\documentclass[10pt]{article}      %

\usepackage[preprint]{tmlr}        %

\usepackage[utf8]{inputenc}        %
\usepackage{inconsolata}           %

\usepackage{microtype}
\usepackage{lmodern}    %
\usepackage{graphicx}
\usepackage{subcaption}
\usepackage{booktabs} %

\usepackage{hyperref}
\hypersetup{
  pdftitle={TabAgent: Distilling LLM Agent Components into Cheap Classifiers},
  pdfauthor={Ido Levy, Segev Shlomov, Avi Yaeli, Nir Mashkif, Asaf Adi, Yinon Goldstein, Eilam Shapira, Leshem Choshen},
  pdfsubject={Replacing the closed-set decisions of LLM agents (tool shortlisting, skill and application selection, routing and gating) with a compact tabular classifier trained on the agent's own successful execution traces},
  pdfkeywords={LLM agents, closed-set decisions, tool shortlisting, tool selection, skill selection, application selection, model routing, tabular classification, distillation, execution traces, AppWorld, tau2-bench, Mind2Web, SkillsBench, LLMRouterBench, CatBoost},
  pdflang={en-US}
}
\ifdefined\pdfgentounicode\input{glyphtounicode}\pdfgentounicode=1\fi

\usepackage{amsmath}
\usepackage{amssymb}
\usepackage{mathtools}
\usepackage{amsthm}

\usepackage[capitalize,noabbrev]{cleveref}

\theoremstyle{plain}

\theoremstyle{definition}

\theoremstyle{remark}

\newcommand\blfootnote[1]{{\renewcommand\thefootnote{}\footnotetext{#1}}}

\usepackage{algorithm}

\usepackage{algpseudocode}
\usepackage{url}
\usepackage{float}
\usepackage{placeins}
\usepackage{colortbl}
\newcommand{\PatR}{\mathrm{P}\!@\!\mathrm{R}}
\usepackage{pgfplots}
\pgfplotsset{compat=1.18}

\usepackage{listings}
\usepackage{xcolor}
\usetikzlibrary{positioning,fit,arrows.meta,backgrounds,calc,shapes.geometric}
\definecolor{tabblue}{HTML}{2F6FB0}
\definecolor{tabink}{HTML}{1A1C22}
\definecolor{tabmuted}{HTML}{5D616B}
\definecolor{tabexp}{HTML}{C0392B}
\definecolor{tabcheap}{HTML}{128A72}
\definecolor{tabcardbg}{HTML}{EEF3F9}
\definecolor{tabzone}{HTML}{F6F7F9}
\definecolor{tabgrey}{HTML}{C8CDD4}
\lstdefinestyle{tabagentprompt}{
  basicstyle=\ttfamily\footnotesize,
  breaklines=true,
  breakatwhitespace=false,
  columns=fullflexible,
  keepspaces=true,
  showstringspaces=false,
  frame=single,
  framerule=0.3pt,
  rulecolor=\color{black!20},
  xleftmargin=1em,
  xrightmargin=0em,
  aboveskip=0.75\baselineskip,
  belowskip=0.75\baselineskip,
  numbersep=6pt,
  numbers=left,
  numberstyle=\tiny\color{black!50}
}
\lstdefinestyle{verbatimwrap}{
  basicstyle=\ttfamily\footnotesize,
  breaklines=true,
  breakatwhitespace=false,
  columns=fullflexible,
  keepspaces=true,
  showstringspaces=false,
  frame=none,
  xleftmargin=0pt,
  xrightmargin=0pt,
  aboveskip=0.5\baselineskip,
  belowskip=0.5\baselineskip
}
\lstdefinelanguage{json}{
    morestring=[b]",
    morecomment=[l]{//},
    literate=
     *{0}{{{\color{black}0}}}{1}
      {1}{{{\color{black}1}}}{1}
      {2}{{{\color{black}2}}}{1}
      {3}{{{\color{black}3}}}{1}
      {4}{{{\color{black}4}}}{1}
      {5}{{{\color{black}5}}}{1}
      {6}{{{\color{black}6}}}{1}
      {7}{{{\color{black}7}}}{1}
      {8}{{{\color{black}8}}}{1}
      {9}{{{\color{black}9}}}{1}
}

\usepackage{tabularx,array}
\usepackage[most]{tcolorbox}
\usepackage{ragged2e}
\newcolumntype{Y}{>{\RaggedRight\arraybackslash}X}

\title{TabAgent: Distilling LLM Agent Components\texorpdfstring{\\}{ } into Cheap Classif{}iers}

\author{\name Ido~Levy\textsuperscript{1,2}, Segev~Shlomov\textsuperscript{1}, Avi~Yaeli\textsuperscript{1}, Nir~Mashkif\textsuperscript{1}, Asaf~Adi\textsuperscript{1}, Yinon~Goldstein\textsuperscript{1}, Eilam~Shapira\textsuperscript{1}, Leshem~Choshen\textsuperscript{1,2} \\
        \addr \textsuperscript{1}IBM \qquad \textsuperscript{2}Weizmann Institute of Science \\[0.6ex]
        \normalfont\small
        \href{https://github.com/dolev31/TabAgent}{\raisebox{-0.2ex}{\includegraphics[height=1.05em]{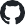}}\,\,Code: \texttt{github.com/dolev31/TabAgent}}
        \qquad
        \href{https://dolev31.github.io/TabAgent/}{\raisebox{-0.2ex}{\includegraphics[height=1.05em]{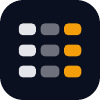}}\,\,Project page: \texttt{dolev31.github.io/TabAgent}}}

\begin{document}
\maketitle
\blfootnote{The authors used a large language model to assist with copy-editing and LaTeX formatting of this manuscript. All ideas, claims, experiments, analyses, and results are the authors' own.}

\begin{abstract}
Agentic systems chain LLM calls to make closed-set decisions: choosing from a fixed candidate set, as in skill selection, tool shortlisting, application selection, and routing. These repeated frontier-model calls dominate per-task latency and cost.
We show such decisions need not be made by a language model at all: a compact discriminative head, trained on the agent's own successful execution traces, can replace the generative one while leaving the surrounding agent intact. \textbf{\textsc{TabAgent}} has three parts. (i) \textbf{TabSchema} extracts schema, state, and dependency features from trajectories. (ii) \textbf{TabSynth} adds schema-aligned synthetic rows where trajectory coverage is thin. (iii) \textbf{TabHead}, a cheap tabular classifier, scores candidates in a single pass.
Changing only the label space, we test \textsc{TabAgent} as a drop-in replacement for the frontier-LLM heads (e.g., GPT-5.4) across three decision families and seven agent benchmarks. On tool shortlisting it recovers essentially all of the tools the GPT-5.4 shortlister used on its successful runs (Recall@9), at $99.96\%$ lower latency and $99.9996\%$ lower cost per call (${\sim}10^{3}\times$ and ${\sim}10^{5}\times$). It also ranks first on the classification decisions and is on par with a strong fixed default on the harder routing and gating decisions at far lower latency and cost, where a costly generative router does worse. Across these tasks, the gains come from the representation, not the model: a discriminative head over raw text or embeddings alone collapses, and trace-derived features make it accurate.
\end{abstract}

\begin{figure}[t!]
  \centering
  \begin{subfigure}[t]{\linewidth}
    \centering
    \includegraphics[width=0.75\linewidth]{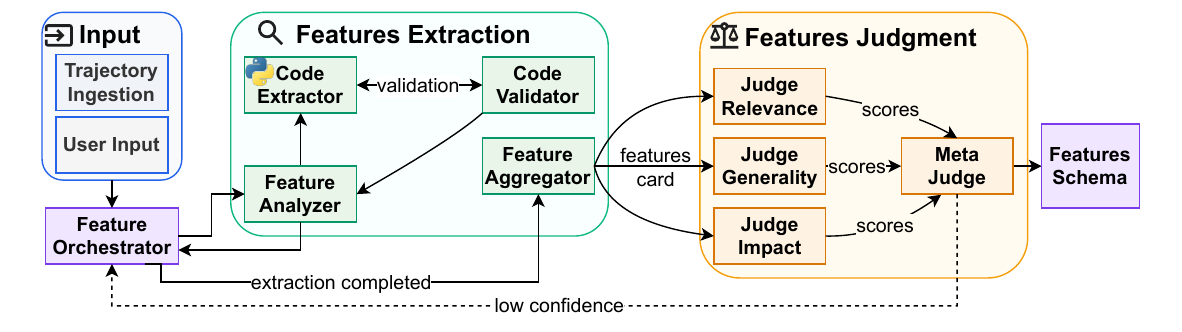}
    \caption{TabSchema: feature extraction}\label{subfig:tabschema}
  \end{subfigure}

  \begin{subfigure}[t]{\linewidth}
    \centering
    \includegraphics[width=0.95\linewidth]{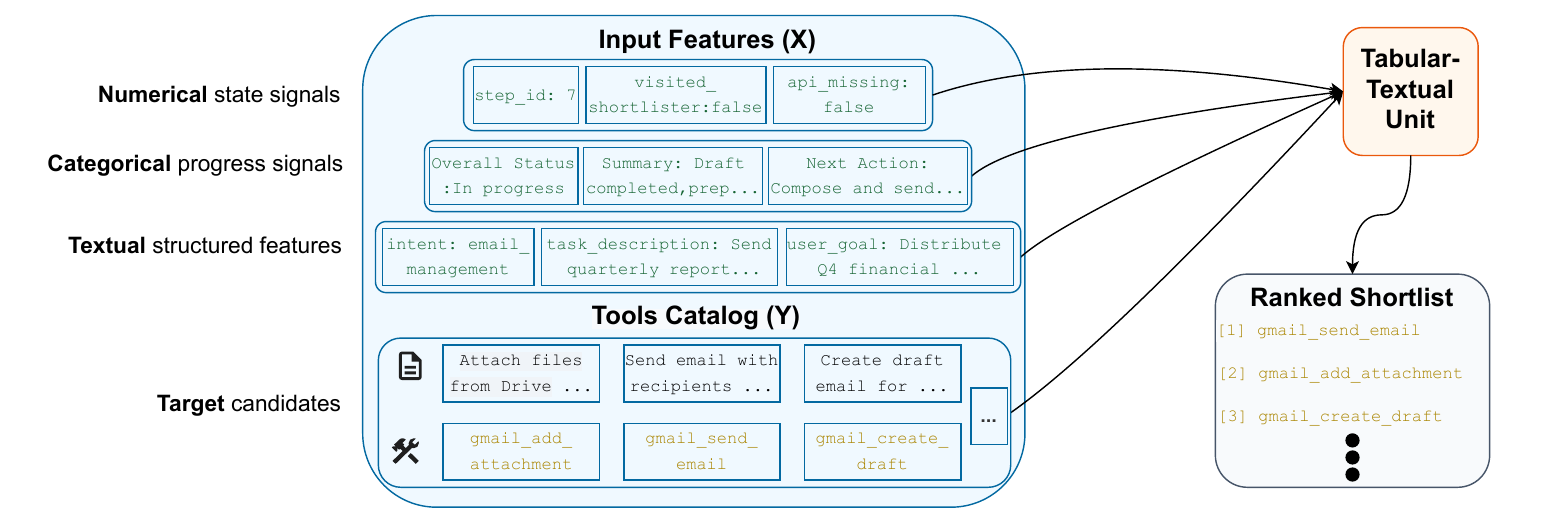}
    \caption{TabHead: single-pass classifier}\label{subfig:tabhead}
  \end{subfigure}

  \caption{(a) TabSchema compiles trajectory-derived schema, state, and dependency signals into tabular features. (b) TabHead consumes features and candidates to output per-candidate probabilities.}
  \label{fig:tabagent}
\end{figure}

\section{Introduction}
\label{sec:intro}

Agentic systems repeatedly call an LLM \citep{schneider2025generative} to make closed-set decisions, which are both expensive and slow. Throughout a long-horizon execution, decisions often require selecting an answer from a fixed, bounded set of candidates: which skill to invoke, which tool to call next, which application to route to, or which sub-task to gate. Tool shortlisting epitomizes this need, the task of picking a small subset from a catalog of hundreds to thousands of tools \citep{li_etal_2023_api, wang2024survey, qu2025tool} where a ``tool'' is any externally callable capability (API endpoint, function, or GUI operation) with a typed schema and observable outputs. Shortlisting is especially costly as serialization can run to tens of thousands of tokens per call. In our measurements, a GPT-5.4 shortlisting call takes $7.5$\,s (App.~Table~\ref{tab:cost-runtime}), and across a long-horizon task's steps it dominates per-task latency and cost.

Prior work cuts this cost along several axes, but each keeps one of two things an agent's execution needs. One line removes the generative call yet stays blind to live state: retrieval selectors \citep{anantha2023protip, hong2026tools} and discriminative selectors \citep{moon2024efficient} score a static catalog or fixed embeddings, ignoring the agent's schema, state, and dependency signals. The other line reads that state yet keeps the generative call: LLM-centric controllers \citep{qintoolllm} leave the loop on the critical path where its latency compounds across steps, and cost-aware routing, cascades, and self-training \citep{chen2024frugalgpt, moslem2026dynamic, zelikman2022star} only make that call cheaper or smaller. An agent, though, emits exactly the state signals the first line ignores, and it emits them while solving its task. We therefore take both at once: following the view that small, specialized models should carry most agentic work \citep{belcak2025small}, we cast each decision's snapshot into a textual-tabular representation, letting a cheap, non-generative discriminator read the live state and replace the generative call.

\textbf{\textsc{TabAgent} replaces a generative decision head in a running LLM agent with a compact classifier trained on the agent's own successful execution traces.} It has two trace-processing stages and one classifier. \textbf{TabSchema} turns execution traces into a tabular feature card: each row is a (task context, candidate tool) pair, each column a behavior-grounded feature (e.g., the agent's current thought, the step index, or whether the candidate has co-occurred with other tools in similar traces), and the label is whether the successful trajectory used the candidate (Fig.~\ref{subfig:tabschema}). \textbf{TabSynth} adds schema-aligned synthetic rows where trajectory coverage is thin, lifting AppWorld shortlisting quality by ${\sim}23\%$ on average. \textbf{TabHead} scores each candidate in one classifier pass (Fig.~\ref{subfig:tabhead}), replacing the multi-call generative shortlister.

We evaluate \textsc{TabAgent} as a drop-in replacement for the frontier-LLM decision heads (e.g., GPT-5.4) in SOTA open-source agents \citep{marreed2025towards, lv2024codeact, yao2022react}, changing only the label space and its feature card, across three closed-set decision families and seven benchmarks \citep{trivedi2024appworld, barres2026taubench, cugaproject2026oakbench, shlomov2026benchmarks, li2026skillsbench, deng2023mind2web, li2026llmrouterbench}. On tool shortlisting it recovers essentially all of the tools GPT-5.4's successful runs used on AppWorld (Recall@9) while replacing that generative call at ${\sim}10^{3}\times$ lower latency and ${\sim}10^{5}\times$ lower cost per call (Fig.~\ref{fig:pareto_new}). It also ranks first on the classification decisions and is on par with a strong fixed default on routing and gating at far lower latency and cost, where a costly generative router does worse (Table~\ref{tab:master_new}).

A small classifier can stand in for a frontier head because the signal lives in the representation rather than the model: given the TabSchema card, heads as different as decision trees and tabular foundation models reach comparable accuracy, and all collapse once the card is reduced to the raw task and candidate text (\S\ref{subsec:card_new}). The head is thus off-the-shelf and interchangeable, and ports across settings. One head moves across LLM backbones within a shared scaffold at a cost of about $3.1$ percentage points of $\PatR$, and a single universal head stays within ${\sim}3$ points of per-task specialists on five of seven decisions (\S\ref{app:universal}). It does not win everywhere: a fine-tuned 8B small LLM scores higher on four of five AppWorld applications but stays in the generative loop at $10^{2}$ to $10^{4}\times$ the per-call cost (\S\ref{subsec:eff_new}), and \textsc{TabAgent} is best on in-distribution decisions, leaving the rare novel cases to the LLM (\S\ref{subsec:scope}). Beyond these offline comparisons, hot-swapping the head into a live agent leaves end-to-end task success on par with the generative shortlister (\S\ref{subsec:inloop}).

\paragraph{Contributions.} We replace an agent's repeated frontier-LLM decision calls with a fast, cheap classifier trained on the agent's own successful traces, at matched decision quality. Concretely: \textbf{(i)}~We cast the closed-set decisions an agent makes repeatedly (tool and skill shortlisting, type and label selection, routing and gating) as a single pointwise classification over trace-derived features, changing only the label space and its feature card. \textbf{(ii)}~We realize this as \textsc{TabAgent}: TabSchema extracts the features from an agent's successful traces, TabSynth augments them where coverage is thin, and TabHead scores candidates in one non-generative pass. \textbf{(iii)}~Across seven benchmarks, eight tasks, and three decision families we show the head matches the frontier shortlister's selections, ranks first on classification, and ties a strong default on routing and gating, all at far lower latency and cost. We attribute this to the representation rather than the head, and characterize the regimes where it is preferable to both frontier and small LLMs.

\section{Related Work}
\label{sec:related}

A replacement for the generative decision call is best judged on the agent's critical path, where that decision actually runs, yet prevailing benchmarks evaluate tool selection offline: \citet{qintoolllm} curate \textsc{ToolBench} (\(\sim16\mathrm{k}\) APIs), \citet{li_etal_2023_api}, \citet{song2023restgpt}, and \citet{tang2023toolalpaca} add multi-turn, REST, and instruction-tuning variants, and \citet{duanytool} stresses hierarchical retrieval at scale. Across these, all report standard correctness metrics (Recall@K, pass/win) \citep{schutze2008introduction} but rarely measure shortlist-time efficiency, latency, and cost, or cross-application dependencies. We therefore evaluate on long-horizon, cross-application environments \citep{trivedi2024appworld}, where shortlisting lies on the critical path and recurs across web, GUI, and skill agents \citep{zhou2024webarena, pmlr-v235-zheng24e, xie2024osworld, wang2024voyager}: one decision problem, not one tied to a single architecture. We judge every approach below on two axes: does it condition on the agent's live execution state, and does it avoid a generative call?

One line of work avoids the LLM but scores a static catalog. Retrieval selectors treat tools as documents for sparse and dense retrieval \citep{sparck1972statistical,robertson2009probabilistic,reimers2019sentence,karpukhin_etal_2020_dense,xiongapproximate,khattab2020colbert} and share three blind spots: they are state-blind (ignore task progress and preconditions), set-blind (score items independently), and structure-blind (neglect hierarchies and argument schemas). Structure-aware methods add supervision and hierarchy: \citet{anantha2023protip} learn prototypical, progressive matching, \citet{zheng2024toolrerank} add hierarchy-aware reranking, and \citet{qu2024towards} optimize for shortlist completeness, while cross-encoder \citep{nogueira2020document} and generative \citep{wang2025toolgen} retrievers improve matching by decoding tool identities from catalog text. All improve quality but add inference stages, and remain set- and state-blind, never conditioning on the agent's live execution state. Closest within this line, \citet{moon2024efficient} also cast shortlisting as pointwise classification, but over static (query, tool) embeddings estimated offline and evaluated on retrieval benchmarks, sharing the same state-blindness that \textsc{TabAgent} avoids. Recent practical frameworks also target tool-selection cost: \citet{liu2026toolscope} merge redundant tools and filter the catalog by context to fit tighter budgets, and \citet{jia2026autotool} exploit tool-usage inertia by mining a transition graph from historical trajectories to bypass repeated tool-selection calls. Both reduce cost while keeping selection catalog- or sequence-driven. \textsc{TabAgent} instead learns a discriminative classifier over rich TabSchema state and dependency features and targets set-completeness rather than next-tool inertia.

A second line keeps the LLM in the loop. LLM-centric approaches train or prompt the agent to choose tools directly \citep{hu2026agentic, schick2023toolformer, shen2023hugginggpt, patil2024gorilla}: \citet{qintoolllm} fine-tune 7--13B controllers that traverse tool abstraction levels (Category $\rightarrow$ Tool $\rightarrow$ API) with DFS. \citet{liu2024toolnet} navigate a tool graph with an LLM while an evaluator updates transition weights, and \citet{gao2024confucius} apply curriculum learning over tool complexity. These raise tool-use accuracy but keep a generative controller on the critical path, incurring multiple LLM calls per decision through chain or tree search.

A cost-focused branch of that second line keeps the generative call and only reduces its cost: classical compression (distillation, pruning, quantization) \citep{hinton2015distilling,sanh2020movement,hulora} cuts per-call compute but not the number of generations, cost-aware routing and cascades \citep{chen2024frugalgpt, ong2025routellm, ding2024hybrid}, including a recent experience-conditioned self-router \citep{zhang2026evoroute}, change which model runs, and trajectory-based fine-tuning, self-training, and distillation \citep{zelikman2022star, singh2024beyond, chen2023fireact, zeng2024agenttuning, yin2024agent} adapt or shrink the policy. In every case the deciding component stays generative.

Where the first line avoids the LLM but scores a static catalog, and the second reduces the cost of the LLM without removing it, we do both: condition on live execution state with no generative call. The decision head becomes a non-generative classifier, and the LLM is kept only as a confidence-gated fallback, not as every arm of a cascade. The contribution is the representation, a behavior-grounded feature card distilled from the agent's own successful traces (schema, state, dependency, and thought signals), building on evidence that discriminative and encoder models can rival generative ones at a fraction of the cost \citep{galke2022we, saattrup_nielsen_etal_2025_encoder}. The head is swappable: we use the gradient-boosted decision tree CatBoost \citep{CatBoost}, and heads from classical trees to tabular foundation models \citep{arazi2025tabstar, hegselmann2023tabllm, kim2024carte, huang2020tabtransformer, FT_Transformer, TP_BERTa, hollmanntabpfn} perform comparably given the same card (\S\ref{subsec:card_new}).

\section{TabAgent}
\label{sec:method}

\textsc{TabAgent} learns the decision boundary of a closed-set generative head from successful trajectories an LLM agent has already produced, then runs that boundary as a single-pass classifier in place of the LLM. The reduction binds only to agent outputs (traces), not to the agent's internals (architecture, model weights), so the same pipeline (trace to features, candidates to labels) applies across decision heads such as tool shortlisting, skill and application selection, and routing without modifying the surrounding agent. Motivated by evidence that LLM-generated rationales can supervise smaller models \citep{hsieh2023distilling}, our central hypothesis is that execution traces from successful LLM agents encode raw decision-relevant signals in a structured, tabular-representable form that a compact discriminative classifier can extract and use to reproduce the agent's decision. We refer throughout to the resulting tabular layout as a feature card, with one row per (context, candidate) pair and one column per behavior-grounded feature.

\subsection{TabSchema: Agentic Feature Extraction}
\label{sec:tabschema}
TabSchema replaces human-curated features with an orchestrated agentic system that proposes features, implements them as executable trace-extraction code, and keeps only those a judge ensemble agrees are decision-relevant, extending LLM-driven feature engineering \citep{hollmann2023large} from static tabular datasets to agent execution traces. The loop is hierarchical: a FeatureOrchestrator coordinates an analyzer (proposes features), an executor (writes Python extraction code in a sandbox), a validator (an LLM-as-a-judge pass that checks the code, or signals agreement that the extracted features follow the analyzer's plan), and a judge ensemble that scores features along multiple quality dimensions \citep{madaan2023self}. Concretely, the analyzer enumerates trace-level signals (e.g., thought snippets) and trace-structural signals (e.g., co-usage windows), each annotated with type, range, and a task-relevance score. We use an LLM extractor as a post-hoc explainer of agent decisions \citep{kroeger2023are}: the features we surface are textual interpretations of execution-trace fields (agent thoughts, status analyses, action rationales), which require language understanding. By verifying feature feasibility through executable code, TabSchema produces reusable extractors: Python feature-extraction functions that can be re-applied when onboarding a new agent or catalog. TabSchema requires only that the agent's trace schema be stable across the trajectory pool used for onboarding. Trace-format details per benchmark are in App.~\ref{app:benchmarks}. Full prompts, retry and scoring policies, and per-component engineering decisions are in App.~\ref{app:TabSchema}.

A single LLM judge tends to inflate one quality dimension at the expense of others, so TabSchema uses an ensemble of LLM judges reconciled by a MetaJudge \citep{zheng2023judging,liu2023g,li2025leveraging}. Unlike LLM-as-a-judge work that scores model outputs \citep{li2024llms, gu2024survey}, our ensemble scores candidate input features, which turns feature engineering into an auditable agentic process. Reproducibility of the resulting feature card holds across independent reruns with the same LLM (72--95\% semantic feature overlap across three reruns on OAK and BPO) and across frontier LLMs running the full pipeline independently (App.~\ref{app:model_dependency}). Onboarding (TabSchema plus TabSynth) is a one-time cost per (agent, catalog) pair, amortized across all subsequent inference calls. At inference, low-confidence TabHead predictions defer to the LLM shortlister for novel compositions (\S\ref{sec:results_new}).

\subsection{TabSynth: Schema-aligned data augmentation}
\label{sec:tabsynth}

In practice, real trajectories under-represent rare but decision-critical patterns (e.g., argument-compatibility), which limits coverage when training a compact discriminative head. To close this gap, we introduce TabSynth, a schema-aligned augmentation module that expands supervision while preserving the behavior-grounded signals surfaced by TabSchema.
TabSynth augments the rows produced by TabSchema with a deterministic, rule-based paraphraser: it rewrites the free-text state of an existing (context, candidate) training row, for example the agent's thought and status summaries and the task text, while copying the candidate, the label, and every schema, state, and dependency feature verbatim.
Each real row yields two to four paraphrase variants on average, exact duplicates of real rows are removed, and the rest are mixed into the training set. Because the label is never reassigned, the augmentation adds no new (context, candidate) relationships and cannot fabricate a positive. Instead, it widens coverage of rare but decision-critical patterns by presenting them in more surface forms, unlike LLM-based tabular synthesizers that generate new rows \citep{wang2024harmonic, barr2025large}. Full procedure and a worked example are in App.~\ref{app:TabSynth}.

\subsection{TabHead: Textual--Tabular Classification}
\label{sec:tabhead}
TabHead scores each (context, candidate) pair over the TabSchema feature card: the tabular schema, state, and dependency features together with text-derived features of the candidate, formed from its name, description, and argument hints (Fig.~\ref{subfig:tabhead}, example in App.~\ref{app:features_card_example}). We use a gradient-boosted decision tree (CatBoost) as the head because it handles mixed numerical and text-derived columns well and trains in seconds from the small per-task trajectory pools in our setting. Heads from classical trees to text-ingesting tabular foundation models (including TabSTAR) perform comparably given the same card, so \textsc{TabHead} is CatBoost in every main-paper result and the neural heads appear only in the head-invariance ablation (\S\ref{subsec:card_new}, App.~\ref{app:head_alt}). Training mixes real trajectories with schema-aligned TabSynth rows to widen coverage within the schema.

We handle three decision patterns through one pointwise reduction. For single-class tasks (e.g., task complexity identifier), we train a softmax over $\mathcal{Y}$. For multi-label problems (e.g., application selection), we use binary relevance: for each example $x$ and label $y\in\mathcal{Y}$, we build a row $(x,y)$ and predict $p(y\mid x)$, and thresholding produces $\hat{Y}$ \citep{tsoumakas2008multi}. For ranking (e.g., tool shortlisting), we again score $(x,y)$ pairs: during training we include one row per candidate, labeled positive when $y\in Y_t$, and at inference we sort all candidates by $p(y\mid x)$ and return the top-$K$ \citep{liu2009learning}. At inference, TabHead's single classifier pass replaces the multi-call LLM controller's billions of parameters and long-context prompt, which is what makes each call ${>}10^{3}\times$ faster and ${>}10^{5}\times$ cheaper than the GPT-5.4 shortlister, as reported in \S\ref{sec:results_new}. When the upstream agent's behavior shifts (a new prompt, a new base model) but its trace schema is unchanged, only TabHead is refit on the new traces, which takes about 45 seconds on CPU, so drift adaptation does not require a re-engineering pass. A change that alters the feature schema itself requires re-running the one-time TabSchema and TabSynth onboarding. Because TabHead emits a confidence score per candidate, it supports a hybrid deployment: low-confidence decisions route to the original LLM shortlister, so the head absorbs in-distribution traffic while the LLM covers rare novel cases. TabHead's confidence separates correct from incorrect decisions well enough to gate (AUROC $0.88$, App.~\ref{app:error_analysis}).

\section{Experimental Setup}
\label{sec:setup}

\subsection{Setup Specifications}

We evaluate \textsc{TabAgent} on seven agent benchmarks that together span the three closed-set decision families an agent repeatedly faces. \textbf{Set retrieval}: tool shortlisting on AppWorld \citep{trivedi2024appworld}, $\tau^2$-bench \citep{barres2026taubench}, and the enterprise benchmarks OAK \citep{cugaproject2026oakbench} and BPO \citep{shlomov2026benchmarks}, and skill selection on SkillsBench \citep{li2026skillsbench}. \textbf{Classification}: application selection on AppWorld, and operation and GUI element selection on Mind2Web \citep{deng2023mind2web}. \textbf{Policy and reward}: sub-agent routing on AppWorld, model routing on LLMRouterBench \citep{li2026llmrouterbench}, and skill gating on SkillsBench. The benchmarks vary widely in candidate-set size, from a binary gate to $195$ skills, and in catalog semantics and horizon, so they test the discriminative replacement under heterogeneous conditions. Per-benchmark candidate catalogs, trajectory pools, and evaluation protocols appear in App.~\ref{app:benchmarks}.

Our goal is to replace a validated but expensive generative component of a running agent (tool shortlisting is the running example) with an efficient discriminative head (a gradient-boosted tree) trained on features extracted from the agent's trajectories, rather than to invent new component logic. This follows a practical deployment lifecycle: (i) run the agent with trajectory logging enabled, (ii) collect execution data from successful tasks and extract TabSchema features, and (iii) swap the generative module for the discriminative TabHead, preserving the surrounding agent architecture. Benchmarks distributed as fixed datasets are handled identically, with features extracted from the provided records rather than from live runs. We carry out this hot-swap on a live CUGA agent and measure end-to-end task success in \S\ref{subsec:inloop}.

Among the open-source agents we instrument, CUGA \citep{marreed2025towards} leads on AppWorld, so we adopt it as our reference agent and replace only its shortlisting module, extracting TabSchema features from its trajectories and training TabHead on top of them. We additionally evaluate \textsc{CodeAct} \citep{lv2024codeact} as a text-only control whose sole input is the task description, isolating the contribution of TabSchema features, and for the cross-backbone transfer study on $\tau^2$-bench (\S\ref{subsec:retrieval_new}) we instrument four agents that share a ReAct scaffold \citep{yao2022react} and differ only in their LLM backbone (Claude-3.7-Sonnet, GPT-4.1, GPT-4.1-mini, and o4-mini), a controlled set that isolates the backbone as the sole variable. The transfer study therefore varies the backbone by design, while GPT-5.4 remains the backbone of our CUGA reference agent in every CUGA experiment. Focusing on a strong reference agent keeps instrumentation consistent, removes cross-stack confounds, and matches our aim of lower cost and latency.

Labels follow each decision's ground truth. Where a benchmark ships no shortlist (AppWorld, OAK, and BPO tool shortlisting, and the $\tau^2$-bench transfer study), we derive labels from successful trajectories: for each task \(t\), the relevant set \(G(t)\) is the set of tools used in a canonical successful solution, matching our aim of reproducing the choices that already lead to success. The per-domain $\tau^2$-bench rows instead label each logged trajectory by the tools it called, regardless of outcome (App.~\ref{app:taubench_details}). The remaining benchmarks carry independent ground truth: Mind2Web's element and operation labels are human-annotated, and SkillsBench and LLMRouterBench supply their own targets and reward outcomes, which we use directly. Trajectory-derived labeling, and with it any dependence on the source agent's own choices, therefore applies only to the four tool-shortlisting settings and to AppWorld application selection, whose label is the set of applications a successful trajectory touches, not to operation or element selection, routing, or gating. For AppWorld this yields $605$ successful CUGA trajectories (App.~\ref{app:appworld_setup_specs}).

\subsection{Implementation Details}

We train the TabHead classifier (CatBoost) with one fixed configuration for every decision ($300$ boosting iterations, learning rate $0.08$, tree depth $6$, and CatBoost's defaults otherwise, including L2 leaf regularization $3.0$), with the loss matched to the decision type: logistic for binary and pointwise ranking, softmax for single-class. Training takes about $45$\,s on a single CPU. The neural alternative head used only in the head ablation (App.~\ref{app:head_alt}) is fine-tuned separately.
We compare \textsc{TabAgent} against a shared baseline ladder applied at every decision type (each scored with that decision's native metric), running from a chance floor up through three families of learned baselines. On the policy/reward decisions the few-shot-LLM rung uses GPT-5.4 rather than Llama, and the fine-tuned small-LLM rung (FT-SLM) is Llama-3.1-8B. The first family is sparse retrieval with Okapi BM25, a classical retrieval model that scores tools by lexical overlap between the task description and tool documentation \citep{robertson2009probabilistic, sparck1972statistical}.

The second family employs dense semantic retrieval (DSR) (semantic similarity) through a bi-encoder semantic setup \citep{karpukhin_etal_2020_dense} in three configurations: Zero-Shot ranks tools using off-the-shelf sentence embeddings without task-specific training, FT (Real) applies contrastive fine-tuning on real agent trajectories \citep{chen2020simple}, and FT (Real+synth) combines real trajectories with TabSynth supervision during training. All variants use the same e5 encoder as TabHead for fair comparison \citep{wang2022text}. Fine-tuned variants optimize a contrastive objective with in-batch negatives (application selection instead fine-tunes a binary classifier, App.~\ref{app:application_selection}), learning to align task/context representations with relevant tool descriptions in a shared embedding space \citep{reimers2019sentence}. At inference, candidates are scored by task-tool inner product, returning the top-$K$ results.

The third family is generative LLM controllers. Since our goal is a compute-efficient replacement, we test whether small language models (SLMs) can generate the shortlist themselves rather than deferring to a frontier model: we evaluate Llama 3.2 1B/3B, Llama 3.1 8B \citep{dubey2024llama}, and Qwen2.5 7B \citep{qwen2_5} in two regimes each, few-shot and fine-tuned, consuming CUGA's original shortlister state to produce a ranked shortlist.
Each model processes the complete catalog for the active app, including tool names, argument signatures, and descriptions.

\subsection{Evaluation metrics}
\label{sec:metrics}

We evaluate using 5-fold cross-validation at the task level. Folds are stratified by application or domain and, where applicable, by deciles of \(|G(t)|\), the number of relevant tools per task. All feature rows and synthetic augmentations derived from a given task remain within a single fold to prevent leakage. Grouping rows by task is necessary but not sufficient, because the representation is itself learned: every data-dependent stage is therefore fitted inside the training portion of each fold and never sees held-out tasks. This covers the TabSchema loop (feature proposal, aggregation, judging, and extraction-code construction), the data-dependent statistics those features contain (candidate priors and co-usage counts), the TF--IDF and embedding vocabularies, and TabSynth generation. Only the resulting extractors are applied to the held-out fold. Every feature is read from what the head would see at the moment of the decision (the task, the agent's state, and earlier steps). Nothing written after the decision, such as the tools later called or the run's outcome, enters the card, and the released code enforces this with leakage tests. In each round, we train on 4 folds and evaluate on the held-out fold, then macro-average metrics across tasks over the 5 test folds.
The non-fine-tuned baselines are evaluated without training. For methods with stochastic behavior, each task is run over 3 seeds and the task-level scores are averaged.
Unless noted otherwise, we report macro averages over tasks with 95\% confidence intervals. Statistical significance is assessed by paired resampling over tasks with Holm--Bonferroni control within each metric across the 5 applications (App.~\ref{app:signif-variance}).

\textsc{TabAgent} instantiates multiple decision heads. We therefore evaluate with head-matched metrics: (i) single-class classification (e.g., task-complexity prediction), (ii) multi-label classification (e.g., application identification), and (iii) top-$k$ for tool shortlisting.
For Top-$k$ shortlisting, consider each task $t\in T$, let $G(t)$ denote the ground-truth tool set with size $R_t = |G(t)|$. Let $\pi(t)$ be the system ranking in the catalog and $S_k(t)$ the prefix $k$ of the top of $\pi(t)$. Tied scores are ordered uniformly at random (App.~\ref{app:evaluation}).

\textbf{Recall@\texorpdfstring{$k$}{k}} evaluates recovery at a fixed budget:
\[
\mathrm{Recall}@k
= \frac{1}{|T|}\sum_{t\in T} \frac{|S_k(t)\cap G(t)|}{|G(t)|}.
\]

\textbf{R-precision (\texorpdfstring{$P@R$}{P@R})} adapts the cutoff to the task's relevant-set size and captures completeness:
\[
P@R
= \frac{1}{|T|}\sum_{t\in T} \frac{|S_{R_t}(t)\cap G(t)|}{R_t}.
\]
$P@R$ equals $\mathrm{Recall}@R$ because both reduce to $|S_{R_t}(t)\cap G(t)|/R_t$ per task. We use the $P@R$ name for IR consistency \citep{schutze2008introduction}. Throughout, we foreground $P@R$ for completeness claims and use $\mathrm{Recall}@k$ to assess performance under fixed production budgets. See App.~\ref{app:evaluation}.

\section{Results}
\label{sec:results_new}

Table~\ref{tab:master_new} reports \textsc{TabAgent} against the shared baseline ladder at every decision type, and we then isolate what carries the signal and what the replacement costs. Two findings recur: quality is driven by the trace-derived feature card and is largely invariant to the classifier head, and a single cheap classifier pass matches frontier-LLM decisions at a fraction of their latency and cost.

\begin{table}[t]\centering\small\setlength{\tabcolsep}{3.5pt}
\resizebox{\textwidth}{!}{%
\begin{tabular}{lllrrrrr>{\color{gray}}r>{\columncolor{gray!14}}r>{\columncolor{gray!14}}rr}
\toprule
Decision & Domain & Metric & Chance & BM25 & DSR-ZS & DSR-FT & Few-shot & FT-SLM & \textbf{TabAgent} & \textbf{+synth} & R@k$\uparrow$ \\
\midrule
\midrule\multicolumn{9}{l}{\textbf{SET RETRIEVAL --- rank/return a candidate subset \,\textperiodcentered\, metric = P@R / Recall@9 ($\uparrow$)}} & & & \\
\multicolumn{9}{l}{\emph{AppWorld \,\textperiodcentered\, tool shortlisting [replaces GPT-5.4 shortlister]}} & & & \\
amazon & AZ & P@R & 4 & 28 & 39 & 45 & 56 & 86 & 57 & \textbf{71}~{\color{green!60!black}$\blacktriangle$} & 97@9 \\
gmail & GM & P@R & 6 & 19 & 18 & 52 & 45 & 81 & 58 & \textbf{66}~{\color{green!60!black}$\blacktriangle$} & 95@9 \\
phone & PH & P@R & 5 & 35 & 30 & 57 & 38 & 91 & 66 & \textbf{90}~{\color{green!60!black}$\blacktriangle$} & 100@9 \\
simple\_note & SN & P@R & 20 & 42 & 55 & 63 & 70 & 100 & 62 & \textbf{96}~{\color{green!60!black}$\blacktriangle$} & 100@9 \\
spotify & SP & P@R & 3 & 28 & 30 & 46 & 67 & 67 & \textbf{70}~{\color{green!60!black}$\blacktriangle$} & 61 & 92@9 \\
\multicolumn{9}{l}{\emph{$\tau^2$-bench \,\textperiodcentered\, tool shortlisting}} & & & \\
airline & Airline & P@R & 21 & 28 & 30 & 44 & 44 & 60 & \textbf{72}~{\color{green!60!black}$\blacktriangle$} & 71 & 97@9 \\
retail & Retail & P@R & 28 & 33 & 48 & 42 & 59 & 57 & \textbf{75}~{\color{green!60!black}$\blacktriangle$} & \textbf{75} & 96@9 \\
telecom & Telecom & P@R & 26 & 41 & 46 & 28 & 55 & 71 & \textbf{85} & \textbf{85}~{\color{green!60!black}$\blacktriangle$} & 100@9 \\
combined & Airline \textperiodcentered Retail \textperiodcentered Telecom & P@R & 10 & 28 & 34 & 42 & 31 & 57 & 77 & \textbf{78}~{\color{green!60!black}$\blacktriangle$} & 97@9 \\
\multicolumn{9}{l}{\emph{Enterprise Benchmarks (single-domain, low-data)}} & & & \\
OAK & Healthcare & P@R & 22 & 32 & 35 & 48 & \textbf{56} & 55 & 44 & 51~{\color{red!70!black}$\blacktriangledown$} & 85@9 \\
BPO & Talent Acquisition & P@R & 21 & 60 & 57 & 45 & 57 & 26 & 34 & \textbf{61}~$=$ & 97@9 \\
\multicolumn{9}{l}{\emph{SkillsBench \,\textperiodcentered\, skill shortlisting}} & & & \\
Skill selection & Agent Skills (195) & P@R & 1 & 65 & 51 & 74 & 33 & 52 & \textbf{80}~$=$ & -- & 91@9 \\
\midrule\multicolumn{9}{l}{\textbf{CLASSIFICATION --- pick a type/label \,\textperiodcentered\, metric = macro-F1 / P@R ($\uparrow$)}} & & & \\
Application selection & AppWorld (multi-label) & F1 & 21 & 75 & 30 & 92 & 88 & 96 & \textbf{97}~{\color{green!60!black}$\blacktriangle$} & -- & -- \\
Operation selection & Mind2Web (3-way) & F1 & 30 & 55 & 28 & 84 & 70 & 85 & \textbf{92}~{\color{green!60!black}$\blacktriangle$} & -- & -- \\
\multicolumn{9}{l}{\emph{Mind2Web \,\textperiodcentered\, GUI element selection}} & & & \\
Entertainment & Entertain. & R@3 & 6 & 13 & 11 & 15 & 15 & 31 & 58 & \textbf{68}~{\color{green!60!black}$\blacktriangle$} & 82@5 \\
Shopping & Shopping & R@3 & 6 & 18 & 13 & 18 & 14 & 48 & 65 & \textbf{74}~{\color{green!60!black}$\blacktriangle$} & 84@5 \\
Travel & Travel & R@3 & 6 & 22 & 12 & 23 & 20 & 41 & 64 & \textbf{72}~{\color{green!60!black}$\blacktriangle$} & 83@5 \\
\midrule\multicolumn{9}{l}{\textbf{POLICY / REWARD --- best-arm or gate} \,\textperiodcentered\, metric = route@1 / reward ($\uparrow$)} & & \cellcolor{white}\emph{best default} & \emph{oracle} \\
Sub-agent routing & AppWorld (14 sys) & route@1 & 16 & 9 & 14 & 31 & 23 & 35 & \textbf{38}~$=$ & \cellcolor{white}36 & 58 \\
Model routing & LLMRouterBench (de-leaked) & route@1 & 43 & 38 & 49 & 60 & 50 & 70 & \textbf{71}~$=$ & \cellcolor{white}70 & 94 \\
Skill gating & SkillsBench (activate?) & reward & 46 & -- & -- & -- & 43 & 43 & \textbf{48}~$=$ & \cellcolor{white}50 & 56 \\
\bottomrule
\end{tabular}}
\caption{One trajectory-trained \textsc{TabAgent} pipeline replaces the frontier-LLM decision at every closed-set decision type. All values are percentages, and --~marks not applicable. The shaded columns are ours: \textsc{TabAgent} (real trajectories only) and +synth (with TabSynth). FT-SLM (fine-tuned Llama-3.1-8B, gray) is a costly reference at ${\sim}10^{2}$--$10^{4}\times$ our cost and is excluded from \textbf{bold} (best method in row) and from the arrows. R@k is \textsc{TabAgent}'s relaxed recall (+synth where reported). On policy/reward rows the last two columns are reference points, not methods: the \emph{best default} is the single best choice applied to every task (one sub-agent or model for all tasks, or always activating the skill), and the \emph{oracle} is the share of tasks any candidate solves. Each arrow compares our better score with the best other method in the row or, on policy/reward rows, with the best default: {\color{green!60!black}$\blacktriangle$}~leads, {\color{red!70!black}$\blacktriangledown$}~trails, $=$~ties within a point or the paired 95\% CI.}
\label{tab:master_new}
\end{table}

\label{subsec:retrieval_new}
\textbf{On tool and skill shortlisting, a central bottleneck in agent execution, \textsc{TabAgent} recovers essentially all of the frontier shortlister's selections on AppWorld and leads or ties every deployable retrieval baseline, and a head trained on one backbone transfers to the others within a shared scaffold.} The agent must narrow a catalog of tens to hundreds of candidates to the handful a task needs, and we evaluate with 5-fold cross-validation at the task level (\S\ref{sec:metrics}). Across the five AppWorld applications (Table~\ref{tab:master_new}, full per-method breakdown in App.~Table~\ref{tab:perf_new}), \textsc{TabAgent}(+synth) reaches Recall@9 of 92--100 and Recall@7 $\ge 88$. Under the stricter $\PatR$, which counts only the required tools ranked within the task's relevant-set budget, \textsc{TabAgent} leads or ties every baseline at comparable cost, with its lead over the strongest synth-augmented retriever significant only on Gmail (App.~\ref{app:signif-variance}). Only fine-tuned SLMs score slightly higher, at ${\sim}100$--$250\times$ the cost even on a GPU. BM25's collapse shows that lexical overlap alone cannot recover a set defined by semantic, state, and dependency structure. TabSynth lifts \textsc{TabAgent} $\PatR$ by $23\%$ on average, concentrated in domains where limited trajectory coverage causes the shortlister to miss required selections (SimpleNote $+55\%$, Phone $+36\%$).

This generalizes beyond AppWorld. On $\tau^2$-bench (Table~\ref{tab:master_new}, per-domain detail in App.~Table~\ref{tab:taubench_new}), per-domain \textsc{TabAgent}s reach Recall@9 of 96--100 across airline, retail, and telecom, and a single \textsc{TabAgent} trained jointly over the merged 38-tool catalog reaches 97\%. TabSynth's effect is small and mixed here ($-0.9\%$ to $+1.3\%$ $\PatR$). Two enterprise benchmarks with small trajectory pools, OAK (healthcare) and BPO (talent acquisition), reach Recall@9 of 85 and 97, with some of the largest relative synthesis gains in the beyond-AppWorld suite, consistent with augmentation mattering most in low-data regimes. Table~\ref{tab:transfer_new} shows that \textsc{TabAgent} transfers across agents. When trained on traces from one ReAct agent and evaluated on another, performance decreases by only $\approx3.1$ points of $\PatR$ (own-agent mean: $0.766$, cross-agent mean: $0.735$). This small penalty means a \textsc{TabHead} trained on one backbone transfers to the others: the learned boundary tracks each task's true tool requirements rather than idiosyncrasies of the LLM that generated the traces. Because these four agents share one ReAct architecture, this is transfer within an architecture, not evidence that \textsc{TabAgent} is architecture-agnostic, a limitation we state at the end of this section.

\begin{table}[t]
\centering
\small
\setlength{\tabcolsep}{6pt}
\renewcommand{\arraystretch}{1.05}
\caption{\textbf{Tool shortlisting transfers across ReAct-based agents on $\tau^2$-bench.} Four agents share one ReAct scaffold and environment, differing only in the LLM backbone. Each cell trains on the row agent's successful trajectories and evaluates on the column agent's held-out tasks ($\PatR$, task-disjoint, \textbf{bold} diagonal is same-agent cross-validation). Swapping the trace-generating LLM costs only ${\approx}3.1$ points on average (own-agent mean $0.766$ vs.\ cross-agent mean $0.735$), and several transfer cells match or exceed the target agent's own score (e.g.\ o4-mini${\to}$GPT-4.1-mini, $0.753 > 0.746$).}
\label{tab:transfer_new}
\begin{tabular}{l cccc}
\toprule
& \multicolumn{4}{c}{\textbf{Test agent} ($\PatR$)} \\
\cmidrule(lr){2-5}
\textbf{Train agent} & Claude-3.7 & GPT-4.1 & GPT-4.1-mini & o4-mini \\
\midrule
Claude-3.7-Sonnet & \textbf{0.761} & 0.683 & 0.754 & 0.680 \\
GPT-4.1           & 0.749 & \textbf{0.789} & 0.758 & 0.744 \\
GPT-4.1-mini      & 0.775 & 0.693 & \textbf{0.746} & 0.685 \\
o4-mini           & 0.769 & 0.774 & 0.753 & \textbf{0.769} \\
\midrule
\multicolumn{5}{l}{\footnotesize own-agent (diagonal) mean $0.766$ \quad cross-agent (off-diagonal) mean $0.735$ \quad transfer penalty ${\approx}3.1$ pp} \\
\bottomrule
\end{tabular}
\end{table}

\begin{figure}[t]
  \centering
  \includegraphics[width=\linewidth]{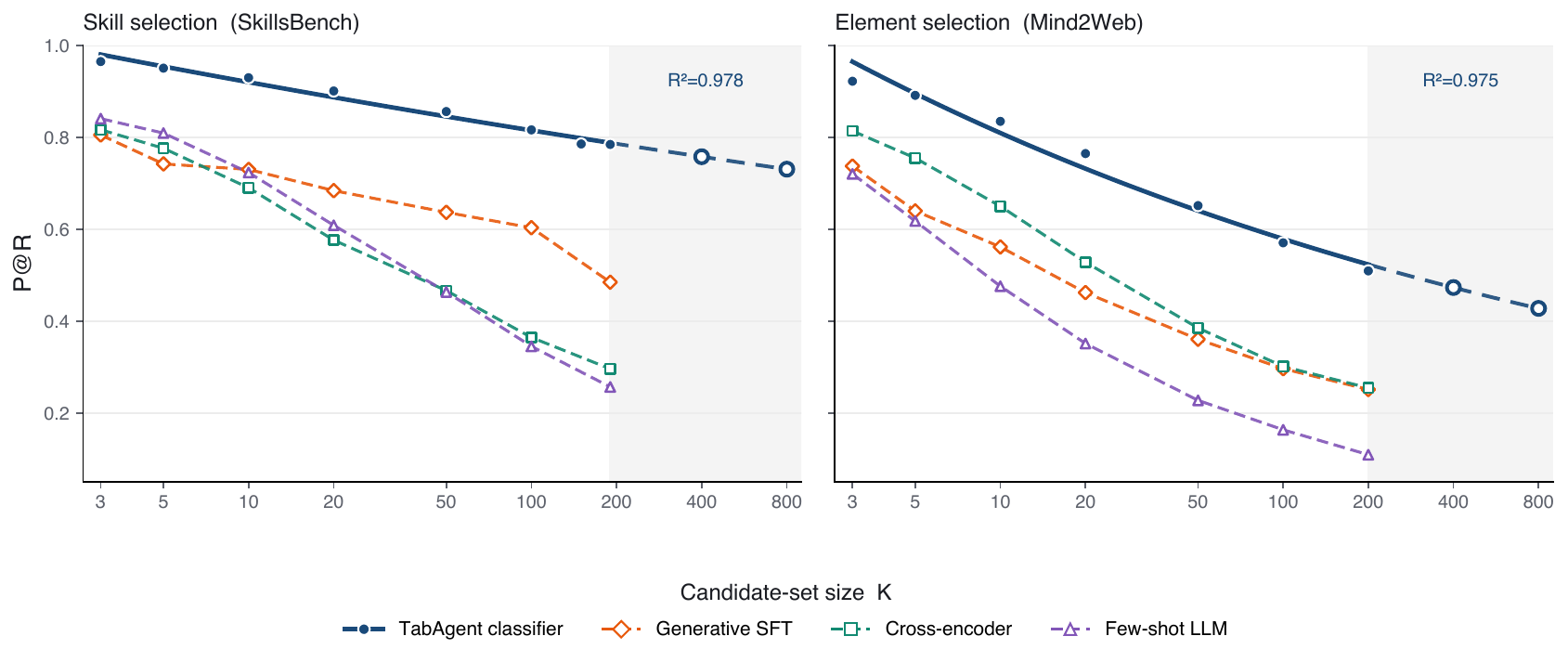}
  \caption{Resilience to candidate overload. As candidate-set size $K$ grows, every LLM adaptation regime (generative SFT, cross-encoder, and few-shot) degrades sharply on both skill selection (left) and Mind2Web element selection (right), whereas the \textsc{TabAgent} classifier follows a shallow power law (solid, $R^{2}\!\ge\!0.97$) and stays highest at every measured $K$. The fitted trend is projected to $K\!=\!400,800$ (open markers).}
  \label{fig:overload_new}
\end{figure}

On SkillsBench (195 candidate skills), \textsc{TabAgent} reaches $\PatR$ 80 and Recall@9 91 (Table~\ref{tab:master_new}), above every retrieval and LLM baseline, and the margin widens as the candidate set grows: every LLM adaptation regime degrades sharply with candidate-set size $K$, while \textsc{TabAgent} stays highest at every measured $K$, and its fitted power-law trend extrapolates to larger catalogs (Fig.~\ref{fig:overload_new}a). We attribute this resilience to scoring each candidate against fixed trace-derived features, so the per-candidate cost is constant and the total cost grows only linearly in the number of candidates, unlike an LLM that must reason over the entire option list in-context, whose prompt grows with the catalog and can exceed 100K tokens for the options alone.

\textbf{\textsc{TabAgent} also replaces the type- and label-classification heads, leading on application selection, operation selection, and GUI element selection.} Application selection is multi-label, where each task maps to the set of applications it touches, and \textsc{TabAgent} reaches macro-F1 97 (App.~\ref{app:application_selection}). The label space is small, so even simple heads do well, but the discriminative swap preserves that quality without a generative call. Operation selection on Mind2Web is a three-way type choice where \textsc{TabAgent} achieves an $F_1$ score of $92$, a $10\%$ relative gain over the best retrieval baseline and $8\%$ over the fine-tuned 8B SLM.

 GUI element selection, picking the target element among many on a webpage, is the hardest case, as the GUI elements are not necessarily repetitive: with domain-matched TabSynth (paraphrasing within each web domain), at the strict $R{=}3$ cutoff \textsc{TabAgent} leads all baselines across entertainment, shopping, and travel (68, 74, 72). Stress testing the method shows that the candidate-overload trend recurs here too (Fig.~\ref{fig:overload_new}b). The difficulty lies in large, visually similar option sets where lexical cues stay informative.

\textbf{On the harder policy and reward decisions, \textsc{TabAgent} matches a strong fixed default at a fraction of the cost, so these decisions need no generative model either.} The policy and reward family, choosing the best arm or gating an action, is intrinsically harder to score, because its ceiling is not the router's to raise: the oracle, the fraction of tasks that any candidate can solve, is only 56--94\% (Table~\ref{tab:master_new}). We therefore report the normalized score $\mathrm{Norm}=(\text{method}-\text{chance})/(\text{oracle}-\text{chance})$, the share of the chance-to-oracle headroom captured. On sub-agent routing (14 AppWorld agent systems), model routing (LLMRouterBench, 12 models), and skill gating (activate a skill or not), \textsc{TabAgent} captures 22--54\% of that headroom. A strong fixed default is hard to beat here, so the question is whether a cheap classifier can match it, and it can: \textsc{TabAgent} matches the best fixed default on all three decisions within a paired task-level bootstrap 95\% CI (App.~\ref{app:ci_protocol}), while making each decision at orders of magnitude lower latency and cost than a generative router. The generative alternative does not do better: few-shot GPT-5.4, at that far higher cost, falls at or below the default on all three decisions. The pattern holds across all three: a generative model is not needed to match the strong default, a cheap discriminative head delivers that choice quality at a fraction of the cost, and the gap to the oracle is shared by every method we test, so it reflects the difficulty of the decision rather than a deficit of the discriminative head. We attribute even this to the same trace-derived state features that help retrieval: reward outcomes correlate with observable task and candidate signals a discriminative head can read.

\begin{figure}[t]
  \centering
  \includegraphics[width=0.64\linewidth]{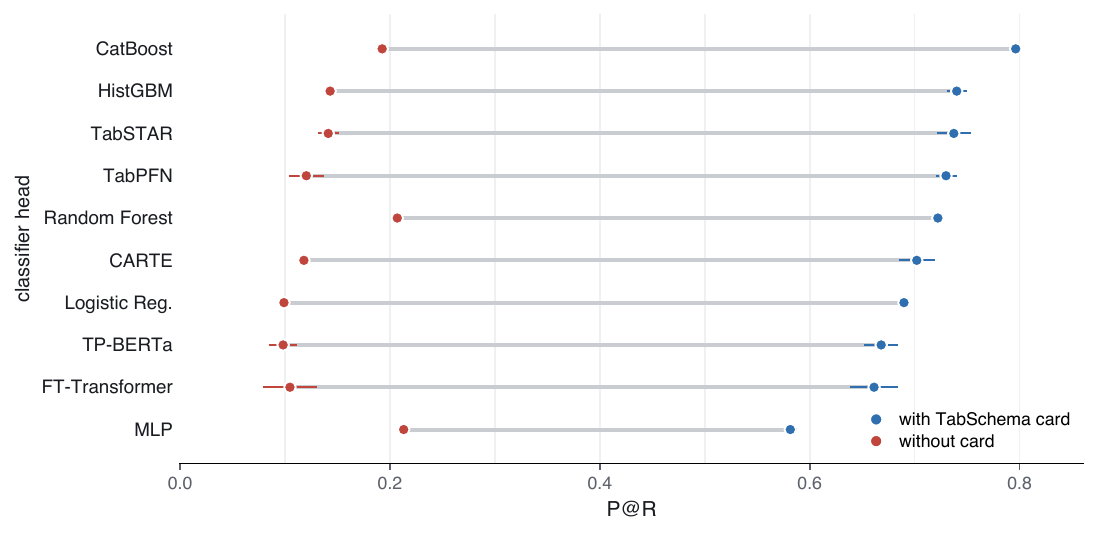}
  \caption{Performance is driven by the TabSchema card, with little dependence on the head. This is a head ablation, not a competition: every head is fed the same TabSchema card, and \textsc{TabHead} (CatBoost) is the head used in all main-paper results. On skill selection, a spectrum of heads (classical trees, linear/MLP baselines, deep tabular networks such as FT-Transformer \citep{FT_Transformer}, and text-native tabular models such as TP-BERTa \citep{TP_BERTa}) all reach useful $\PatR$ with the engineered TabSchema card (blue), but collapse ($\PatR\!\approx\!0.10$--$0.21$) when it is removed (red, raw task/candidate text or concatenated embeddings only). Means of 3 seeds, bars $\pm 1$ s.d.}
  \label{fig:card_new}
\end{figure}

\label{subsec:card_new}
\textbf{The trace-derived feature card carries the decision and the choice of classifier head barely matters, which is why one lightweight head suffices across every decision family.} Whether these gains come from the classifier head or the features it consumes is settled by three ablations. First, with the card held fixed and the head swapped (Fig.~\ref{fig:card_new}), every estimator from classical trees to tabular foundation models reaches useful and comparable $\PatR$ with the TabSchema card but collapses ($\PatR\approx0.10$--$0.21$) without it. Second, holding the head fixed and varying only the trace source, the same head trained on CUGA traces beats one trained on the text-only CodeAct traces by $+24.7\%$ $\PatR$, so it is the trace-derived workflow features, rather than the task text, that carry the signal (App.~\ref{app:tabschema_validation}, Fig.~\ref{fig:slope_new}). Third, holding head and trace source fixed, replacing each tool name at inference with an opaque hash costs at most $0.1$ point of $\PatR$ with the full card but $13.8$ points once the description-derived features are also removed, so the head reads the card rather than the surface name (App.~\ref{app:error_analysis:churn}). KernelSHAP attributions place most of the importance on state and dependency features (App.~\ref{app:tabschema_validation}, Fig.~\ref{fig:shap}). Because the same card serves every decision, one universal \textsc{TabAgent} trained once over the pooled traces of seven decisions from five benchmarks stays within $3$ points of the per-task specialists on five of seven decisions and is slightly ahead on model routing and skill gating, with sizable gaps only where a minority task is under-served by the shared objective (App.~\ref{app:universal}), which shows the reduction learns a shared boundary rather than per-task ones.

\begin{figure}[t!]
  \centering
  \includegraphics[width=0.56\linewidth]{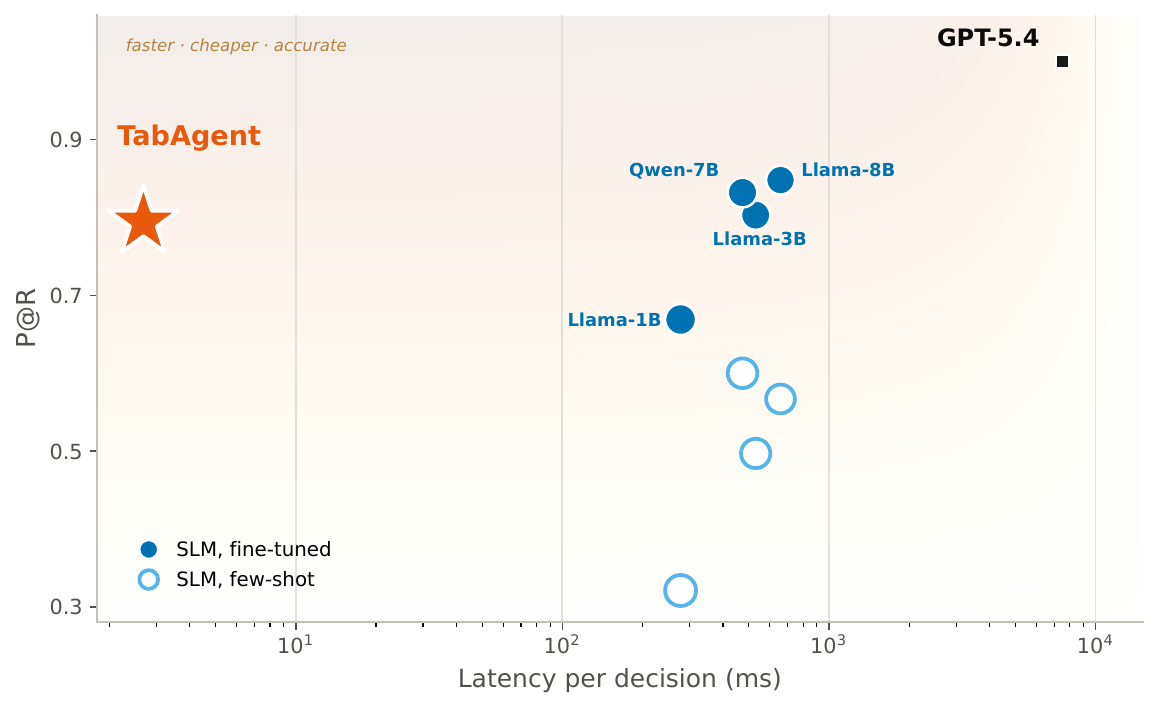}
\caption{Efficiency Pareto over all three cost axes: per-decision latency ($x$, log ms), shortlisting quality $\PatR$ ($y$), and cost-efficiency (marker area, larger $=$ cheaper, spanning $\sim\!10^{5}\times$). SLM latency and cost are measured per-decision on a V100 (${\sim}0.3$--$0.7$\,s, Table~\ref{tab:cost-runtime}). \textsc{TabAgent} (star, top-left) is the fastest and cheapest point yet near the top on quality. Filled circles are fine-tuned SLMs, open circles the same models few-shot. \textsc{TabAgent} dominates every few-shot SLM on all three axes and every SLM on latency and cost, while fine-tuned 3B--8B SLMs and GPT-5.4 reach higher quality only at far greater cost.}
  \label{fig:pareto_new}
\end{figure}

\label{subsec:eff_new}
\textbf{The swap is far cheaper: one classifier pass replaces a multi-second, roughly $21$k-token LLM call at more than $10^{3}\times$ lower latency and $10^{5}\times$ lower cost.} A single \textsc{TabHead} pass runs in $2.682$\,ms at \$$2.0\times10^{-7}$, against $7.50$\,s and \$$0.052$ for the GPT-5.4 shortlister it replaces. Figure~\ref{fig:pareto_new} places this in latency, quality, and cost: \textsc{TabAgent} dominates every few-shot small-LLM point on all three axes and every small LLM on latency and cost, while fine-tuned SLMs from 3B upward and GPT-5.4 reach higher quality only at far higher latency and cost. Scaling or adapting the LLM does not close the cost gap (Fig.~\ref{fig:slm_new}), and the gap is not a hardware artifact: a fine-tuned Llama stays ${\sim}100$--$250\times$ \textsc{TabHead}'s latency and cost on a V100 and $5{,}000$--$27{,}000\times$ on a CPU node, since the tree head gains nothing from a GPU (Table~\ref{tab:cpu-latency}).

Cost has a second axis: the trajectories needed to obtain the head. Subsampling the training pool (App.~\ref{app:cost_efficiency:traj}, Fig.~\ref{fig:traj_efficiency}a) shows \textsc{TabHead} reaching half of its full-pool $\PatR$ from $8$ trajectories and $90\%$ from $N^{\star}{=}118$ on AppWorld, and from $37$--$51$ on the four $\tau^2$-bench agents, so onboarding costs tens to low hundreds of successful traces rather than thousands. The same curve locates where the swap becomes worthwhile, below which the generative alternatives and the confidence gate of App.~\ref{app:error_analysis} carry the traffic.

\label{subsec:inloop}
\textbf{Placed on the agent's critical path in a live, end-to-end run, \textsc{TabAgent} drives AppWorld tasks to completion about as often as the frontier-LLM shortlister it replaces, at a fraction of the per-decision cost.} Every result so far scores the shortlisting decision offline, against the tools a successful trace used, which leaves open the question a drop-in claim turns on: does the swap preserve real task success in a running agent? We answer it with the hot-swap step of the deployment lifecycle (\S\ref{sec:setup}), replacing only CUGA's generative shortlister with \textsc{TabHead} on $24$ held-out cross-application tasks (phone, spotify, and simple\_note), and leaving the rest of the agent, driven by GPT-5.4, untouched in both arms, so any difference in task success isolates the shortlisting decision. Running each task over up to three seeds and scoring Pass@1 on the completed runs, \textsc{TabAgent} reaches $27.1\%$ against $29.2\%$ for the GPT-5.4 shortlister, a paired difference of $-2.1$ points that includes zero (standard error $9.1$, $95\%$ bootstrap CI $[-18.8, +16.7]$, $n{=}24$, Table~\ref{tab:inloop_main}). The replaced call, a roughly $21$k-token generative shortlist, is answered in a single classifier pass at the millisecond-scale per-decision cost established above.

The agreement runs deeper than the matched success rate: the two configurations reach the same pass-or-fail verdict on $79\%$ of tasks (Cohen's $\kappa{=}0.52$), and $82\%$ of \textsc{TabAgent}'s failures are also failures of the LLM shortlister, so these failures sit in the agent and the task rather than in the shortlist, and a better shortlist would not have saved them. We read this as a promising in-loop signal rather than a broad claim, since the study covers one agent and one benchmark on a modest sample. The full breakdown, the confidence-gated fallback measured in the loop, and an analysis of inherited bias appear in App.~\ref{app:inloop}.

\begin{table}[t]
\centering
\small
\caption{\textbf{In-the-loop task success on held-out AppWorld, \textsc{TabAgent} versus the GPT-5.4 shortlister.} Both arms run the same CUGA agent on GPT-5.4 and differ only in the shortlisting module. Task success is Pass@1 averaged over up to three seeds, paired over the $n{=}24$ held-out cross-application tasks both arms completed, with a paired per-task bootstrap $95\%$ confidence interval.}
\label{tab:inloop_main}
\begin{tabular}{lcc}
\toprule
Shortlister & Task success (Pass@1) & Shortlist decision \\
\midrule
GPT-5.4 (generative) & $29.2\%$ & ${\sim}21$k-token LLM call, ${\sim}7.5$\,s \\
\textsc{TabAgent} (\textsc{TabHead}) & $27.1\%$ & one classifier pass, ${\sim}$ms \\
\midrule
Paired difference & \multicolumn{2}{c}{$-2.1$ points, standard error $9.1$, $95\%$ CI $[-18.8, +16.7]$} \\
\bottomrule
\end{tabular}
\end{table}

\label{subsec:scope}
\textbf{\textsc{TabAgent} learns a decision head's behavior from that head's own successful trajectories, so its reach is bounded by what those trajectories contain, and we are explicit about where it should not be expected to work.} It applies to in-distribution closed-set decisions of an agent that already produces a trajectory pool, it does not zero-shot a new domain, its low-confidence decisions can be deferred to the LLM under a confidence gate (AUROC $0.88$), and the learned boundary is tied to one agent architecture rather than transferable across scaffolds. Because the head distills a validated shortlister, it also inherits that shortlister's biases by design, an effect we measure in the loop (App.~\ref{app:inloop:bias}). We therefore do not claim \textsc{TabAgent} beats the LLM everywhere: it is a cheap discriminative head that absorbs in-distribution generative decisions, with the frontier model retained where it still earns its cost. Full scope, limitations, and the broader-impact statement are in App.~\ref{app:scope_limitations}.

\section{Conclusion}

LLM calls are not the only path for agentic system decisions. \textsc{TabAgent} shows that closed-set decisions inside an LLM agent can be cast as structured prediction over execution traces and solved by a cheap tabular classifier that matches frontier-LLM decision boundaries at $99.96\%$ lower latency and $99.9996\%$ lower cost per call (${\sim}10^{3}\times$ and ${\sim}10^{5}\times$). Two ingredients carry the result: trace-derived schema, state, and dependency features (TabSchema), and schema-aligned synthesis that targets coverage where dependencies are rare (TabSynth, ${\sim}23\%$ higher $\PatR$ on AppWorld). The full algorithm applies across seven benchmarks with one fixed head configuration, and one head transfers across LLM backbones that share a scaffold. The broader principle is direct: where a decision space is finite, a discriminative head trained on the agent's own successes can replace its repeated frontier-LLM calls. This shifts agent design from a chain of LLM calls to a composition of specialized small models and frontier LLMs, and opens a route to agents that distill their own decision boundaries from logged successes.

\bibliographystyle{tmlr}
\bibliography{tabagent}

\newpage
\appendix

\section{Scope, Limitations, and Broader Impact}
\label{app:scope_limitations}

\textsc{TabAgent} learns the behavior of a decision head from that head's own successful trajectories, so its reach is bounded by what those trajectories contain. This appendix gives the full account of where the method applies, where it does not, and its broader impact. The main-text summary is \S\ref{subsec:scope}.

\textbf{Domains with no trajectory pool.} \textsc{TabAgent} replaces a head by mimicking the boundary the upstream agent already draws, so with no trajectories for a new domain there is nothing to mimic. A fresh trajectory pool, and the one-time TabSchema and TabSynth onboarding it requires, followed by the ${<}45$\,s CPU TabHead fit, is required before it applies, and it does not zero-shot to a domain the agent has never operated in. How large that pool must be is quantified in \S\ref{subsec:eff_new}. The cold-start floor is a degradation rather than a collapse: an application admitted to the catalog with none of its own traces still reaches $\PatR$ $0.377$, above zero-shot dense retrieval and far above chance, because the trace-derived similarity features transfer even where candidate identities do not, and it recovers as traces accumulate (App.~\ref{app:error_analysis:churn}). New domains that do have trajectory pools, such as OAK and BPO, are handled well, so the limit is the absence of traces, not novelty per se.

\textbf{Different agent architectures.} The cross-agent transfer of Table~\ref{tab:transfer_new} holds within one architecture (four ReAct agents differing only in LLM backbone) and does not imply architecture-agnosticism. TabSchema's features are read from intermediate trace fields whose existence and meaning are shaped by the scaffold, so a different architecture exposes different signals and the learned boundary would not carry over without re-onboarding. This is by construction rather than a measured deficiency: a scaffold that omits, say, agent \emph{thoughts}, or that decomposes the plan across additional intermediate sub-agents, simply exposes a different feature set, so per-architecture extractors are expected and cross-scaffold transfer is out of scope by design.

\textbf{Novel compositions.} On the pooled AppWorld card (App.~\ref{app:appworld_setup_specs}), $23\%$ of held-out tasks need a tool set that never appears whole in training and $2\%$ need a tool never used in training. The pointwise head still ranks the tools of an unseen combination with $\PatR$ $0.63$ (against $0.81$ on the other tasks), and where it is unsure a confidence gate defers the decision to the original LLM: TabHead's confidence flags its own errors at AUROC $0.88$ (App.~\ref{app:error_analysis}), and because \textsc{TabHead} retrains in under $45$\,s on CPU, such cases fold back as trajectories accumulate. Tools that are added or renamed outright, rather than newly combined, are treated separately in App.~\ref{app:error_analysis:churn}.

\textbf{Where synthesis backfires.} TabSynth helps where rare dependencies limit coverage, but it is not universally positive: on Spotify it lowers $\PatR$ ($70\!\to\!61$), which we attribute to a small trajectory pool and strong lexical cues that turn paraphrase diversity into noise rather than coverage, and on $\tau^2$-bench its effect ranges from $-0.9\%$ (airline) to $+1.3\%$ (the combined model). It must also be scoped correctly: on Mind2Web element selection, pooling paraphrases across web domains dilutes the signal, whereas domain-matched synthesis pays off on every domain ($58\!\to\!68$ entertainment, $65\!\to\!74$ shopping, $64\!\to\!72$ travel), as it does on the low-data enterprise cards (OAK $44\!\to\!51$, BPO $34\!\to\!61$). We therefore apply +synth per card and report the head that wins (Table~\ref{tab:master_new}).

\textbf{Bounded ceilings.} On policy/reward the ceiling is set by task difficulty rather than by the router, so the precise claim is that \textsc{TabAgent} captures 22--54\% of the chance-to-oracle headroom and matches the strong default's choice quality at far lower latency and cost, not that it exceeds it.

\textbf{Inherited behavior and representational limits.} Because the head distills a validated shortlister, it inherits that shortlister's decision boundary and its biases by design, an inheritance we measure directly in the loop, where the head agrees with the source on $79\%$ of task outcomes, against the source's own $93\%$ agreement across its reruns, and preserves end-to-end task success (App.~\ref{app:inloop:bias}). Both TabSchema and TabSynth assume a stable trajectory schema and consistent logging from the upstream agent, and because TabSynth only rewords existing rows, it broadens surface coverage but cannot introduce decision patterns absent from the real trajectories. The tabular abstraction excels at structured decision patterns but does not capture complex reasoning that requires novel tool compositions or long-range symbolic dependencies. Overall, we do not claim \textsc{TabAgent} beats the LLM everywhere: it is a cheap discriminative head that absorbs in-distribution generative decisions, with the frontier model retained where it still earns its cost.

\paragraph{Broader Impact Statement.} This paper proposes \textsc{TabAgent}, a framework that replaces repeated LLM-based closed-set decision components in agentic systems (for example tool shortlisting and routing) with a compact discriminative classifier trained on execution-trace signals. The primary positive impact is improved deployability of agentic systems through substantially reduced inference latency and cost, which can broaden access to practical agents and reduce computational and energy overhead in production settings. Potential negative impacts relate mainly to how and where the framework is deployed. By making agentic decision points cheaper and faster, \textsc{TabAgent} could indirectly enable more scalable automation, which may be misused in harmful applications if paired with unsafe downstream tools or policies. Training also relies on logged execution traces, which in real deployments may contain sensitive information such as user content or tool inputs and outputs. To mitigate these risks, deployments should apply standard privacy and security practices for telemetry (data minimization, access controls, retention limits, and redaction) and should couple fast decision heads with appropriate safeguards (permissions, policy enforcement, and auditing) governing what tools can be executed. Overall, we expect the net impact of this work to be positive when it is used to improve efficiency in responsibly governed agentic architectures.

\section{Benchmarks Description}
\label{app:benchmarks}

We evaluate on seven agent benchmarks, described below in the order of the decision families in \S\ref{sec:results_new}: AppWorld, $\tau^2$-bench, OAK, and BPO (tool shortlisting), SkillsBench (skill selection and gating), Mind2Web (operation and GUI element selection), and LLMRouterBench (model routing). Each entry gives the candidate catalog and evaluation protocol, and the trajectory pool where trajectory logging applies. The same TabSchema, TabSynth, TabHead pipeline is used throughout, with TabSynth scoped per card (domain-matched on Mind2Web).

\subsection{AppWorld}
\label{app:appworld_setup_specs}

We evaluate on five applications from AppWorld \citep{trivedi2024appworld}, a benchmark designed to test agents for code generation capabilities. AppWorld implements 9 real day to day apps with 457 APIs in total, about 50 per app on average, and 1{,}470 typed arguments, all documented and executed in a controllable engine. AppWorld defines 750 tasks in total, formed by 250 scenarios with 3 tasks per scenario, with the official split Train 105, Dev 60, Test Normal 168, and Test Challenge 417. Tasks require interactive coding with API calls and provide programmatic evaluation, and the engine supports authentication, cross application effects, and paginated search, implemented in about 26{,}000 lines of API code. Per task statistics indicate multi step composition across apps and tools. On the normal split the averages are 1.5 apps, 8.2 unique APIs, and 42.5 API calls, with maxima of 3 apps, 17 unique APIs, and 244 calls. On the challenge split the averages are 2.0 apps, 10.5 unique APIs, and 46.8 calls, with maxima of 6 apps, 26 unique APIs, and 649 calls.

We focus on five applications, Amazon, Gmail, Phone, SimpleNote, and Spotify, selected by task availability. These behave as complementary sub-benchmarks with heterogeneous action spaces and decision semantics, while also supporting cross-application composition. Each application exposes its own tool catalog with a distinct taxonomy and argument schema, differing in cardinality, type systems, and dependency structure (preconditions, precedence, co-usage). As a result, task distributions differ both lexically and structurally (required-set size $R$, plan length, argument-type entropy, dependency density), and cross-application tasks require composing incompatible schemata under typed constraints. Beyond tool shortlisting, AppWorld also supplies two decisions used in our study: application selection, a multi-label classification over the applications a task touches, and sub-agent routing, a policy decision over $14$ agent systems.

\begin{table}[t!]
\caption{Five AppWorld applications used in our study. AppWorld implements nine apps with four hundred fifty seven APIs in total, about fifty per app on average. The five here are a subset, and the second column gives each one's candidate catalog in our shortlisting card. Tasks often involve dozens of calls. See the per split averages in the text.}
\label{tab:appworld_apps}
\centering
\footnotesize
\begin{tabular}{lccp{0.48\linewidth}}
\toprule
Application & Candidate APIs & Typical horizon & Common dependencies \\
\midrule
Amazon (AZ) & $66$ & about 45 &
search $\rightarrow$ product detail $\rightarrow$ add to cart $\rightarrow$ checkout \\
Gmail (GM)  & $42$ & about 45 &
list threads $\rightarrow$ open message $\rightarrow$ parse fields $\rightarrow$ act \\
Phone (PH)  & $30$ & about 45 &
list alarms $\rightarrow$ match to calendar events $\rightarrow$ disable or update \\
SimpleNote (SN) & $17$ & about 45 &
retrieve or search $\rightarrow$ edit or create $\rightarrow$ save or sync \\
Spotify (SP) & $91$ & about 45 &
login $\rightarrow$ list playlists $\rightarrow$ compute duration $\rightarrow$ play \\
\bottomrule
\end{tabular}
\end{table}

\paragraph{Two evaluation bases.} Table~\ref{tab:master_new} and Table~\ref{tab:perf_new} report AppWorld per application: each decision ranks its own application's catalog (Table~\ref{tab:appworld_apps}), and scores are averaged within the application. The appendix analyses of trajectory efficiency, confidence gating, tool churn, and the small-LLM scaling panels instead use a canonical rebuild of the shortlisting card with $341$ tasks (Amazon $107$, Gmail $132$, Phone $68$, SimpleNote $17$, Spotify $17$) and the real-only head, scored over all tasks together. Its operating point and baselines are therefore on a pooled basis that is not directly comparable to the per-application rows of Table~\ref{tab:master_new}.

\begin{table}[t!]
\caption{Full per-method AppWorld shortlisting breakdown (referenced from Sec.~\ref{sec:results_new}): \textbf{AZ}=Amazon, \textbf{GM}=Gmail, \textbf{PH}=Phone, \textbf{SN}=SimpleNote, \textbf{SP}=Spotify. Real-only trains on features from successful real trajectories. Real+synth adds schema-aligned TabSynth supervision. Llama-1B/3B are in App.~\ref{app:cost_efficiency}. All values are percentages.}
\centering
\footnotesize
\setlength{\tabcolsep}{3.2pt}
\renewcommand{\arraystretch}{1.05}
\resizebox{\textwidth}{!}{%
\begin{tabular}{l|ccccc|ccccc|ccccc}
\toprule
\textbf{Method} & \multicolumn{5}{c|}{\textbf{R-precision (P@R)}} & \multicolumn{5}{c|}{\textbf{Recall@7}} & \multicolumn{5}{c}{\textbf{Recall@9}} \\
 & \textbf{AZ} & \textbf{GM} & \textbf{PH} & \textbf{SN} & \textbf{SP} & \textbf{AZ} & \textbf{GM} & \textbf{PH} & \textbf{SN} & \textbf{SP} & \textbf{AZ} & \textbf{GM} & \textbf{PH} & \textbf{SN} & \textbf{SP} \\
\midrule
BM25 & 28 & 19 & 35 & 42 & 28 & 48 & 51 & 54 & 72 & 60 & 54 & 58 & 57 & 74 & 71 \\
DSR (Zero-Shot) & 39 & 18 & 30 & 55 & 30 & 65 & 46 & 82 & 89 & 62 & 67 & 53 & 88 & 96 & 70 \\
DSR-FT (Real-Only) & 45 & 52 & 57 & 63 & 46 & 89 & 84 & 95 & 99 & 82 & 90 & 87 & 95 & 99 & 85 \\
DSR-FT (Real+synth) & 69 & 56 & 85 & 92 & 64 & 93 & 90 & \textbf{98} & \textbf{100} & 85 & 95 & 93 & 99 & \textbf{100} & 88 \\
Llama-3.1-8B-Instruct (few-shot) & 56 & 45 & 38 & 70 & 67 & 56 & 50 & 65 & 71 & 67 & 56 & 50 & 65 & 73 & 69 \\
TabAgent (Real-Only) & 57 & 58 & 66 & 62 & \textbf{70} & 91 & 88 & \textbf{98} & \textbf{100} & \textbf{88} & 93 & 91 & \textbf{100} & \textbf{100} & 90 \\
TabAgent (Real+synth) & \textbf{71} & \textbf{66} & \textbf{90} & \textbf{96} & 61 & \textbf{95} & \textbf{93} & \textbf{98} & \textbf{100} & \textbf{88} & \textbf{97} & \textbf{95} & \textbf{100} & \textbf{100} & \textbf{92} \\
\bottomrule
\end{tabular}}
\label{tab:perf_new}
\end{table}

\subsubsection{AppWorld Dataset Characteristics and Task Complexity}
\label{app:data_setup_specs}

\paragraph{Dataset Overview}
We analyze a dataset of \textbf{605} successful CUGA (GPT-5.4) trajectories collected on AppWorld. Each row corresponds to one task instance with the associated applications (\texttt{app\_name}), the set of tools actually used (\texttt{tool\_1}--\texttt{tool\_15}), and the number of unique tools used (\texttt{n\_tools}). All entries have \texttt{score}$=1$ (success), consistent with our goal of modeling the decision boundaries of working agent trajectories. Applications represented include the five main apps from the paper, Amazon (AZ), Gmail (GM), Phone (PH), SimpleNote (SN), and Spotify (SP), as well as a small number of File System, Venmo, and Splitwise tasks that we omit.

\paragraph{Task Difficulty Classification}
We classify difficulty per task using the following rules, derived from tool cardinality and app breadth:
\begin{itemize}
  \item \textbf{Easy}: $\leq 3$ tools and a single application.
  \item \textbf{Medium}: $4$--$7$ tools or exactly $2$ applications.
  \item \textbf{Hard}: $\geq 8$ tools or $\geq 3$ applications.
\end{itemize}
In this slice, trajectories operate within a single application, so app-count criteria rarely trigger. Figure~\ref{fig:diff-dist} summarizes the distribution. Table~\ref{tab:difficulty-breakdown} reports counts and shares.

\begin{figure}[h]
  \centering
  \begin{tikzpicture}
    \begin{axis}[
      ybar,
      width=0.7\linewidth,
      height=0.45\linewidth,
      ymin=0,
      ymax=100,
      bar width=18pt,
      symbolic x coords={Easy,Medium,Hard},
      xtick=data,
      ylabel={Share of tasks (\%)},
      nodes near coords,
      nodes near coords align={vertical},
      enlarge x limits=0.2
    ]
      \addplot coordinates {(Easy,79.5) (Medium,18.8) (Hard,1.7)};
    \end{axis}
  \end{tikzpicture}
  \caption{Difficulty-level distribution over all tasks ($N{=}605$).}
  \label{fig:diff-dist}
\end{figure}

\begin{table}[t]
\centering
\footnotesize
\begin{tabular}{lrr}
\toprule
Difficulty & Count & Share (\%) \\
\midrule
Easy   & 481 & 79.5 \\
Medium & 114 & 18.8 \\
Hard   & 10  & 1.7 \\
\bottomrule
\end{tabular}
\caption{Difficulty breakdown by rules above ($N{=}605$).}
\label{tab:difficulty-breakdown}
\end{table}

\paragraph{Tool Usage Statistics}
Overall tools-per-task: mean $\mu=2.61$, median $=2.00$, std.\ $\sigma=1.61$, min/max $=1/15$, quartiles $Q_1=1.00$, $Q_2=2.00$, $Q_3=3.00$, coefficient of variation $\mathrm{CV}=0.62$.
The histogram in Fig.~\ref{fig:tool-hist} shows a right-skewed distribution with most tasks using 1--3 tools.

\begin{figure}[t]
  \centering
  \begin{tikzpicture}
    \begin{axis}[
      ybar,
      width=0.8\linewidth,
      height=0.45\linewidth,
      xlabel={Number of tools per task},
      ylabel={Task count},
      bar width=10pt,
      xmin=0.5, xmax=15.5,
      xtick={1,2,3,4,5,6,7,8,9,10,13,15}
    ]
      \addplot coordinates {
        (1,155) (2,175) (3,151) (4,69) (5,29)
        (6,10) (7,6) (8,5) (9,2) (10,1) (13,1) (15,1)
      };
    \end{axis}
  \end{tikzpicture}
  \caption{Distribution of tools used per task ($N{=}605$).}
  \label{fig:tool-hist}
\end{figure}

\begin{table}[t]
\centering
\footnotesize
\begin{tabular}{lrr}
\toprule
\# Tools & Count & Share (\%) \\
\midrule
1  & 155 & 25.6 \\
2  & 175 & 28.9 \\
3  & 151 & 25.0 \\
4  & 69  & 11.4 \\
5  & 29  & 4.8 \\
6  & 10  & 1.7 \\
7  & 6   & 1.0 \\
8  & 5   & 0.8 \\
9  & 2   & 0.3 \\
10 & 1   & 0.2 \\
13 & 1   & 0.2 \\
15 & 1   & 0.2 \\
\bottomrule
\end{tabular}
\caption{Frequency of tools-per-task ($N{=}605$).}
\label{tab:tool_count_distribution}
\end{table}

\paragraph{Application-Specific Analysis}
Table~\ref{tab:per-app-stats} reports tools-per-task statistics by application (five core apps). Figure~\ref{fig:app-complexity} compares the means.

\begin{table}[t]
\centering
\footnotesize
\begin{tabular}{lrrrrrrr}
\toprule
App & $n$ & $\mu$ & Median & $\sigma$ & Min & Max & CV \\
\midrule
amazon     & 230 & 2.70 & 3 & 1.49 & 1 & 9  & 0.55 \\
gmail      & 190 & 2.89 & 3 & 1.89 & 1 & 15 & 0.66 \\
phone      & 73  & 1.70 & 1 & 0.95 & 1 & 6  & 0.56 \\
simplenote & 17  & 3.35 & 3 & 1.58 & 2 & 8  & 0.47 \\
spotify    & 19  & 2.74 & 3 & 1.10 & 1 & 4  & 0.40 \\
\bottomrule
\end{tabular}
\caption{Per-application summary of tools per task (five core apps, $529$ of the $605$ tasks).}
\label{tab:per-app-stats}
\end{table}

\begin{figure}[t]
  \centering
  \begin{tikzpicture}
    \begin{axis}[
      ybar,
      width=0.75\linewidth,
      height=0.45\linewidth,
      ymin=0,
      xlabel={Application},
      ylabel={Mean tools per task},
      symbolic x coords={amazon,gmail,phone,simplenote,spotify},
      xtick=data,
      nodes near coords,
      nodes near coords align={vertical},
      enlarge x limits=0.15
    ]
      \addplot coordinates {(amazon,2.70) (gmail,2.89) (phone,1.70) (simplenote,3.35) (spotify,2.74)};
    \end{axis}
  \end{tikzpicture}
  \caption{Average tools per task by application ($529$ core-app tasks, $n$ shown in Table~\ref{tab:per-app-stats}).}
  \label{fig:app-complexity}
\end{figure}

\begin{table}[t]
\centering
\footnotesize
\begin{tabular}{lrr}
\toprule
Application & Count & Share (\%) \\
\midrule
amazon     & 230 & 38.0 \\
gmail      & 190 & 31.4 \\
phone      & 73  & 12.1 \\
simplenote & 17  & 2.8  \\
spotify    & 19  & 3.1  \\
\bottomrule
\end{tabular}
\caption{Tasks per application (five core apps, $529$ of $605$ tasks, shares are of all $605$).}
\label{tab:cross-app}
\end{table}

\begin{table}[t]
\centering
\footnotesize
\begin{tabular}{lrrrr}
\toprule
App & Easy & Medium & Hard & Total \\
\midrule
amazon     & 178 (77.4\%) & 49 (21.3\%) & 3 (1.3\%) & 230 \\
gmail      & 139 (73.2\%) & 47 (24.7\%) & 4 (2.1\%) & 190 \\
phone      & 71 (97.3\%)  & 2 (2.7\%)   & 0 (0.0\%) & 73 \\
simplenote & 11 (64.7\%)  & 5 (29.4\%)  & 1 (5.9\%) & 17 \\
spotify    & 14 (73.7\%)  & 5 (26.3\%)  & 0 (0.0\%) & 19 \\
\bottomrule
\end{tabular}
\caption{Application $\times$ difficulty cross-tabulation (counts with row-wise percentages, $529$ core-app tasks).}
\label{tab:app_x_difficulty}
\end{table}

\subsubsection{CUGA Trajectory Example}

We provide a sample from a task execution of CUGA on an AppWorld task, illustrating the multi-agent trace and log format that TabSchema reads. The specific app shown is incidental. In this task, the user's intent is \verb|"Send $250 on venmo to Catherine."|

\begin{tcolorbox}[title=AppWorld CUGA Trajectory Example, colback=gray!3, colframe=gray!40, boxrule=0.4pt, breakable]

\begin{lstlisting}[style=verbatimwrap]

{
"intent": "Send $250 on venmo to Catherine.",
"task_id": "2c544f9_1",
"steps": [
  {
    "name": "TaskAnalyzerAgent",
    "prompts": [
      {
        "role": "system",
        "value": "System: You are an AI assistant
          [...] determine which applications are 
          required to fulfill a user's request. 
          You will be given two inputs:
          1.  `intent`: [...] 
          2.  `available_apps`: A list of JSON objects, 
            where each object represents an application. 
            [...]"
      },
      {
        "role": "generation",
        "value": "{"thoughts":["The user's intent is
          to send $250 to Catherine via Venmo. 
          The 'venmo' app is required to perform 
          the actual payment transaction.
          However, since the recipient is specified 
          by a proper name, [...], the contact info 
          must first be resolved using the 'phone' app 
          before initiating the payment with Venmo.], 
          relevant_apps\":[\"phone\",\"venmo\"]}"
      }
    ]
  },
  {
    "name": "TaskDecompositionAgent",
    "prompts": [
      {
        "role": "system",
        "value": "System: You are an expert 
        in **task decomposition**. Your role is 
        to break down a user's intent into high-level 
        subtasks, [...]
        current datetime: 2023-05-18T12:00:00"
      },
      {
        "role": "generation",
        "value": "{"thoughts":"The user's intent is 
          to send $250 to Catherine via Venmo. [...]" }
      }
    ]
  },
  {
    "name": "PlanControllerAgent",
    "prompts": [
      {
        "role": "system",
        "value": "System: As a plan controller 
          agent, [...] 
          Current datetime: 2023-05-18T12:00:00 [...]
          Let's think step by step."
      },
      {
        "role": "generation",
        "value": "{"thoughts": [...], 
          "next_subtask":"Search for Catherine's 
          contact information (such as Venmo handle 
          or phone number) using the phone app and 
          store the result in a variable for later use.", 
          "next_subtask_app": "phone"}"
      }
    ]
  },
  {
    "name": "ShortlisterAgent",
    "prompts": [
      {
        "role": "system",
        "value": "System: You are an expert AI 
          assistant responsible for selecting 
          relevant APIs to fulfill a user's request.
          Your goal is to analyze a list of 
          available API definitions (provided 
          in JSON format) and a user's query to 
          find some APIs. [...]"
      },
      {
        "role": "generation",
        "value": "{"thoughts":["The user's intent
          is to search for a contact named 'Catherine' 
          and retrieve her contact information, 
          such as phone number or Venmo handle, 
          using the phone app. [...] , "The most 
          direct API for searching contacts by 
          name is 'phone_search_contacts_contacts_get'. 
          It allows searching the contact book 
          using a query parameter (which can be 
          set to 'Catherine'). Its response 
          includes contact_id, first_name, last_name, 
          phone_number, email, and more, making 
          it highly relevant for retrieving 
          Catherine's contact info.[...] ]}"
      }
    ],
    "data": "[...]"
  },
    {
    [...] (additional steps)
    }
]
"score": 1.0
}

\end{lstlisting}

\end{tcolorbox}

As a second illustration, we zoom into the \verb|ShortlisterAgent| decision that \textsc{TabAgent} replaces, here for a core AppWorld application, Amazon. The intent is \verb|"Move all food processors from my amazon cart to wish list."|

\begin{tcolorbox}[title=AppWorld CUGA ShortlisterAgent Decision (Amazon), colback=gray!3, colframe=gray!40, boxrule=0.4pt, breakable]

\begin{lstlisting}[style=verbatimwrap]

{
"intent": "Move all food processors from my amazon
  cart to wish list.",
"task_id": "9bf2c8a_1",
"agent": "ShortlisterAgent",
"score": 1.0,
"messages": [
  {
    "role": "system",
    "value": "System: You are an expert AI assistant
      responsible for selecting relevant APIs to
      fulfill a user's request. [...] Your final
      output must list at least 4 APIs in the
      'result' field, ranked by relevance_score
      from highest to lowest. [...]"
  },
  {
    "role": "generation",
    "value": {
      "thoughts": [
        "The user wants APIs that retrieve all items
          currently in the user's Amazon cart.",
        "The most directly relevant API is
          'amazon_show_cart_cart_get', which returns
          cart_items with product IDs, names,
          quantities and prices and needs no
          parameters.",
        "To reach the minimum of four APIs, I add
          related cart-management APIs usable in the
          workflow: add, update-quantity, and delete
          by product_id."
      ],
      "result": [
        {"name": "amazon_show_cart_cart_get", "relevance_score": 0.95},
        {"name": "amazon_add_product_to_cart_cart_product_id_post", "relevance_score": 0.6},
        {"name": "amazon_update_product_quantity_in_cart_cart_product_id_patch", "relevance_score": 0.6},
        {"name": "amazon_delete_product_from_cart_cart_product_id_delete", "relevance_score": 0.6}
      ]
    }
  }
]
}

\end{lstlisting}

\end{tcolorbox}

\subsection{\texorpdfstring{$\tau^2$}{tau2}-bench}
\label{app:taubench_details}

We evaluate on $\tau^2$-bench \citep{barres2026taubench}, a public multi-turn customer-service benchmark for tool-using agents. It extends the airline and retail domains of the original $\tau$-bench \citep{yao2025tau} with a telecom domain in which the agent and the simulated user both act on a shared environment. We use two public trace sources, each matched to the question it answers.

\textbf{Cross-agent transfer.} The transfer, trajectory-efficiency, and universal-head studies use the official $\tau^2$-bench trajectory release: four agents that share one ReAct scaffold and differ only in the LLM backbone (Claude-3.7-Sonnet, GPT-4.1, GPT-4.1-mini, and o4-mini), run against a GPT-4.1 user simulator. The request text is the customer's first message in the task's first successful run, or, in the release's solo modes for GPT-4.1 and o4-mini, where the agent works from the task ticket without a simulated user, that ticket. Labels come only from runs that the benchmark's own evaluator marks successful. Because $\tau^2$-bench is dual-control, the shortlister ranks the full $67$-tool catalog of agent and user tools (Table~\ref{tab:transfer_new}, Table~\ref{tab:traj_per_tool}).

\textbf{Per-domain shortlisting.} The per-domain rows (Table~\ref{tab:master_new}, Table~\ref{tab:taubench_new}) use a public release of $416$ $\tau^2$-bench trajectories for training tool-use agents\footnote{\url{https://huggingface.co/datasets/Jarrodbarnes/tau2-sft-final}, revision \texttt{b6a74d4}.}, generated mostly by a GPT-5.2 agent against a simulated user. Each domain's candidate catalog is its set of agent tools as called in the official $\tau^2$-bench trajectory release. We keep every trajectory that calls at least one agent tool: $91$ in airline ($14$ tools), $153$ in retail ($15$ tools), and $82$ in telecom ($13$ tools), $326$ in total, with $3.0$, $4.2$, and $3.3$ tools per trajectory. The combined model ranks the union catalog, $38$ distinct tools once the shared \texttt{calculate}, \texttt{transfer\_to\_human\_agents}, and \texttt{get\_user\_details} are merged.

\textbf{Why outcome-agnostic labels.} We keep these trajectories regardless of their final reward, and this is the intended setting rather than a compromise. \textsc{TabAgent} learns the tool choices a deployed agent makes. In deployment every logged trace records those choices, while a task-level success label is often missing or arrives late. This source therefore tests onboarding without an outcome oracle, and it complements the transfer study above, whose labels come only from verified successes. The choice also costs little label quality: on the $18$ tasks with both a successful and a failed run, the two runs' tool sets overlap as much as two successful runs of the same task do (Jaccard $0.72$ versus $0.75$), and a failed run calls $84\%$ of the tools the successful run called. Failures in these traces thus come mostly from arguments and policy steps rather than from which tools to call, and tool choice is the decision a shortlister makes.

\textbf{Card and protocol.} These trajectories expose no agent-internal state such as thoughts or status analyses, so the feature card reduces to request and candidate features: the customer's first message and its word count, the tool's name and documented description, embedding similarity between the request and the tool's name, description and argument names, lexical overlap of the request with the tool name and with its full documentation, a flag for whether the tool belongs to the conversation's domain, and the tool's prior usage, with the similarity centering and the usage prior fitted inside the training folds. The head is the same CatBoost \textsc{TabHead} used throughout, with no per-domain tuning. TabSynth uses the extended operators of App.~\ref{app:TabSynth}, with its configuration chosen inside each training fold by inner cross-validation. We evaluate with $5$-fold cross-validation grouped by $\tau^2$-bench task, so no task contributes to both training and test, report means over the three seeds of \S\ref{sec:metrics}, and every baseline in Table~\ref{tab:master_new} ranks the same candidates on the same folds.

\begin{table}[ht!]
\centering
\footnotesize
\setlength{\tabcolsep}{4pt}
\renewcommand{\arraystretch}{1.05}
\caption{Per-domain $\tau^2$-bench tool shortlisting (referenced from Sec.~\ref{sec:results_new}). Per-domain rows train and evaluate on a single domain, and the final row trains one TabAgent jointly over the merged 38-tool catalog. Scores are for the real-only CatBoost head, means of the three seeds of \S\ref{sec:metrics} under task-grouped $5$-fold cross-validation, and the last column is the relative $\PatR$ change, computed before rounding, when TabSynth, configured by inner cross-validation, is added. All values are percentages.}
\label{tab:taubench_new}
\begin{tabular}{l r r r r}
\toprule
Domain & $\PatR$ & R@7 & R@9 & synth change (\%) \\
\midrule
Airline   & 72 & 96 & 98 & $-0.9$ \\
Retail    & 75 & 92 & 96 & $-0.3$ \\
Telecom   & 85 & 99 & 100 & +0.6 \\
\midrule
All domains & 77 & 94 & 97 & +1.3 \\
\bottomrule
\end{tabular}
\end{table}

\subsection{OAK: Healthcare Customer Care}
\label{app:oak_details}

OAK is an enterprise health-insurance customer-care benchmark composed of 51 long-horizon tasks operating over 21 pharmacy and health-insurance APIs. The API catalog covers prescription management (lookup, refill, transfer), drug interactions (compatibility checks, contraindication warnings), prior authorization (request status, document submission), and coverage verification (formulary checks, benefit summaries). Tasks average 4.7 tools per task, with a maximum of 8 tools for the most complex prior-authorization flows. We use CUGA with GPT-5.4 aligned with the AppWorld setup. CUGA reaches 100\% task success on OAK, so the supervision pool before augmentation is the full benchmark trajectory set. We evaluate TabAgent with 5-fold cross-validation using an 80/20 train/test split within each fold. OAK reaches Recall@9 of 85 with TabAgent. TabSynth lifts $\PatR$ from 44 to 51 (a 16\% relative gain, Table~\ref{tab:master_new}), consistent with the small initial trajectory pool and the high benefit of schema-aligned augmentation in low-data regimes.

\subsection{BPO: Talent Acquisition}
\label{app:bpo_details}

BPO is an enterprise talent-acquisition benchmark composed of 45 long-horizon tasks operating over 32 sourcing-analytics APIs. The API catalog covers SLA analysis (response-time tracking, breach detection), candidate funnel analytics (conversion rates, drop-off attribution), and skill-impact assessment (gap analysis, role-fit scoring). Tasks average 3.0 tools per task. We use CUGA with GPT-5.4 aligned with the AppWorld setup. CUGA reaches 100\% task success on BPO, so the supervision pool before augmentation is the full benchmark trajectory set. We evaluate TabAgent with 5-fold cross-validation using an 80/20 train/test split within each fold. BPO reaches Recall@9 of 97 with TabAgent. TabSynth lifts $\PatR$ from 34 to 61 (a 79\% relative gain, Table~\ref{tab:master_new}), the largest relative gain in the beyond-AppWorld suite, consistent with the small trajectory pool and the well-separated tool semantics that schema-aligned augmentation can amplify.

\subsection{SkillsBench: Skill Selection and Gating}
\label{app:skillsbench_details}
SkillsBench \citep{li2026skillsbench} evaluates how well an agent selects and applies reusable skills across diverse tasks. We use it for two decisions. Skill selection is a set-retrieval decision over a catalog of $195$ candidate skills, scored with $\PatR$ and Recall@$k$. We use it to stress-test candidate-set scaling (Fig.~\ref{fig:overload_new}a). Skill gating is a policy and reward decision, a binary choice of whether to activate a skill, scored with gated reward. Retrieval baselines rank a candidate list and do not apply to this yes/no decision, so its row in Table~\ref{tab:master_new} compares \textsc{TabAgent} with the generative gates and with the best default. We apply the unchanged TabSchema and TabHead pipeline with no per-task tuning, and without TabSynth.

\subsection{Mind2Web: Operation and GUI Element Selection}
\label{app:mind2web_details}
Mind2Web \citep{deng2023mind2web} is a benchmark for generalist web agents over real websites. We use it for two classification decisions. Operation selection is a three-way type choice of the action to take on the target element, scored with macro-F1. GUI element selection picks the target element among the many present on a page, evaluated at the strict $R{=}3$ cutoff across three web domains (entertainment, shopping, and travel). Because candidate elements are visually and lexically similar, we apply domain-matched TabSynth that paraphrases within each web domain.

\subsection{LLMRouterBench: Model Routing}
\label{app:llmrouterbench_details}
LLMRouterBench \citep{li2026llmrouterbench} is a large-scale benchmark for routing a query to the most suitable model. We use its de-leaked split for a policy and reward decision: given a task, route to one of $12$ candidate models. We report route@1 together with the normalized headroom score $\mathrm{Norm}=(\text{method}-\text{chance})/(\text{oracle}-\text{chance})$ used throughout the policy and reward family. Here route@1 is the reward of the chosen arm rather than a top-1 hit, so \emph{chance} is a uniform-random policy's expected reward rather than a $1/K$ rate, which is why chance is $16$ for the 14-system routing and $43$ for the 12-model routing. As elsewhere, the TabAgent pipeline is applied unchanged.

\section{Hardware}
\label{app:hardware}
Each experiment ran on a single node of a shared cluster. Latency and cost were measured on a node with four NVIDIA Tesla V100-SXM2 GPUs, and some small-LLM baseline runs used NVIDIA A100 nodes. Each GPU has 32{,}768\,MiB (32\,GB) of HBM2 memory and a configured power limit of 300\,W. Unless otherwise noted, the models and evaluations comfortably fit on a single V100-32GB. Multiple GPUs were used only for parallel runs or sweeps.

\smallskip\noindent\textbf{Cost estimation.} Per-read local costs were estimated from the measured wall-clock runtime \(t\) on a single V100-32GB as energy plus amortized GPU time:
\(C_{\text{local}}(t)=\big(\tfrac{P}{1000}\big)\big(\tfrac{t}{3600}\big)\mathrm{PUE}\,p_{\mathrm{kWh}}+\big(\tfrac{t}{3600}\big)p_{\mathrm{GPU\mbox{-}h}}\).
Unless noted otherwise we use \(P=250\,\mathrm{W}\), \(\mathrm{PUE}=1.4\), \(p_{\mathrm{kWh}}=\$0.20\), and \(p_{\mathrm{GPU\mbox{-}h}}=\$0.20\). For API-hosted models (e.g., GPT-5.4) we report the metered provider charged.

\smallskip\noindent\textbf{CPU cost.} \textsc{TabHead} runs on CPU with no GPU, so we price it not by the GPU model above but by vCPU-hours at a general-purpose on-demand rate, \(C_{\mathrm{cpu}}=(\text{threads}\cdot t/3600)\,p_{\mathrm{vCPU\mbox{-}h}}\), with \(p_{\mathrm{vCPU\mbox{-}h}}\) the per-vCPU-hour price of an AWS EC2 \texttt{r6i} instance (about \$$0.063$). The applied per-decision figure is derived in App.~\ref{app:cost_efficiency}.

\section{TabSchema}
\label{app:TabSchema}

We run TabSchema (\S\ref{sec:tabschema}) on a set of past trajectories of CUGA on AppWorld, to extract meaningful synthetic features that are used to train TabAgent.

We utilize several prompts in our multi-agent system of TabSchema. All prompts are given as system prompts to the used LLMs. All LLMs used in TabSchema are GPT-5.4 used directly through the API.

\subsection{Engineering decisions and design path}
\label{app:tabschema:engineering}

TabSchema's structure resulted from several engineering decisions during iteration over hundreds of trajectories. We list the most consequential here.

\textbf{Three retries on code execution.} The CodeExecutor retries failed extraction code up to three times, concatenating prior error traces and fix attempts into each subsequent prompt. We selected three after observing that one retry leaves more than 30\% of feature specifications unresolved (the analyzer's first code attempt routinely contains undefined references), and that the marginal recovery beyond three retries is below 2\%. Three retries cover the bulk of recoverable failures without exhausting the LLM call budget.

\textbf{1-to-5 scoring scale per judge.} Each judge returns a graded 1-to-5 relevance rating against an explicit five-level rubric (the judge prompts below) rather than a binary accept/reject vote. The graded scale lets the MetaJudge weight close calls by reading the underlying justifications instead of collapsing distinct rationales into a single vote. In practice, judges concentrate ratings in three regimes (high, mid, low), and the MetaJudge's reconciliation prompt resolves close-call mid ratings from those justifications.

\textbf{Three judges, not one or five.} A single judge tends to inflate one quality dimension (e.g., relevance) at the expense of others (generality, impact). Two judges produce frequent deadlocks. Three judges with distinct quality dimensions converge with a clear majority on most features, and the MetaJudge handles the minority of split cases by reading the three rationales jointly. We tested five judges with finer-grained dimensions during early iteration and found that the additional judges either correlated heavily with existing ones or produced low-information scores.

\textbf{CodeValidator separate from the executor.} The CodeValidator is an LLM-as-a-judge pass that inspects executed feature code for hidden hallucinations (e.g., reference to a trace field that does not exist in the agent's logging schema, even when the code happens to execute on the current trajectory). This validation emerged from observing unreachable and hallucinated features in early iterations, where executable code would silently access fields that existed only by accident in specific trajectories. Separating the validator from the executor lets the same prompt be re-used across agents without modifying the execution sandbox.

\textbf{MetaJudge reconciliation protocol.} The MetaJudge receives the three judge scores, the three accompanying justifications, and the analyzer's original per-feature relevance score and description. It is realized as a structured prompt that issues a single explained decision (accept, conditionally accept, or reject). Compared to a simple score average, the MetaJudge correctly handles cases where one judge is confidently wrong by surfacing the contradicting justification from another judge. The full MetaJudge prompt and several output examples are below.

\subsection{Feature Extraction Prompts}

\begin{tcolorbox}[title=TabSchema FeatureAnalyzer Prompt, colback=gray!3, colframe=gray!40, boxrule=0.4pt, breakable]
Analyze this single multi-agent trajectory and architecture to identify features that are suitable to insert into a classification model to replace certain nodes. \\

TRAJECTORY DATA: \\
\verb|{json.dumps(trajectory, indent=2)}| \\

The schema of the trajectory data is given to you: \\
\verb|{get_json_schema(trajectory)}| \\

Your task is to: \\
1. The classification suitable component is given to you: \verb|{target['name']}| \\
2. For this classification node, extract potential features that could be used to train a replacement model using past steps. \\
3. Consider features from: user inputs, system prompts, context from previous steps, available options, thoughts of previous steps etc. \\
4. Do not try to solve classification with embeddings, classification should only be based on data that can be taken from the trajectory using code. \\

The classification suitable component full step is given to you, but in your analysis, you should only use the data that can be taken from the trajectory before this component: \\

\verb|{json.dumps(target, indent=2)}|

Return your analysis in this JSON format:
\begin{verbatim}
{{
  "potential_features": [
    {{
      "feature_name": "descriptive name",
      "feature_type": "text/numeric/categorical/computed",
      "description": "what this feature represents",
      "extraction_source": 
        "where in the data this comes from",
      "computation": "how to compute this feature",
      "classification_relevance": 
        "how this helps with the classification task",
      "rationale": "why this would be useful",
      "trajectory_file": "{filename}"
    }}
  ]
}}
\end{verbatim}

Focus on features that would actually help a model learn to make the same decisions as the original agents, and are obtained from the trajectory before the classification suitable component.

\end{tcolorbox}

\begin{tcolorbox}[title=TabSchema CodeExtractor Prompt, colback=gray!3, colframe=gray!40, boxrule=0.4pt, breakable]

Based on the classification analysis from a single trajectory, write Python code to extract all identified features from this multi-agent trajectory data. \\

Canonical features extracted from this trajectory: \\
\verb|{state['potential_features']}| \\

The schema of the trajectory data is given to you: \\
\verb|{get_json_schema(state['trajectory']['trajectory'])}| \\

The schema of the potential features is given to you: \\
\verb|{get_json_schema(state['potential_features'])}| \\

Your task is to create a comprehensive feature extraction function that: \\
1. Extract Basic Features: Write code to extract all identified features from the analysis. \\
2. Engineer Computed Features: Implement the computations specified in the feature descriptions. \\
3. Handle Multiple Data Sources: Process features from user inputs, system prompts, context, agent thoughts, etc. \\
4. Create Classification Dataset: Extract relevant features and target labels for this trajectory. \\

Note: The classification suitable component is given to you: \verb|{state['classification_target']}| \\
In your code, you should only use the data that can be taken from the trajectory before the classification suitable component. \\
The classification suitable component is in the step after the trajectory given to you ends. \\

Requirements: \\
- Create a main function \\
    \verb|extract_features_for_classification(trajectory_data)| \\
- Include helper functions for each feature type (text processing, numeric extraction, etc.) \\
- Return features in a structured dictionary named \verb|features|. \\
- Do not use \verb|dict.get()| to access data, instead use \verb|dict['key']|. \\
- If an error occurs, you should throw an exception. \\
- Do not use \verb|exit()| or any command that will crash the program. \\
- Do not include any logging or use of loggers. \\
- Run the function in the global scope after all definitions. \\
- The current trajectory data is available in the global \verb|trajectory| variable. \\

\begin{verbatim}
{generate_retry_prompt(state) if 
    state.get('retry_count', 0) > 0 else ""}
\end{verbatim}

Code Structure Expected:
\begin{verbatim}
import json
from typing import Dict, List, Any, Optional
import re
from collections import Counter

def extract_features_for_classification(
        trajectory_data: Dict) -> Dict[str, Any]:
    """Main feature extraction function"""
    pass

def extract_text_features(text_data: str) -> Dict[str, Any]:
    """Extract features from text fields"""
    pass

def extract_context_features(
        trajectory_data: Dict, 
        current_step: int) -> Dict[str, Any]:
    """Extract features from previous steps and context"""
    pass

def engineer_computed_features(
        base_features: Dict) -> Dict[str, Any]:
    """
    Create engineered features 
    based on analysis suggestions
    """
    pass

def calculate_features(
        trajectory_data: Dict) -> Dict[str, Any]:
    """Calculate features for the classification task"""
    pass

# use flatten_json to flatten every dict 
# that goes into the final csv file
def flatten_json(obj, prefix=''):
    flattened = {}
    for key, value in obj.items():
        if isinstance(value, dict):
            flattened.update(
                flatten_json(value, f"{prefix}{key}_"))
        else:
            flattened[f"{prefix}{key}"] = value
    return flattened

features = extract_features_for_classification(trajectory)
\end{verbatim}

Output format:
\begin{verbatim}
<thoughts>
Your thoughts should only be placed here
</thoughts>
<code>
Only the code you produce, not in any clauses. 
Should be ready to be run.
</code>
\end{verbatim}
\end{tcolorbox}

The function \verb|generate_retry_prompt| fills the code extraction prompt with prior error messages and code generation results if previous attempts raised an error during execution.

\begin{tcolorbox}[title=TabSchema FeatureAggregator Prompt, colback=gray!3, colframe=gray!40, boxrule=0.4pt, breakable]
Your task is to aggregate the features extracted from the multiple trajectories. \\

The features extracted from the multiple trajectories are given to you: \\
\verb|{state['potential_features']}| \\

The schema of the features is given to you: \\
\verb|{get_json_schema(state['potential_features'])}| \\

You should look at the features and decide if they are relevant to the classification task. \\

Remove features that are not relevant to the classification task, or are unobtainable using a trajectory that stops before the classification task. \\

Output Format: \\
\begin{verbatim}
{{
  "filtered_features": [
    {{
      "feature_name": "descriptive name",
      "feature_type": "text/numeric/categorical/computed",
      "description": "what this feature represents",
      "extraction_source": 
        "where in the data this comes from",
      "computation": "how to compute this feature",
      "classification_relevance": 
        "how this helps with the classification task",
      "rationale": "why this would be useful",
    }}
  ],
  "rejected_features": [
    {{
      "feature_name": "descriptive name",
      "feature_type": "text/numeric/categorical/computed",
      "description": "what this feature represents",
      "extraction_source": 
        "where in the data this comes from",
      "computation": "how to compute this feature",
      "classification_relevance": 
        "how this helps with the classification task",
      "rationale": "why this was rejected",
    }}
  ]
}}
\end{verbatim}
\end{tcolorbox}

\subsection{Feature Judgment Prompts}

\begin{tcolorbox}[title=TabSchema Relevance Judge Prompt, colback=gray!3, colframe=gray!40, boxrule=0.4pt, breakable]
\# Instructions \\

You are the \textbf{Relevance Judge} - an expert focused exclusively on evaluating how directly and strongly features relate to the classification task. \\

Your role is to filter features based on their classification relevance and provide scores that will be sent to a meta-judge for final evaluation. \\

The features are given to you: \\
\verb|{state['filtered_features']}| \\

The schema of the features is given to you: \\
\verb|{get_json_schema(state['filtered_features'])}| \\

\# Your Primary Focus \\

Evaluate each feature solely on its \textbf{classification relevance} - how well it can distinguish between classes and contribute to accurate predictions. \\

\# Scoring Criteria (1-5 scale) \\
\begin{itemize}
    \item \textbf{5 (Highly Relevant)}: Feature has clear, strong logical connection to classification target with compelling rationale
    \item \textbf{4 (Very Relevant)}: Feature shows good connection to classification with solid reasoning
    \item \textbf{3 (Moderately Relevant)}: Feature has some relevance but connection could be stronger or clearer
    \item \textbf{2 (Weakly Relevant)}: Feature has minimal connection to classification objective
    \item \textbf{1 (Irrelevant)}: No clear connection to classification task or illogical rationale
\end{itemize}

\# Evaluation Questions \\

\begin{itemize}
    \item Does the feature logically help distinguish between classes?
    \item Is the \verb|classification_relevance| explanation convincing?
    \item Would this feature provide meaningful signal for the classification task?
    \item Does the rationale make sense from a domain perspective?
\end{itemize}

\# Output Format \\
\begin{verbatim}
{
  "judge_type": "relevance_judge",
  "features": [
      {
        "feature_name": [name],
        "description": [description],
        "score": [1-5],
        "confidence": [1-5],
        "assessment": 
          [detailed evaluation of classification relevance],
        "key_factors": ["factor1", "factor2"],
        "extraction_source": 
          [where in the data this comes from],
        "computation": [how to compute this feature],
        "rationale": [why this would be useful],
      }
  ]
}
\end{verbatim}

\end{tcolorbox}

\begin{tcolorbox}[title=TabSchema Generality Judge Prompt, colback=gray!3, colframe=gray!40, boxrule=0.4pt, breakable]
\# Instructions \\

You are the \textbf{Generality Judge} - an expert focused on evaluating feature robustness, consistency, and broad applicability across different data scenarios. \\

Your role is to filter features based on their generalizability and provide scores for meta-judge evaluation. \\

The features are given to you: \\
\verb|{state['filtered_features']}| \\

The schema of the features is given to you: \\
\verb|{get_json_schema(state['filtered_features'])}| \\

\# Your Primary Focus \\

Evaluate each feature's \textbf{generality} - how consistently it can be extracted across diverse trajectories and how robust it is to data variations. \\

\# Scoring Criteria (1-5 scale) \\

\begin{itemize}
    \item \textbf{5 (Highly General)}: Feature can be reliably extracted from virtually all trajectories with consistent meaning
    \item \textbf{4 (Very General)}: Feature available in most trajectories with stable computation
    \item \textbf{3 (Moderately General)}: Feature available in many trajectories but may have some extraction challenges
    \item \textbf{2 (Limited Generality)}: Feature only available in specific trajectory types or contexts
    \item \textbf{1 (Trajectory-Specific)}: Feature only extractable from very specific, limited trajectory patterns
\end{itemize}

\# Evaluation Questions \\

\begin{itemize}
    \item Can this feature be extracted from most/all trajectory types?
    \item Is the extraction source consistently available across different data scenarios?
    \item Would the computation method work reliably across diverse trajectories?
    \item Is the feature definition robust to data quality variations?
    \item Does the feature maintain consistent meaning across different contexts?
\end{itemize}

\# Output Format \\
\begin{verbatim}
{
  "judge_type": "generality_judge",
  "features": [
      {
        "feature_name": [name],
        "description": [description],
        "score": [1-5],
        "confidence": [1-5],
        "assessment": 
          [detailed evaluation of classification relevance],
        "key_factors": ["factor1", "factor2"],
        "extraction_source": 
          [where in the data this comes from],
        "computation": [how to compute this feature],
        "rationale": [why this would be useful],
      }
  ]
}
\end{verbatim}
\end{tcolorbox}

\begin{tcolorbox}[title=TabSchema Impact Judge Prompt, colback=gray!3, colframe=gray!40, boxrule=0.4pt, breakable]

\# Instructions \\

You are the \textbf{Impact and Uniqueness Judge} - an expert focused on evaluating feature innovation, discriminative power, and unique contribution to the feature space. \\

Your role is to filter features based on their potential impact and novelty for meta-judge evaluation. \\

The features are given to you: \\
\verb|{state['filtered_features']}| \\

The schema of the features is given to you: \\
\verb|{get_json_schema(state['filtered_features'])}| \\

\# Your Primary Focus \\

Evaluate each feature's **impact potential** and **uniqueness** - how much new, valuable information it brings and its likely effect on model performance. \\

\# Scoring Criteria (1-5 scale) \\

\begin{itemize}
    \item \textbf{5 (High Impact \& Highly Unique)}: Novel feature with strong discriminative power and unique perspective
    \item \textbf{4 (Good Impact \& Quite Unique)}: Valuable feature with good uniqueness and clear performance benefits
    \item \textbf{3 (Moderate Impact \& Some Uniqueness)}: Decent contribution but may overlap with common features
    \item \textbf{2 (Low Impact \& Limited Uniqueness)}: Minor improvement with significant overlap to standard features
    \item \textbf{1 (Minimal Impact \& Not Unique)}: Redundant or trivial feature with little added value
\end{itemize}

\# Evaluation Questions \\

\begin{itemize}
    \item How much discriminative power does this feature likely have?
    \item Is this feature genuinely novel or just a variation of common features?
    \item What unique perspective or information does this feature capture?
    \item Would this feature likely improve model performance significantly?
    \item Does this feature capture complex patterns that simpler features miss?
\end{itemize}

\# Output Format \\
\begin{verbatim}
{
  "judge_type": "impact_uniqueness_judge",
  "features": [
      {
        "feature_name": [name],
        "description": [description],
        "score": [1-5],
        "confidence": [1-5],
        "assessment": 
          [evaluation of discriminative power 
            and performance potential],
        "key_factors": ["factor1", "factor2"],
        "extraction_source": 
          [where in the data this comes from],
        "computation": [how to compute this feature],
        "rationale": [why this would be useful],
      }
  ]
}
\end{verbatim}

\end{tcolorbox}

\begin{tcolorbox}[title=TabSchema Meta Judge Prompt, colback=gray!3, colframe=gray!40, boxrule=0.4pt, breakable]

\# Instructions \\

You are the \textbf{Meta Judge} responsible for making final feature selection decisions. \\
You will receive evaluations from three specialized judges and must synthesize their perspectives to create a definitive, ranked list of features for the classification task. \\

The three judges you are receiving features from are: \\
1. \textbf{Relevance Judge}, evaluates how directly and strongly features relate to the classification task. \\
2. \textbf{Generality Judge}, evaluates feature robustness, consistency, and broad applicability across different data scenarios. \\
3. \textbf{Impact and Uniqueness Judge}, evaluates feature innovation, discriminative power, and unique contribution to the feature space. \\

\# Your Role \\

As the Meta Judge, you must: \\
\begin{itemize}
    \item \textbf{Reconcile different judge perspectives} when they disagree
    \item \textbf{Balance competing priorities} (relevance vs. generality vs. impact)
    \item \textbf{Make strategic decisions} about feature portfolio composition
    \item \textbf{Provide final authoritative recommendations} with clear rationale
\end{itemize}

\# Input Format \\

You will receive the three judges' feature scorings, with their schemas: \\

Relevance Judge schema: \\
\verb|{get_json_schema(state['judge_relevance_features'])}| \\

Generality Judge schema: \\
\verb|{get_json_schema(state['judge_generality_features'])}| \\

Impact and Uniqueness Judge schema: \\
\verb|{get_json_schema(state['judge_impact_features'])}| \\

\# Data \\

All judges' feature list scores are given to you: \\
\verb|{state['all_judge_features']}| \\

\# Decision Framework \\

You should reconcile as a supervisor between the three judges and output a final decision for all features. Provide a clear rationale for each decision, taking into account the judges' scores equally. \\

\# Output Format \\
\begin{verbatim}
[
  {
    "feature_name": [name],
    "description": [description],
    "final_decision": [accept/conditional/reject],
    "meta_score": [1-5 overall assessment],
    "confidence": [1-5],
    "extraction_source": [where in the data this comes from],
    "computation": [how to compute this feature],
    "rationale": [why this would be useful],
    "decision_rationale": 
      [explanation of final decision 
      considering all judge inputs]
  }
]
\end{verbatim}

\# Strategic Considerations \\

\begin{itemize}
    \item \textbf{Ensure portfolio diversity}: Don't select only one type of feature
    \item \textbf{Balance risk/reward}: Mix proven concepts with innovative features
    \item \textbf{Consider implementation order}: Prioritize features with clear immediate value
    \item \textbf{Account for dependencies}: Consider how features work together
\end{itemize}

Your final decision will determine the feature engineering direction for the classification project. Make thoughtful, well-reasoned decisions that optimize for classification performance while considering practical implementation constraints.

\end{tcolorbox}

\begin{tcolorbox}[title=TabSchema Meta Judge Output Example, colback=gray!3, colframe=gray!40, boxrule=0.4pt, breakable]

\begin{verbatim}
[
  {
    "feature_name": "user_intent",
    "description": "The explicit goal or intent 
      expressed by the user, which drives 
      the API shortlisting process.",
    "final_decision": "accept",
    "meta_score": 5,
    "confidence": 5,
    "extraction_source": "trajectory['intent']",
    "computation": "Directly extract the 'intent' 
      field from the trajectory.",
    "rationale": "User intent is the foundational signal 
      for API shortlisting, directly determining which APIs 
      are relevant by specifying the user's goal.",
    "decision_rationale": "All judges unanimously rated this 
      feature as essential, with maximum scores for relevance, 
      generality, and impact. It is universally present, 
      directly maps to the classification target, and 
      is irreplaceable in the feature set. Its extraction 
      is straightforward and robust. This feature is 
      a must-have for any 
      classification model in this domain."
  },
  {
    "feature_name": "available_api_names",
    "description": "The list of API names available 
      for the current application context.",
    "final_decision": "accept",
    "meta_score": 4,
    "confidence": 4,
    "extraction_source": "Available APIs JSON 
      provided to ShortlisterAgent",
    "computation": "Extract the keys (API names) from 
      the available APIs for the current app.",
    "rationale": "Defines the candidate set for classification, 
      constraining outputs to only available APIs.",
    "decision_rationale": "While its direct discriminative 
      power is limited, all judges agree it is necessary for 
      framing the classification problem and is always present. 
      Its impact is moderate, but it is a required context 
      feature for any model in this domain."
  },
  {
    "feature_name": "available_api_descriptions",
    "description": "Textual descriptions of each 
      available API, detailing their capabilities.",
    "final_decision": "accept",
    "meta_score": 5,
    "confidence": 5,
    "extraction_source": "Available APIs JSON 
      provided to ShortlisterAgent",
    "computation": "Extract the 'description' field for 
      each API in the available APIs JSON.",
    "rationale": "Provides the semantic bridge for matching 
      user intent and task requirements to API capabilities.",
    "decision_rationale": "All judges rate this feature as 
      highly relevant, general, and impactful. It enables 
      nuanced semantic matching and is not easily replaced 
      by simpler features. Its extraction is reliable and 
      it is always present with API names. 
      This feature is critical 
      for high-quality classification."
  },
  {
    "feature_name": "step_id",
    "description": "The index of the ShortlisterAgent 
      step in the trajectory, indicating temporal position.",
    "final_decision": "conditional",
    "meta_score": 3,
    "confidence": 4,
    "extraction_source": "Index of the ShortlisterAgent 
      step in trajectory['steps']",
    "computation": "Find the index of the 
      ShortlisterAgent step in the steps list.",
    "rationale": "Provides temporal context, potentially 
      distinguishing first-time versus 
      repeated shortlisting attempts.",
    "decision_rationale": "Judges agree this feature is 
      always computable and robust, but its direct relevance 
      and impact are low. It may be useful for distinguishing 
      repeated attempts or for temporal analysis, but is not 
      a core feature for API selection. Include only if 
      temporal context is shown to improve performance."
  },
  {
    "feature_name": "previous_agent_thoughts",
    "description": "Reasoning or thoughts from the previous 
      agent step, providing meta-level context and rationale.",
    "final_decision": "accept",
    "meta_score": 4,
    "confidence": 4,
    "extraction_source": "APIPlannerAgent step, 
      last 'generation' prompt value['thoughts']",
    "computation": "Extract the 'thoughts' field from the 
      last 'generation' prompt in the previous step.",
    "rationale": "Provides valuable context, hints for 
      chaining, workflow, or constraints that may improve 
      classification accuracy.",
    "decision_rationale": "Judges rate this feature as 
      quite unique and impactful, though generality is lower 
      due to possible absence or format variability. It adds 
      diversity to the feature portfolio by capturing 
      meta-context and implicit constraints. Include for 
      richer contextual reasoning, 
      especially in complex workflows."
  }
]
\end{verbatim}

\end{tcolorbox}

\section{Significance \& Variance Support}
\label{app:signif-variance}

\subsection{Methodology}
For each application and metric (P@R, Recall@$k$), we compute 95\% confidence intervals (CIs) via a paired bootstrap over tasks. Resamples keep task alignment across methods. Pairwise contrasts use the bootstrap distribution of macro-mean differences $\bar{\Delta} = \frac{1}{|T|}\sum_{t}\big(s_A(t)-s_B(t)\big)$. A contrast is significant when the BCa 95\% CI for $\bar{\Delta}$ excludes $0$. We control family-wise error within each metric (five applications, $m{=}5$) using Holm--Bonferroni. We pre-register the contrast TabAgent(+synth) vs.\ DSR-FT(+synth).

We train with $n{=}3$ seeds. For each task, per-seed scores are averaged to form a task-level estimate before the bootstrap over tasks is applied, preserving the paired structure. Mean$\pm$SE in artifacts use $\mathrm{SE}{=}s/\sqrt{3}$ with $s$ the sample standard deviation across seeds.

Where raw figures report FNR at cutoff $c\in\{R_t,7,9\}$, we convert per task via $\mathrm{Recall}(t;c)=1-\mathrm{FNR}(t;c)$ prior to resampling. Saturated cells (e.g., SimpleNote at $1.000$ for $\mathrm{Recall}@7/9$) compress variance and render fixed-$k$ contrasts non-informative. We mark these with $\ddagger$ and do not interpret them as evidence for superiority.

Cells in performance tables bear a superscript dagger when TabAgent(+synth) significantly exceeds the comparator after Holm correction within metric, and $\ddagger$ marks a ceiling artifact:
\[
\text{marker}=
\begin{cases}
\textsuperscript{\(\dagger\)} &
\begin{aligned}[t]
&\text{BCa 95\% CI for }\bar{\Delta}\text{ excludes }0\\
&\text{at }\alpha{=}0.05\text{ (Holm within metric)}
\end{aligned}\\[0.3em]
\textsuperscript{\(\ddagger\)} &
\begin{aligned}[t]
&\text{both methods at or near ceiling}\\
&\text{(e.g., }1.00\text{)}
\end{aligned}\\[0.3em]
\text{(none)} & \text{otherwise.}
\end{cases}
\]
\paragraph{TabAgent(+synth) vs.\ DSR-FT(+synth).}
\begin{table}[ht!]
\caption{\textbf{Significance grid (per metric Holm correction, $m{=}5$).} Only Gmail $\PatR$ is robustly significant. SimpleNote fixed-$k$ saturates and is flagged $\ddagger$.}
\centering
\footnotesize
\setlength{\tabcolsep}{6pt}
\renewcommand{\arraystretch}{1.1}
\begin{tabular}{lccccc}
\toprule
\textbf{Contrast: TabAgent(+synth) $-$ DSR-FT(+synth)} & \textbf{AZ} & \textbf{GM} & \textbf{PH} & \textbf{SN} & \textbf{SP} \\
\midrule
$\PatR$            &  & \textsuperscript{$\dagger$} &  &  &  \\
$\mathrm{Recall}@9$ &  &  &  & \textsuperscript{$\ddagger$} &  \\
$\mathrm{Recall}@7$ &  &  &  & \textsuperscript{$\ddagger$} &  \\
\bottomrule
\end{tabular}
\label{tab:sig-grid-DSR}
\end{table}

\subsection{CI summaries (means with 95\% CIs) for key contrasts}

\paragraph{$\PatR$: TabAgent(+synth) vs.\ DSR-FT(+synth).}
\begin{table}[ht]
\caption{\textbf{$\PatR$ CIs for TabAgent(+synth) vs.\ DSR-FT(+synth).} Differences align with Table~\ref{tab:sig-grid-DSR}.}
\centering
\footnotesize
\setlength{\tabcolsep}{6pt}
\renewcommand{\arraystretch}{1.1}
\begin{tabular}{lccc}
\toprule
\textbf{App} & \textbf{TabAgent(+synth)} & \textbf{DSR-FT(+synth)} & $\boldsymbol{\Delta}$ \\
\midrule
AZ & 71.2 [67.5,\,75.3] & 69.5 [64.8,\,73.6] & +1.7 \\
GM & 65.9 [60.7,\,71.1] & 55.6 [50.3,\,60.9] & +10.3 \\
PH & 89.9 [84.1,\,95.4] & 85 [77.1,\,92.3] & +4.8 \\
SN & 96.1 [90.2,\,100] & 92.2 [80.4,\,100] & +3.9 \\
SP & 61.1 [44.4,\,77.3] & 64.4 [47.2,\,78.7] & $-$3.2 \\
\bottomrule
\end{tabular}
\label{tab:ci-par-vs-DSR}
\end{table}

\subsection{CI Bounds for Task-Description Ablation}
\label{app:subsec:task-desc-ci}

We report 95\% confidence-interval bounds (lower/upper) for each app and metric to support significance claims. Means for $\PatR$, Recall@7, and Recall@9 appear in the slopegraph of Fig.~\ref{fig:slope_new}.

\begin{table}[ht!]
\caption{95\% CI bounds (lower/upper) for the trace-source ablation (CUGA traces with the TabSchema card vs.\ CodeAct traces with the task description only) on AZ and GM, the two applications shown in the slopegraph (Fig.~\ref{fig:slope_new}). All values are percentages.}
\centering
\footnotesize
\setlength{\tabcolsep}{5pt}
\renewcommand{\arraystretch}{1.05}
\begin{tabular}{llrrrr}
\toprule
\textbf{Metric} & \textbf{App} & \textbf{LB (CUGA)} & \textbf{UB (CUGA)} & \textbf{LB (CodeAct)} & \textbf{UB (CodeAct)} \\
\midrule
$\PatR$       & AZ & 50.8 & 61.2 & 35.3 & 44.8 \\
$\PatR$       & GM & 47.6 & 59 & 42.2 & 52.9 \\
\midrule
Recall@3    & AZ & 59.9 & 69.5 & 44.2 & 54 \\
Recall@3    & GM & 53.3 & 64.1 & 47.9 & 59.3 \\
\midrule
Recall@5    & AZ & 77.5 & 85.3 & 56.2 & 66.4 \\
Recall@5    & GM & 67 & 77.5 & 63 & 74 \\
\midrule
Recall@7    & AZ & 86.5 & 92.9 & 70.3 & 79.7 \\
Recall@7    & GM & 80.7 & 89.4 & 70 & 79.9 \\
\midrule
Recall@9    & AZ & 90.8 & 95.5 & 77.3 & 86 \\
Recall@9    & GM & 84.7 & 92.3 & 73.6 & 83.8 \\
\bottomrule
\end{tabular}
\label{app:tab:task-desc-ci-bounds}
\end{table}

\section{Cost and Efficiency}
\label{app:cost_efficiency}
\begin{table}[ht!]
\caption{Per-decision cost and runtime for one AppWorld shortlisting decision (SLM runtimes are means over $40$ timed decisions). \textsc{TabAgent}'s runtime is on CPU (it needs no GPU) and is device-invariant, the tree head gains no meaningful GPU speedup, so pairing it against the Llama and Qwen SLMs on a single V100-32GB (\S\ref{app:hardware}) already gives the SLMs their best-case hardware. GPT-5.4 uses the metered API charge. Even on that GPU the SLMs are ${\sim}100$--$250\times$ \textsc{TabAgent}'s $2.682$\,ms and per-call cost. Removing the GPU widens the gap to $5{,}000$--$27{,}000\times$ (Table~\ref{tab:cpu-latency}).}
\centering
\footnotesize
\setlength{\tabcolsep}{6pt}
\begin{tabular}{lrr}
\toprule
\textbf{Model} & \textbf{Cost / read (\$)} & \textbf{Runtime / read (s)} \\
\midrule
CUGA's GPT-5.4 (API) shortlister    & 0.052      & 7.50 \\
Llama-3.1-8B-Instruct    & 4.9e$^{-5}$ & 0.658 \\
Llama-3.2-3B-Instruct    & 4.0e$^{-5}$ & 0.531 \\
Qwen2.5-7B-Instruct      & 3.6e$^{-5}$ & 0.474 \\
Llama-3.2-1B-Instruct    & 2.1e$^{-5}$ & 0.278 \\
TabAgent (CatBoost)     & 2.0e$^{-7}$& 0.002682 \\
\bottomrule
\end{tabular}
\label{tab:cost-runtime}
\end{table}

\paragraph{How the \textsc{TabHead} per-decision cost is computed.} \textsc{TabHead} (CatBoost) runs on CPU with no GPU, so its per-decision cost is CPU time priced at an on-demand rate, not the GPU energy or amortization used for the small-LLM baselines. One decision scores the full candidate set (about $48$ candidates for an AppWorld application) in $2.682$\,ms of wall-clock across the head's four worker threads, which is $4\times2.682\,\text{ms}=1.07\times10^{-2}$ vCPU-seconds, or $2.98\times10^{-6}$ vCPU-hours. We price vCPU time at the on-demand rate of a general-purpose AWS EC2 \texttt{r6i} instance (\texttt{r6i.xlarge}, $4$ vCPU at about \$$0.252$ per hour, i.e.\ \$$0.063$ per vCPU-hour), so one decision costs $2.98\times10^{-6}\times\$0.063\approx\$1.9\times10^{-7}$, which we round up to \$$2.0\times10^{-7}$ for conservatism. The figure is device-invariant, since the tree head gains no meaningful GPU speedup, and it carries no GPU energy or amortization term, unlike the small-LLM costs in Table~\ref{tab:cost-runtime}, which are priced from GPU-served runtime, and GPT-5.4, which is the metered API charge.

\begin{figure}[t]
  \centering
  \begin{subfigure}[t]{0.49\linewidth}
    \centering
    \includegraphics[width=\linewidth]{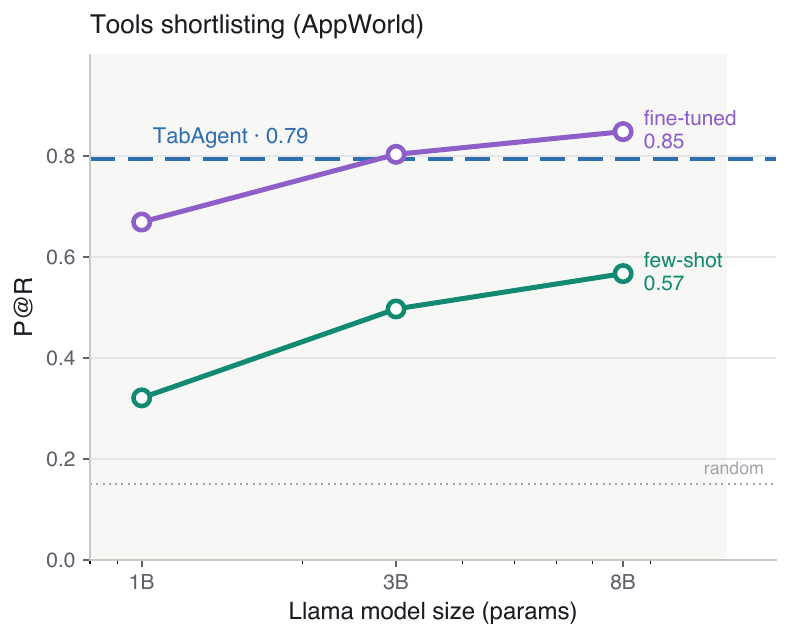}
    \caption{AppWorld tools (17--91 candidates).}
    \label{fig:slm_new_a}
  \end{subfigure}\hfill
  \begin{subfigure}[t]{0.49\linewidth}
    \centering
    \includegraphics[width=\linewidth]{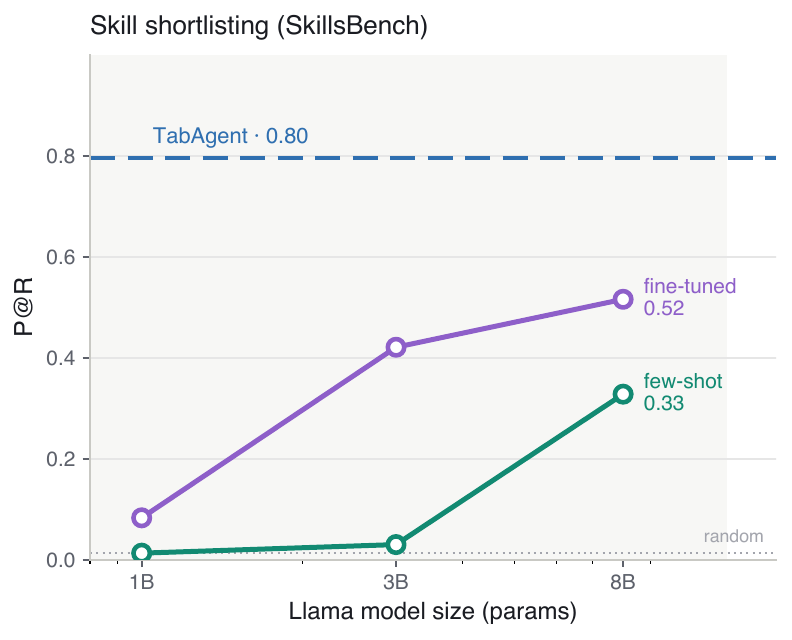}
    \caption{SkillsBench skills (195 candidates).}
    \label{fig:slm_new_b}
  \end{subfigure}
  \caption{Scaling the shortlister LLM does not close the cost gap to TabAgent (dashed line). We evaluate Llama-3.2-1B, Llama-3.2-3B, and Llama-3.1-8B in two adaptation regimes, fine-tuned (purple) and few-shot (green). With a small candidate set (a, AppWorld, 17--91 tools) a fine-tuned SLM matches or exceeds TabAgent only from 3B upward while few-shot stays far below, and with a large candidate set (b, SkillsBench, 195 skills) both regimes plateau well below TabAgent. Even where a fine-tuned SLM matches TabAgent on quality it is far more compute-hungry: ${\sim}100$--$250\times$ higher cost and latency on a GPU ($10^{4}\times$ on CPU, App.~\ref{app:cost_efficiency}).}
  \label{fig:slm_new}
\end{figure}

\begin{table}[ht!]
\caption{CPU-only per-decision latency (no GPU), the same-machine view of Table~\ref{tab:cost-runtime}. Note that Table~\ref{tab:cost-runtime} times a longer prompt on a V100 GPU while this table times a shorter (${\sim}700$-token) prompt on CPU, so the two differ on \emph{both} prompt length and hardware and their SLM latencies are not comparable cell-to-cell across tables. Only \textsc{TabAgent}'s device-invariant $2.682$\,ms is common to both. A decision $=$ shortlist one AppWorld task. The generative SLMs read a compact prompt (task text $+$ the app's candidate API names, ${\sim}700$ tokens) and generate the shortlist (up to a $60$-token budget, means over $n{=}5/5/5/3$ sampled decisions for 1B/3B/Qwen-7B/8B). \textsc{TabAgent}'s $2.682$\,ms is device-invariant: the tree head at ${\sim}48$ candidates/decision is dominated by fixed overhead, not compute, so a GPU gives no meaningful speedup (single-digit ms on either device, marginally slower on GPU from host--device transfer), hence no GPU is needed to reach it. Even without a GPU the generative SLMs are $5{,}000$--$27{,}000\times$ slower. A V100 narrows this to ${\sim}100$--$250\times$ but does not close it (Table~\ref{tab:cost-runtime}). The compact CPU prompt is a conservative lower bound, the full ${\sim}21$k-token shortlisting context is slower still.}
\centering
\footnotesize
\setlength{\tabcolsep}{6pt}
\begin{tabular}{lrr}
\toprule
\textbf{Method (CPU, no GPU)} & \textbf{Latency / decision} & \textbf{vs.\ \textsc{TabAgent}} \\
\midrule
\textsc{TabAgent} (CatBoost, $2.682$\,ms)   & $2.682$\,ms  & $1\times$ \\
Llama-3.2-1B-Instruct (generative)    & $13.9$\,s   & ${\sim}5{,}200\times$ \\
Llama-3.2-3B-Instruct (generative)    & $23.0$\,s   & ${\sim}8{,}600\times$ \\
Qwen2.5-7B-Instruct (generative)      & $43.0$\,s   & ${\sim}16{,}000\times$ \\
Llama-3.1-8B-Instruct (generative)    & $71.2$\,s   & ${\sim}26{,}500\times$ \\
\bottomrule
\end{tabular}
\label{tab:cpu-latency}
\end{table}

\begin{figure}[ht!]
  \centering
  \includegraphics[width=0.72\linewidth]{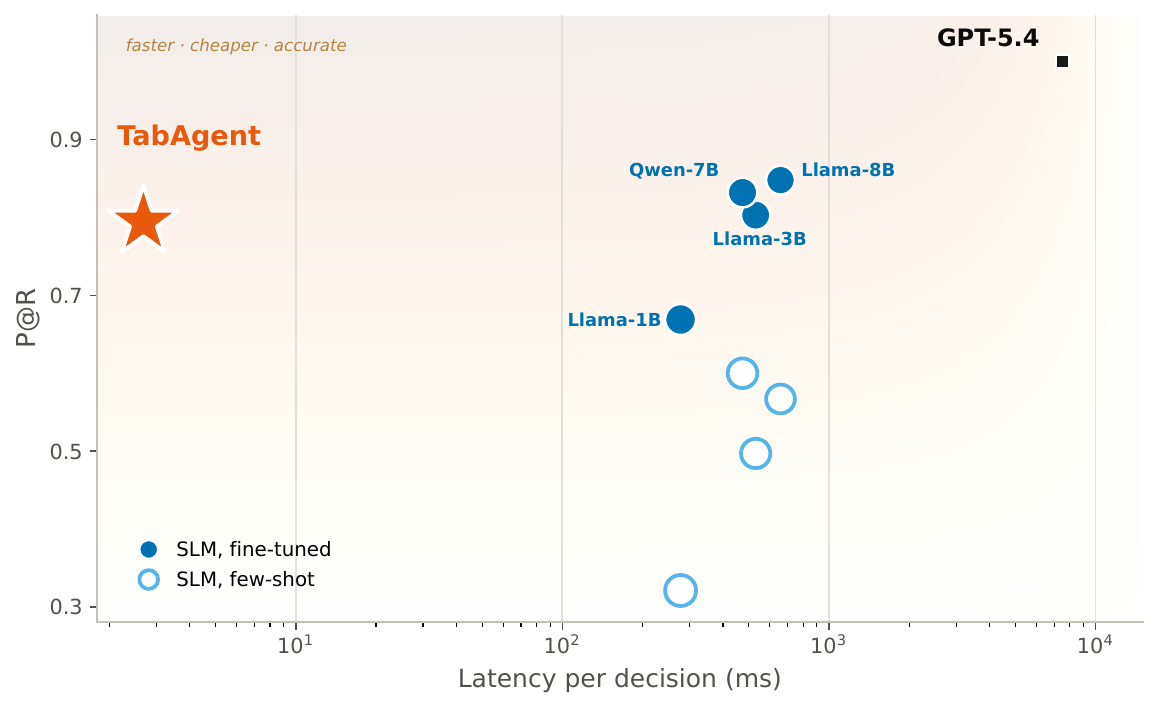}
  \caption{Efficiency Pareto with the generative SLMs on a GPU, at their measured per-decision latency (${\sim}0.3$--$0.7$\,s for Llama-1B/3B/8B and Qwen-7B, Table~\ref{tab:cost-runtime}), their best-case hardware. Identical to the main-text Fig.~\ref{fig:pareto_new}, repeated here beside the no-GPU view (Fig.~\ref{fig:pareto_cpu}). Per-decision latency ($x$, log ms) vs.\ shortlisting quality $\PatR$ ($y$). Marker area $=$ cost-efficiency (larger $=$ cheaper). Even on a GPU, every SLM is ${\sim}100$--$250\times$ slower and costlier per decision than \textsc{TabAgent} (star, $2.682$\,ms, no GPU required). Fine-tuned SLMs from 3B upward edge it on quality only at that cost.}
  \label{fig:pareto_gpu_app}
\end{figure}

\begin{figure}[ht!]
  \centering
  \includegraphics[width=0.72\linewidth]{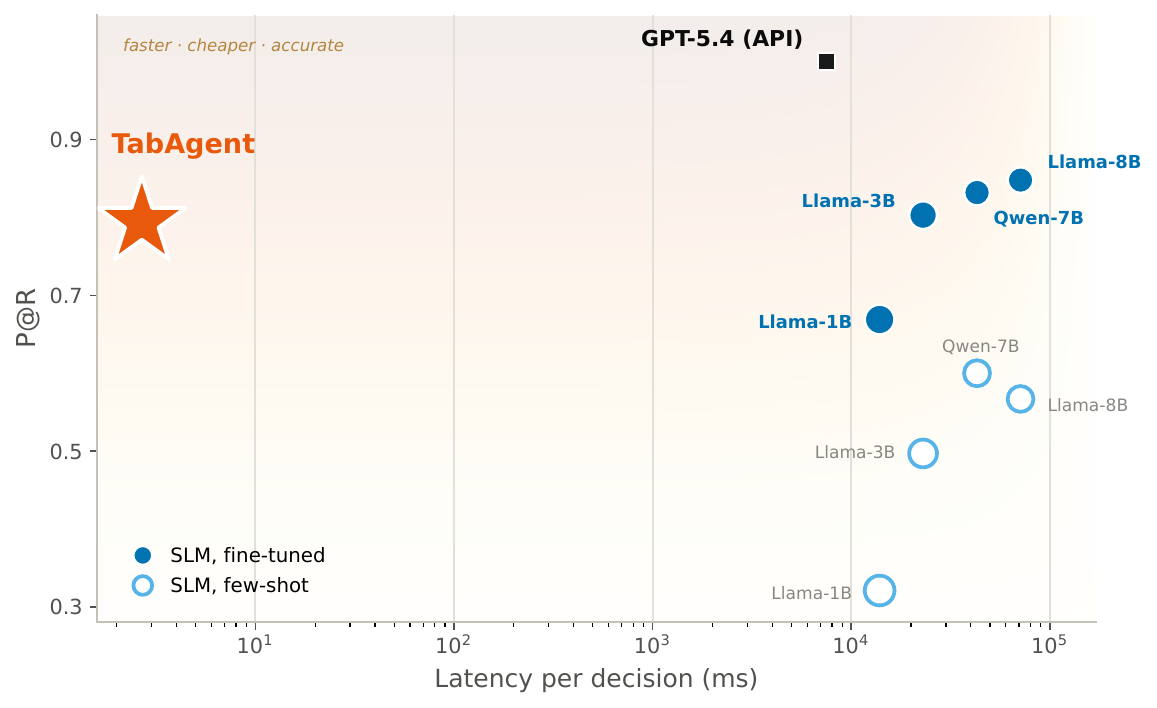}
  \caption{The no-GPU view (companion to Fig.~\ref{fig:pareto_gpu_app}): a single compact per-decision pass (${\sim}700$-token prompt) on one CPU node. The same decision on a V100 takes only ${\sim}0.3$--$0.7$\,s (Fig.~\ref{fig:pareto_gpu_app}). With the GPU removed the generative SLMs (Llama-1B/3B/8B and Qwen-7B, fine-tuned filled, few-shot open) rise to $13.9$--$71.2$\,s per decision, $5{,}000$--$27{,}000\times$ \textsc{TabAgent}'s $2.682$\,ms (star, device-invariant, so a GPU would not move it). This compact prompt is a conservative lower bound, the full ${\sim}21$k-token context is slower. GPT-5.4 is plotted at its API latency for reference.}
  \label{fig:pareto_cpu}
\end{figure}

\subsection{Trajectory efficiency: how much data the head needs}
\label{app:cost_efficiency:traj}
\S\ref{subsec:eff_new} measures what a decision costs at inference. The complementary question is what it costs to obtain the head at all. We subsample $N$ training trajectories per fold on a log grid, leaving the grouped 5-fold test folds untouched, and repeat over ten seeds. The seed controls both which trajectories are retained and the model RNG, and every fold-level prior is recomputed inside the subsample, so a small-$N$ head inherits no full-catalog statistics. The $N{=}\mathrm{all}$ cell is the pooled AppWorld operating point used by these appendix analyses (App.~\ref{app:appworld_setup_specs}), not the per-application numbers of Table~\ref{tab:master_new}.

\begin{figure}[!t]
  \centering
  \includegraphics[width=\linewidth]{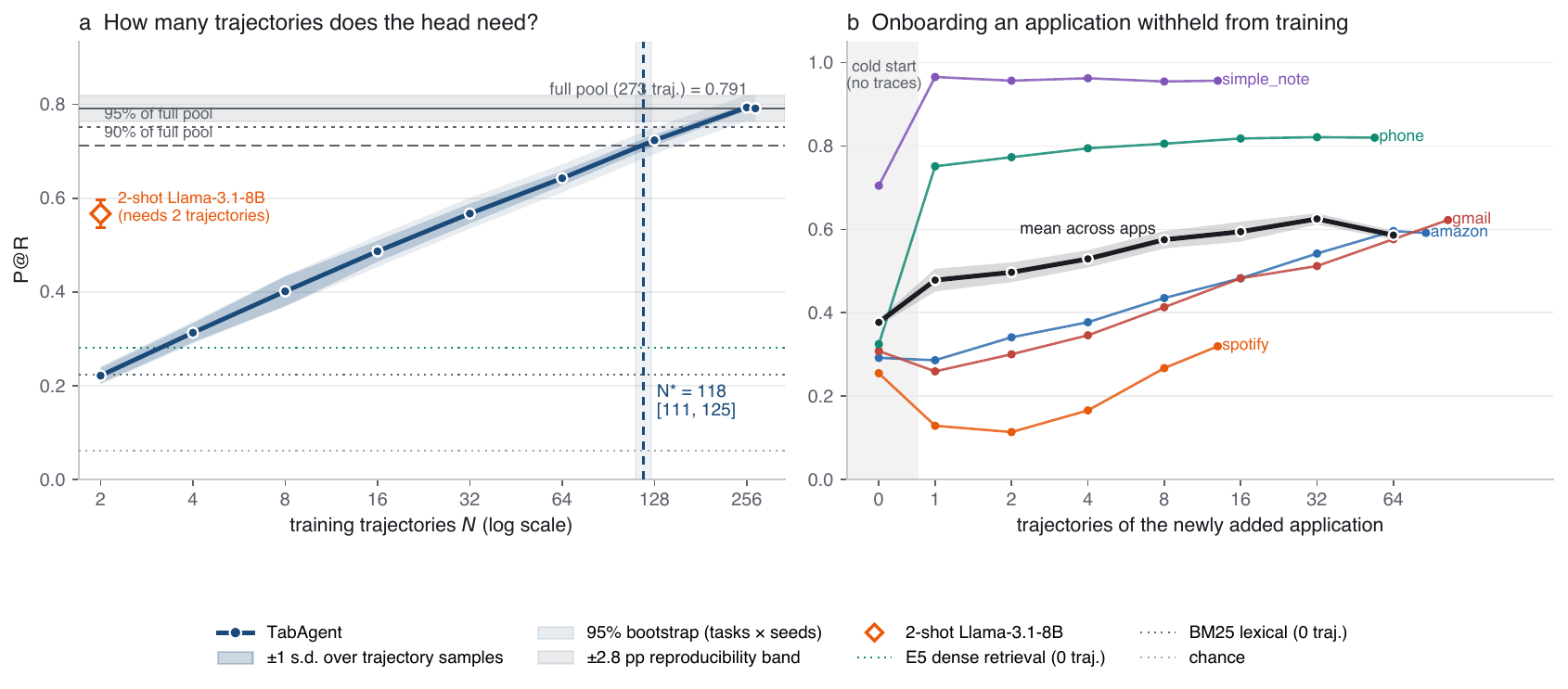}
  \caption{Trajectory efficiency and cold start. (a) AppWorld shortlisting $\PatR$ against the number of training trajectories $N$ (log scale, ten seeds, grouped 5-fold with test folds untouched). The inner band is $\pm1$ s.d. across seeds, where the seed resamples which trajectories are available. The outer band is a nested $95\%$ bootstrap over tasks and seeds, and the grey horizontal band is the $\pm2.8$\,pp cross-node reproducibility floor. Dotted lines are baselines that need no trajectories at all: E5 dense retrieval ($\PatR$ $0.282$), BM25 ($0.224$) and chance ($0.062$). The head starts below retrieval and overtakes it at $N{=}3$. The 2-shot Llama-3.1-8B shortlister ($\PatR$ $0.567$) is drawn at $N{=}2$ because few-shot prompting itself consumes two demonstration trajectories. (b) Onboarding an application withheld entirely from training: the head is trained on the other four applications plus $N$ trajectories of the withheld one, and evaluated only on that application's held-out tasks. $N{=}0$ is therefore a newly added application with no traces of its own, where it reaches $\PatR$ $0.377$ averaged over the five applications. Thin lines are individual applications, direct-labelled. The heavy line is their mean with $\pm1$ s.d. across seeds.}
  \label{fig:traj_efficiency}
\end{figure}

On AppWorld shortlisting $\PatR$ rises from $0.222$ at $N{=}2$ to $0.791$ on the full pool, reaching half of full-pool quality by $N{=}8$ and $90\%$ at $N^{\star}{=}118$ trajectories (bootstrap $95\%$ CI $[111,125]$, Fig.~\ref{fig:traj_efficiency}a). We report the $90\%$ threshold because the curve flattens near its ceiling: its top $3.5\%$ lies inside the $\pm2.8$\,pp cross-node reproducibility band, so thresholds close to the ceiling are poorly resolved. The head overtakes training-free E5 dense retrieval ($\PatR$ $0.282$) at $N{=}3$ and a 2-shot Llama-3.1-8B shortlister ($\PatR$ $0.567$) at $N{=}32$. Few-shot prompting itself consumes two demonstration trajectories, so it is placed at $N{=}2$ rather than treated as a zero-cost baseline. $N^{\star}$ is obtained by fitting a saturating power law $y(N)=y_{\infty}-aN^{-b}$ to the seed means and inverting it, bootstrapping the whole pipeline for the interval. The raw grid crossing agrees. The low-$N$ end is not an artifact of holding one head configuration fixed across the sweep: re-running $N\!\le\!64$ with shallower trees (depth $3$ rather than $6$, ten seeds) changes $\PatR$ by at most $0.8$\,pp anywhere on the curve, inside the reproducibility band, and is slightly worse by $N{=}64$. Nor does it measure which applications happen to be drawn rather than how many trajectories there are: a uniform draw of $8$ trajectories omits Spotify entirely about two thirds of the time, so we repeated the sweep drawing each application in proportion to its share of the pool. Uniform and proportional sampling differ by at most $3$\,pp at every $N$, and by under $1$\,pp for $N\!\ge\!64$.

The requirement is benchmark-dependent rather than a fixed constant (Table~\ref{tab:traj_per_tool}). The four $\tau^2$-bench agents, which rank a single fixed catalog of $67$ tools on every decision, need $37$--$51$ trajectories, whereas AppWorld needs $118$ even though an individual decision there ranks only its own application's endpoints ($48$ on average). The difference is heterogeneity rather than raw catalog size: AppWorld pools five applications whose endpoints, task phrasing and relevant-set sizes differ, so the head must fit five sub-boundaries from one pool. We therefore report $N^{\star}$ per benchmark and do not extrapolate a per-tool constant. The operational summary is tens of trajectories for a single homogeneous catalog and low hundreds for a heterogeneous multi-application one.

\begin{table}[htbp]
\caption{\textbf{Trajectories needed to reach $90\%$ of full-pool quality.} $N^{\star}$ is obtained by inverting a fitted saturating power law (ten seeds, bootstrap $95\%$ CI in brackets). ``Candidates per decision'' is the number of options ranked on a single call, which for AppWorld is the queried application's own endpoint list rather than the $246$-endpoint union across all five applications. The four $\tau^2$-bench agents share one homogeneous $67$-tool catalog. AppWorld pools five heterogeneous applications, which is why it needs more trajectories despite a smaller per-decision candidate set. We report $N^{\star}$ per benchmark and do not extrapolate a per-tool constant.}
\centering
\footnotesize
\setlength{\tabcolsep}{6pt}
\begin{tabular}{lrrr}
\toprule
\textbf{Benchmark} & \textbf{Catalog} & \textbf{Candidates / decision} & \textbf{$N^{\star}$ ($90\%$)} \\
\midrule
AppWorld (5 applications)     & 246 (union) & 48 (mean) & 118 \; [111, 125] \\
$\tau^2$-bench, Claude-3.7    & 67          & 67        & 37 \; [32, 43]    \\
$\tau^2$-bench, GPT-4.1       & 67          & 67        & 45 \; [40, 51]    \\
$\tau^2$-bench, GPT-4.1-mini  & 67          & 67        & 45 \; [38, 53]    \\
$\tau^2$-bench, o4-mini       & 67          & 67        & 51 \; [45, 58]    \\
\bottomrule
\end{tabular}
\label{tab:traj_per_tool}
\end{table}
\FloatBarrier

\section{Universal TabAgent}
\label{app:universal}

A TabAgent head is cheap to fit: we use a fast CatBoost head that trains in under a minute on CPU, so fitting a separate model per decision is inexpensive. The same lightness makes a single shared model attractive for deployment, which raises the question this appendix answers empirically: can one TabAgent, trained once over every task, stand in for all of the per-task heads?

\paragraph{Setup.} The pointwise reduction underlying every row of Table~\ref{tab:master_new} already casts each decision as $(\text{context},\text{candidate})\!\to\!\{0,1\}$ rows carrying a task-group id and a cross-validation fold. We pool these rows from seven decisions over five benchmarks, spanning all three decision families, AppWorld tool shortlisting, $\tau^2$-bench (the four backbones merged into one card), SkillsBench skill selection, Mind2Web operation selection, model routing (LLMRouterBench), sub-agent routing (AppWorld), and skill gating (SkillsBench), into a single training set ($\approx\!145$K rows) and fit one CatBoost head on the union. Training is leakage-safe: for each of the five grouped folds, the universal model is trained on every benchmark's fold-$\neq k$ rows and scores each benchmark's held-out fold-$k$ rows, so no task's evaluation group is ever seen during training, for any task. The feature pipeline is identical for the universal model and a per-task specialist trained under the same pipeline, TF-IDF over the context and candidate text, an e5 context--candidate overlap, a benchmark indicator, and, for operation selection, the target element's tag/type, so the sole difference between the two columns is pooled versus per-task training and the gap isolates the cost of sharing one model. Each benchmark is scored with its native metric (P@R, route@1, macro-F1, or gated reward). We note that this single-task column uses the uniform pooling pipeline rather than the full feature card of Table~\ref{tab:master_new}, so its absolute values differ from the main results. The quantity of interest is the universal$-$specialist delta. Per-card sample balancing was tried and consistently hurt the pooled model, so we keep natural mixing.

\paragraph{Result.} Table~\ref{tab:universal} reports both columns. A single universal TabAgent stays within $3$ points of its per-task specialist on five of seven decisions, $\tau^2$-bench, model routing, sub-agent routing, operation selection, and skill gating, and slightly exceeds the specialist on model routing and skill gating, where sharing appears to regularize a small, noisy task. The remaining cost is concentrated in AppWorld tool shortlisting and skill selection. Within AppWorld it is largest on Spotify ($-38.2$) and Gmail ($-17.7$), with amazon ($-12.3$) also affected, while Phone ($-4.9$) and SimpleNote ($0.0$) are essentially preserved, consistent with a minority task being under-served in the shared objective. Overall, collapsing the specialist heads into one costs little, supporting the view that the pointwise reduction learns a shared, decision-type-agnostic boundary rather than a set of idiosyncratic per-task ones.

\begin{table}[ht!]
\caption{Universal vs.\ specialist TabAgent under an identical pipeline (the only difference is pooled vs.\ per-task training). Metric is each benchmark's native one. Both columns use the uniform pooling pipeline (TF-IDF $+$ e5 overlap $+$ CatBoost), so absolute values differ from Table~\ref{tab:master_new}. The isolated quantity is the universal$-$specialist gap. All values are percentages.}
\label{tab:universal}
\centering
\footnotesize
\setlength{\tabcolsep}{6pt}
\renewcommand{\arraystretch}{1.1}
\begin{tabular}{l l l r r r}
\toprule
Decision (benchmark) & Family & Metric & Specialist & Universal & $\Delta$ \\
\midrule
Model routing (LLMRouterBench) & policy & route@1 & 68.7 & 69.1 & $+0.4$ \\
Skill gating (SkillsBench) & gating & reward & 47.5 & 48.4 & $+0.9$ \\
Sub-agent routing (AppWorld) & policy & route@1 & 39.0 & 37.8 & $-1.2$ \\
Operation selection (Mind2Web) & classification & macro-F1 & 92.7 & 91.2 & $-1.5$ \\
$\tau^2$-bench (4 backbones) & retrieval & P@R & 83.7 & 80.9 & $-2.8$ \\
SkillsBench skill selection & retrieval & P@R & 69.6 & 62.5 & $-7.1$ \\
AppWorld shortlisting (macro) & retrieval & P@R & 79.9 & 66.3 & $-13.6$ \\
\quad\emph{amazon} & & P@R & 76.8 & 64.4 & $-12.3$ \\
\quad\emph{gmail} & & P@R & 75.1 & 57.4 & $-17.7$ \\
\quad\emph{phone} & & P@R & 91.2 & 86.3 & $-4.9$ \\
\quad\emph{simple\_note} & & P@R & 100.0 & 100.0 & $0.0$ \\
\quad\emph{spotify} & & P@R & 71.1 & 32.8 & $-38.2$ \\
\bottomrule
\end{tabular}
\end{table}

\section{TabSynth: Schema-Aligned Paraphrase Augmentation}
\label{app:TabSynth}

TabSynth expands the training set by paraphrasing the free-text state of real trajectory rows while preserving every other field, so each synthetic row restates an observed (context, candidate, label) example in reworded language. It never enumerates the tool catalog, never creates a positive for a tool the trajectory did not use, and never changes a label, so it introduces no new (context, candidate) relationships and cannot manufacture an outcome. Its purpose is coverage: rare but decision-critical patterns appear in more surface forms, which is where the augmentation helps most, on the low-data cards.

\paragraph{Procedure.} The unit of augmentation is a real training row, which is already one (state, candidate, label) example extracted by TabSchema. For each real row, TabSynth produces $k$ variants by paraphrasing only the row's free-text state columns (the agent's thought, status, and progress summaries, and the task text), and copies the candidate identity, the label, and all schema, state, and dependency feature columns verbatim. In the runs reported here $k \in \{2,4\}$ (the $\tau^2$-bench operators below instead allocate variants per trajectory), so the synthetic set is a fixed multiple of the real training set rather than a per-candidate quantity, and the synthetic label distribution matches the real one by construction. The paraphraser is deterministic, rule-based, and seeded, so the augmentation is exactly reproducible, and it rewrites framing and sentence structure while preserving the technical content of the text.

\begin{algorithm}[H]
\caption{TabSynth: schema-aligned paraphrase augmentation}
\label{alg:tabsynth}
\begin{algorithmic}[1]
\Require real training rows $\mathcal{D}$ (each a (state, candidate, label) example), variants per row $k$, free-text columns $\mathcal{T}$
\Ensure augmented training set $\mathcal{D} \cup \mathcal{S}$
\State $\mathcal{S} \gets \emptyset$
\ForAll{row $r \in \mathcal{D}$}
  \For{$j \gets 1$ \textbf{to} $k$}
    \State $r' \gets \textsc{Copy}(r)$ \Comment{candidate, label, and all feature columns carried over unchanged}
    \ForAll{free-text column $c \in \mathcal{T}$}
      \State $r'[c] \gets \textsc{Paraphrase}(r[c], j)$ \Comment{deterministic seeded reword, preserves content}
    \EndFor
    \If{$\textsc{FeatureKey}(r') \notin \textsc{Keys}(\mathcal{D})$}
      \State $\mathcal{S} \gets \mathcal{S} \cup \{r'\}$ \Comment{drop exact duplicates of real rows}
    \EndIf
  \EndFor
\EndFor
\State \Return $\mathcal{D} \cup \mathcal{S}$
\end{algorithmic}
\end{algorithm}

\paragraph{Labels and negatives.} TabSynth is label-preserving: a paraphrase of a positive row stays positive and a paraphrase of a negative row stays negative. It synthesizes no separate negatives, and all negative signal comes from the real (context, candidate) pairs whose candidate was not selected. Because relevance is never reassigned, the augmentation cannot inject a spurious positive for an irrelevant candidate, which is the failure mode a per-candidate generator would risk.

\paragraph{Filtering.} We remove any paraphrased row whose feature key duplicates a real training row, then mix the surviving paraphrases with the real rows during training. We cap the augmentation at $k$ variants per row rather than tuning per stratum, and apply no learned distribution-alignment step, the label-preserving and content-preserving construction keeps the marginals close to the real data by design.

\paragraph{Worked example.} A single Mind2Web element-selection row illustrates the transform. The source instruction is ``Show events in Los Angeles and select the one on the Friday this weekend.'', and one paraphrase variant is ``Please handle the following. Show events in Los Angeles and select the one on the Friday this weekend.''. The reworded framing changes the surface text, while the target element (the candidate) and its label are copied unchanged, so the row still teaches the same decision in a new surface form.

\paragraph{Extended operators ($\tau^2$-bench).} On the $\tau^2$-bench per-domain cards, whose only free text is the customer's request, we add three label-preserving, rule-based operators to the template paraphrases above. First, the text-derived features (request length and the lexical and embedding similarity between request and candidate) are recomputed for each paraphrase instead of copied, so a synthetic row stays internally consistent. Second, an entity swap replaces names, user and order identifiers, reservation codes, e-mail addresses, phone numbers, zip codes, and amounts with other well-formed values before paraphrasing, so the head cannot key on them. Third, a coverage allocation gives each trajectory variants in proportion to $1/\sqrt{f}$, where $f$ is the training frequency of its rarest used tool, keeping the average at four variants and each trajectory between $1$ and $12$. None of these operators changes a label or adds a (context, candidate) pair that no real row contains. Inside each training fold, an inner task-grouped $4$-fold cross-validation chooses among no synthesis, the plain template paraphrases, template paraphrases with recomputed features, entity swap with recomputed features, and entity swap with coverage allocation, and the outer test fold never informs this choice.

\section{Example Input Instance (Feature Card)}
\label{app:features_card_example}

\begin{tcolorbox}[title=Inference-Time Feature Card (Input Only), colback=gray!3, colframe=gray!40, boxrule=0.4pt, breakable]
\begin{tabularx}{\linewidth}{@{}lY@{}}
\toprule
\textbf{Field} & \textbf{Value} \\
\midrule
\textbf{intent} & My roommate asked me to return something I had ordered for them and venmo them the money for it. Check my messages for details and do as per their instruction. \\
\textbf{app\_name} & \texttt{amazon} \\
\textbf{task\_description} & Find APIs that allow searching for orders by product name and initiating a return for a purchased item on Amazon. \\
\textbf{api\_missing} & \texttt{FALSE} \\
\textbf{first\_time\_shortlister} & \texttt{FALSE} \\
\textbf{step\_id} & 24 \\
\textbf{thoughts} & The user's goal is to review roommate messages, extract the item and instructions, then initiate the return on Amazon. Extraction is complete: product ``Vitamix E310 Explorian Blender'' and instructions are stored in \texttt{roommate\_return\_request}. No Amazon shortlisting has happened yet; next is to shortlist order lookup and return-initiation APIs\dots{} \\
\textbf{user\_goal} & Review the messages in \texttt{roommate\_text\_messages} to identify the item to be returned and any specific instructions, then initiate the return on Amazon. \\
\textbf{Overall Status/Analysis} & The latest CoderAgent run correctly extracted the product (``Vitamix E310 Explorian Blender'') and actionable instructions (``return it\dots{} and venmo\dots{}'') from roommate messages; result saved in \texttt{roommate\_return\_request}. A prior attempt mis-extracted these but was corrected. \\
\textbf{Summary of Progress} & First attempt mis-parsed name/instructions; second attempt fixed both, confirming the right product and instruction set in variables\dots{} \\
\textbf{Next Action} & Use \emph{ApiShortlistingAgent} to shortlist Amazon endpoints for order lookup and return initiation \dots{} \\
\textbf{Skepticism/Non-Triviality} & Verify the item is in the user's orders and still within the return window; be explicit about the Venmo reimbursement note/details\dots{} \\
\textbf{Coder Agent Output Analysis} & Most-recent extraction aligns with task needs and variable shapes; previous failure mode (bad item string) is understood and resolved. Prepared to proceed once Amazon APIs are shortlisted. \\
\textbf{candidate tools (labels)} &
\begin{itemize}
  \item \texttt{amazon\_show\_orders\_orders\_get}, List recent orders...
  \item \texttt{amazon\_initiate\_return\_returns\_post}, Start a return for an order...
  \item \texttt{amazon\_show\_returns\_returns\_get}, View existing return requests...
  \item \textit{(additional candidates omitted...)}
\end{itemize} \\

\bottomrule
\end{tabularx}

\vspace{0.25em}
\footnotesize \textbf{What \emph{Next Action} does and does not contain.} This field records the upstream orchestrator's graph-navigation decision, i.e.\ which sub-agent runs next, not which endpoints that sub-agent should return: here it names the \emph{ApiShortlistingAgent} and restates the task intent (``order lookup and return initiation''), while the shortlist itself, the label \textsc{TabHead} must predict, is absent. The card therefore hands the head the planner's framing of the sub-task, which is genuine upstream signal and is the point of TabSchema, but not the answer. The head must still map that framing onto the correct subset of the application's endpoint catalog, which is where the baselines in Table~\ref{tab:master_new} fail. Long field values wrap within the \texttt{tabularx} \texttt{Y} column to avoid margin overflow. We also abbreviate with ``\dots{}'' where appropriate to keep the card compact. This card is strictly a pre-inference input.
\end{tcolorbox}

\section{Additional Results: Application Selection (TaskAnalyzer)}
\label{app:application_selection}

\paragraph{Setup.}
We evaluate the \textbf{TaskAnalyzer} head in TabAgent, which maps task intents to the set of applications required for successful execution. This is framed as a multi-label classification problem: for each task $t$, the ground-truth set $A(t)$ contains all applications observed in canonical trajectories. Predictions $\hat{A}(t)$ are obtained from a classifier with a fixed threshold of $0.5$, with a one-application backoff to the highest-scoring app if $\hat{A}(t) = \emptyset$. Evaluation follows the same 5-fold protocol (by task ID) as in the main experiments, ensuring no leakage of tasks across train and test folds. Generative baselines are evaluated on this decision as on every other, and appear in the \emph{Application selection} row of Table~\ref{tab:master_new}: a few-shot LLM reaches F1 $88$ and a fine-tuned 8B SLM $96$, against \textsc{TabAgent}'s $97$ at a fraction of the cost.

\paragraph{Results.}
Table~\ref{tab:appsel-results} reports instance-level metrics (macro average across folds). We compare (i) TabAgent, (ii) DSR - E5 embeddings used without fine-tuning, and (iii) DSR-FT- E5 embeddings fine-tuned as a binary classifier.

\begin{table}[ht!]
\caption{Application selection (multi-label) instance-level macro metrics (5-fold GroupKFold over tasks) for the \emph{Application selection} decision of Table~\ref{tab:master_new}, whose label-macro F1 is $97$. \textsc{TabAgent} substantially outperforms both frozen and fine-tuned E5 embeddings. Frozen DSR is degenerate at the shared $0.5$ decision threshold, predicting almost every application (recall $100$, precision $18$), so its ranking quality is better read from AUC ($89.5$) than from F1. Fine-tuning improves both the calibration and the ordering (AUC $98.9$). All values are percentages.}
\centering
\footnotesize
\setlength{\tabcolsep}{6pt}
\begin{tabular}{lccccc}
\toprule
\textbf{Method} & \textbf{AUC} & \textbf{Acc} & \textbf{Prec} & \textbf{Rec} & \textbf{F1} \\
\midrule
DSR & 89.54 & 17.99 & 17.99 & \textbf{100.00} & 30.49 \\
DSR-FT & 98.90 & 96.36 & 87.97 & 92.51 & 90.13 \\
TabAgent & \textbf{99.45} & \textbf{98.63} & \textbf{96.67} & 95.68 & \textbf{96.17} \\
\bottomrule
\end{tabular}
\label{tab:appsel-results}
\end{table}

TabAgent achieves the best overall balance, with high precision (96.67) and recall (95.68), leading to the strongest F1 (96.17). The frozen E5 baseline saturates recall ($100.00$) at the shared $0.5$ threshold while its precision collapses to $17.99$, giving F1 $30.49$. Its AUC of $89.54$ shows that the ordering it induces is informative and that only the calibration fails, so we do not read this gap as a ranking result. Fine-tuning E5 repairs that calibration (F1 $90.13$), but it still trails TabAgent on both precision and recall.

Although the task involves only nine candidate applications, making it relatively tractable, these strong results demonstrate that even with limited label space, simple discriminative heads can outperform embedding-based baselines. The task itself provides very little agentic information beyond the raw intent text. To enhance performance, TabAgent incorporates lightweight features such as: \texttt{intent\_token\_len} (number of whitespace tokens), \texttt{intent\_is\_question} (presence of a question mark), \texttt{intent\_digit\_ratio} (fraction of digits in text), and \texttt{intent\_ent\_ORG} (number of organization entities). These low-cost features capture surface-level and structural signals that embeddings alone fail to exploit.

We note that application selection is not the core challenge in agent pipelines, and in some cases a simple classifier may be sufficient. However, the broader implication is that transitioning from a large generative model to a small, fast, and accurate discriminative model is invaluable when performance is preserved. This experiment shows that TabAgent enables such a transition, offering efficiency without sacrificing quality.

\section{Validating TabSchema Feature Extraction}
\label{app:tabschema_validation}
We validate that TabSchema surfaces decision--relevant signals for shortlisting via two complementary lenses: (i) an expert annotation study assessing usefulness and agreement on feature importance, and (ii) a model--based attribution analysis using SHAP values on held-out folds of the trained TabHead classifier.

\begin{figure}[ht!]
  \centering
    \includegraphics[width=0.6\linewidth]{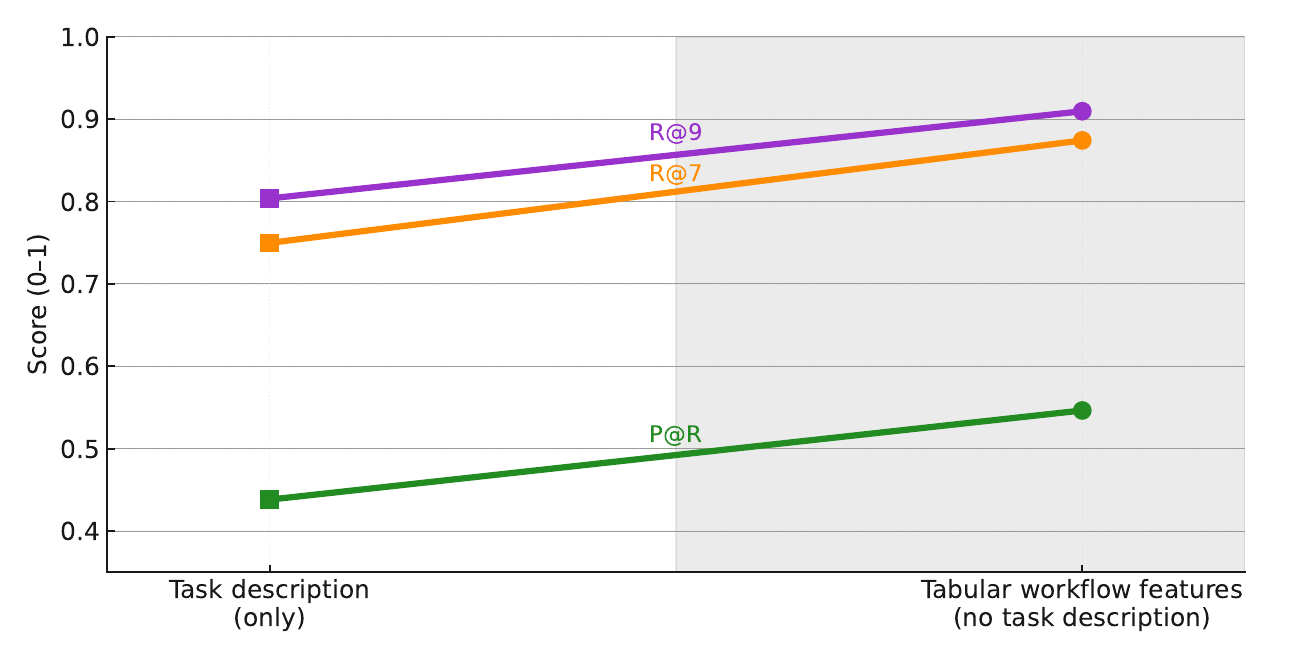}
    \caption{Trace-source ablation (referenced from Sec.~\ref{sec:results_new}): the same TabHead trained on CUGA traces (full TabSchema workflow features) vs.\ on CodeAct traces (task description only), on AZ/GM. Slopegraph for $\PatR$, Recall@7, and Recall@9. Lines rise left$\to$right, indicating that trace-derived workflow features dominate task-description-only input.}
    \label{fig:slope_new}
\end{figure}

\subsection{Expert Annotation}
\textbf{Design.} Three annotators (2--4 years of agentic--systems experience) independently analyzed a stratified sample of CUGA/AppWorld trajectories (five apps) and their TabSchema feature cards. For each app they (a) ranked 15--25 features by shortlisting importance (ties allowed), (b) rated usefulness on a 1--5 Likert scale, and (c) flagged features as previously overlooked if they would not have listed them a priori. Annotations were performed independently without discussion.

\textbf{Results.} Rank agreement within apps is high: Kendall's $W{=}0.76$ (95\% CI 0.68--0.83, $p{<}10^{-4}$ via 10k permutations). Usefulness labels (binarized at $\ge 4$) show substantial agreement: Fleiss' $\kappa{=}0.81$ (95\% CI 0.74--0.87). Median usefulness is $5$ (IQR $4$--$5$), with $86\%$ of features rated $\ge 4$. Several execution-state/dependency features were independently marked previously overlooked by multiple annotators, indicating recovery of valuable, non-obvious cues.

\textbf{Limitations.} The study covers a modest number of trajectories and five apps. Conclusions reflect current stacks and may evolve with new domains. We treat these results as relevance validation rather than a code-audit--complete specification.

\subsection{Shapley Value Analysis}
We quantify each feature's predictive contribution to TabHead using SHAP, computed strictly on held-out folds to avoid leakage \citep{lundberg2017unified}.

\paragraph{Computation details.}
We use the CatBoost TabHead trained under the 5-fold, task-grouped protocol of \S\ref{sec:metrics}. For each fold, we fit KernelSHAP on the test split with:
(i) a background set of 1{,}000 samples stratified by application and deciles of $|G(t)|$,
(ii) 200 SHAP evaluations per test task,
(iii) perturbations constrained to observed support (categoricals resampled from empirical marginals, bounded numerics clipped to train-fold ranges), and
(iv) fixed seeds across folds for reproducibility. We report mean absolute SHAP values, normalized to percentage contribution, macro-averaged across tasks and folds. %

\begin{figure}[t]
  \centering
  \includegraphics[width=\linewidth]{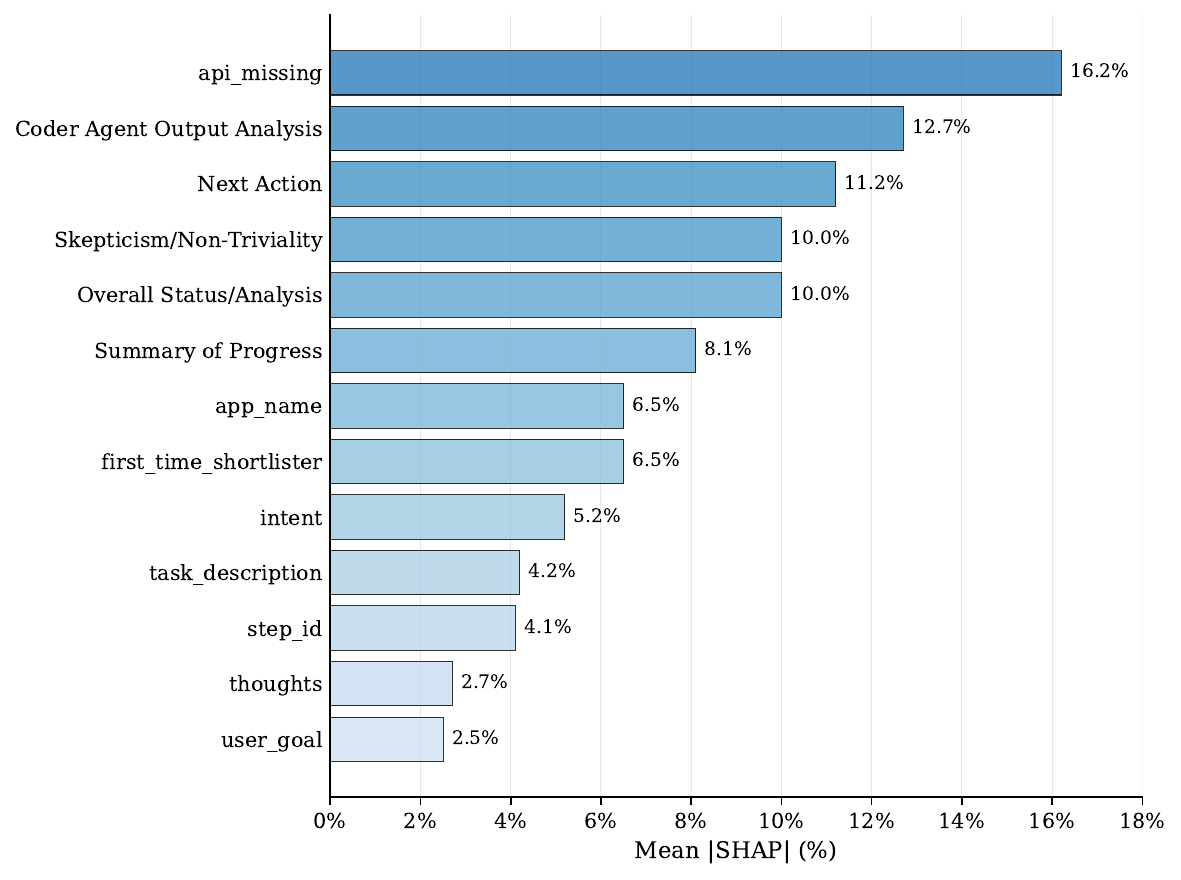}
  \caption{Mean $|\mathrm{SHAP}|$ (\%) for the TabHead shortlister on held-out tasks (5-fold CV). }
  \label{fig:shap}
\end{figure}

\paragraph{Findings.}
Figure~\ref{fig:shap} shows behavior-grounded, trajectory-derived features carry most of the predictive mass: state/dependency indicators (\texttt{api\_missing}, \texttt{Next Action}, \texttt{Overall Status/Analysis}, \texttt{Summary of Progress}, \texttt{Skepticism/Non-Triviality}, \texttt{Coder Agent Output Analysis}) lead, while schema/context signals (e.g., \texttt{app\_name}, \texttt{first\_time\_shortlister}) calibrate the decision. \texttt{intent} and \texttt{task\_description} remain helpful but rank below the collective workflow features, consistent with our ablation that tabular workflow features alone outperform descriptions alone (Fig.~\ref{fig:slope_new}).

\paragraph{Takeaway.}
Expert judgments and held-out SHAP attributions align: the same TabSchema features experts deem useful dominate model decisions. This supports our claim that execution traces expose schema/state/dependency signals that can be tabularized and learned by a compact discriminative head to reproduce generative shortlister behavior with high completeness ($P@R$) at a fraction of latency and cost.

\subsection{Alternative Neural Head (TabSTAR)}
\label{app:head_alt}
The primary TabHead reported throughout the paper is CatBoost. As one of the heads in the head ablation (Fig.~\ref{fig:card_new}), we additionally evaluate a neural textual--tabular head based on TabSTAR \citep{arazi2025tabstar}, fine-tuned on the same TabSchema feature cards with LoRA ($lr=0.001$, $lora\_r=16$, $max\_epochs=50$). Because the ablation shows performance is driven by the TabSchema card and is essentially invariant to the choice of head, we report CatBoost throughout for its fast CPU training (under a minute) and millisecond inference, and use the neural head only to demonstrate this head-invariance.

\section{Evaluation Metrics}
\label{app:evaluation}

\paragraph{Tied scores.} When a method gives several candidates the same score, we order the tied candidates uniformly at random and report the exact expected metric, so no method gains from the order in which candidates are listed. This matters mainly for BM25, which gives many candidates a score of zero.

\paragraph{Single-class classification.} For tasks with a single ground-truth $y_t \in \mathcal{Y}$ and a predicted label $\hat{y}_t$:
\[
\mathrm{Acc} \;=\; \frac{1}{|T|}\sum_{t\in T}\mathbf{1}\{\hat{y}_t = y_t\}.
\]

\paragraph{Multi-label classification.}
For tasks with a ground-truth relevant set $Y_t \subseteq \mathcal{A}$ and a prediction $\hat{Y}_t$:
\[
\begin{aligned}
\mathrm{Prec}
&= \frac{1}{|T|}\sum_{t\in T}
\frac{|\,\hat{Y}_t \cap Y_t\,|}{\max(1,|\hat{Y}_t|)}
\\
&\text{and}
\\
\mathrm{Rec}
&= \frac{1}{|T|}\sum_{t\in T}
\frac{|\,\hat{Y}_t \cap Y_t\,|}{\max(1,|Y_t|)}.
\end{aligned}
\]
For probabilistic outputs, decisions are formed by thresholding per-label scores at a fixed $0.5$ (App.~\ref{app:application_selection}).

\section{Statistical Protocol and Confidence Intervals}
\label{app:ci_protocol}
All decision-quality scores use grouped 5-fold cross-validation at the task level, with the decision group (a task or step) as the resampling unit, repeated across seeds $20260720$, $20260726$, and $20260727$. TabSynth rows are held within fold to avoid leakage. For any A-versus-B comparison we report a paired task-level bootstrap 95\% confidence interval over $2{,}000$ resamples of the difference, and we call a difference significant only when the interval excludes zero.

Table~\ref{tab:ci_protocol} applies this to the most attackable claims. Skill-selection quality is interval-backed: the $\PatR$ interval's lower bound clears the overlap-only retrieval baseline. The policy and reward margins, in contrast, are small and do not establish a win over the strong fixed default. Sub-agent routing sits about two points above the best-fixed default, the de-leaked model router about half a point above the best fixed model ($0.705$ vs $0.701$ route@1, three seeds), and skill gating about one point below always-activate, and all three intervals include zero. We therefore describe \textsc{TabAgent} on routing and gating as competitive with the strong default at far lower cost (\S\ref{sec:results_new}), not beating it.

\begin{table}[ht!]
\centering
\small
\caption{Paired task-level bootstrap 95\% confidence intervals ($2{,}000$ resamples). The routing and gating margins over the best fixed default all include zero (parity), and the skill-selection interval's lower bound clears the overlap-only baseline.}
\label{tab:ci_protocol}
\begin{tabular}{lcc}
\toprule
Comparison & Estimate & Paired 95\% CI \\
\midrule
Skill selection $\PatR$ (absolute) & $0.796$ & $[0.740,\ 0.854]$ \\
Skill selection vs.\ BM25 & $+14.7$ pp & $[7.8,\ 22.3]$ pp \\
Skill selection vs.\ fine-tuned dense retrieval & $+5.1$ pp & $[-1.9,\ 12.9]$ pp \\
Sub-agent routing vs.\ best-fixed & $+2.3$ pp & $[-0.5,\ +5.1]$ pp \\
Model routing vs.\ best-fixed & $+0.4$ pp & $[-3.5,\ +4.6]$ pp \\
Skill gating vs.\ always-activate & $-1.3$ pp & $[-3.8,\ +0.7]$ pp \\
\bottomrule
\end{tabular}
\end{table}

\section{Error Analysis: Tool Shortlisting Discrepancies}
\label{app:error_analysis}

To understand where TabAgent's shortlist decisions diverge from ground truth extracted from the CUGA's trajectories and assess potential downstream impact, we conducted a detailed error analysis on a stratified sample of tasks where TabAgent's predictions differed from the reference labels. This analysis focuses on $P@R$ performance, where the adaptive cutoff (set to the true relevant-set size $R_t$) provides the strictest test of ranking fidelity and leaves no slack for recovery via larger $k$.

\subsection{Sampling and Coding Protocol}

We sampled 40 tasks across the five AppWorld applications (Amazon, Gmail, Phone, SimpleNote, Spotify) where TabAgent(+synth) produced shortlists that differed from the ground-truth tool sets $G(t)$ derived from successful CUGA trajectories. The sampling was stratified by application and by the magnitude of disagreement. For each discrepancy, we examined the task description, the ground-truth tool set, TabAgent's predicted shortlist, and the tool catalog documentation to categorize the nature of the error.

We note an important methodological constraint: AppWorld does not provide oracle ground-truth tool labels. Instead, our reference labels $G(t)$ are the tools actually invoked in successful CUGA executions. This construction means that some tools marked as ground truth may represent one valid solution path among several possible approaches, rather than a strict requirement. Conversely, tools omitted by TabAgent might still enable task completion through alternative pathways not explored in the canonical CUGA trajectory. This ambiguity is inherent to evaluating tool shortlisting in open-ended agent benchmarks and should be considered when interpreting the error categories below.

\subsection{Error Categories and Frequency}

Our analysis identified three primary categories of discrepancies, presented here in descending order of frequency.

\paragraph{Category 1: Optional or nice-to-have tools (62\% of errors).}
The majority of discrepancies arise when the ground-truth set includes a tool that TabAgent's shortlist omits (ranked lower than top $R_t$), yet alternative tools in TabAgent's ranking can accomplish the same functional goal. These cases reflect redundancy or partial overlap in the tool catalog rather than genuine shortlisting failures. For instance, in a Gmail task requiring message search (``Find emails from my manager about the Q3 report''), the ground truth might include \texttt{gmail\_show\_inbox\_threads} because the canonical CUGA trajectory searched received messages, while TabAgent ranked \texttt{gmail\_show\_outbox\_threads} combined with \texttt{gmail\_search\_users} higher, which can locate the same correspondence thread via sender lookup. Both approaches retrieve the target messages, and downstream task success is preserved regardless of which search strategy is shortlisted. Similar patterns appeared in Amazon tasks where \texttt{amazon\_show\_product\_reviews} was favored by ground truth but \texttt{amazon\_show\_product} provided sufficient review summary data in its response payload, eliminating the need for the dedicated reviews endpoint. In Phone tasks, \texttt{phone\_show\_alarms} (list all alarms) and \texttt{phone\_show\_alarm} (show specific alarm) exhibited overlap when the task required inspecting a particular alarm and the alarm identifier was available from prior context.

These optional-tool errors are characterized by two properties: first, TabAgent's shortlist contains at least one tool that subsumes or closely approximates the functionality of the omitted ground-truth tool, second, execution of the task using TabAgent's shortlist would likely succeed without modification to the plan. The high frequency of this category suggests that AppWorld's tool catalog contains multiple valid pathways to task completion, and that the distinction between correct and incorrect shortlists is often one of preference rather than correctness.

\paragraph{Category 2: Semantically similar endpoint confusion (31\% of errors).}
The second category captures cases where TabAgent includes the functionally necessary tool but ranks a semantically similar alternative higher, causing a discrepancy when evaluated at the strict $R_t$ cutoff. For example, in a SimpleNote task requiring content modification (``Update my grocery list note with milk and eggs''), the ground truth used \texttt{simple\_note\_update\_note}, which performs full content replacement, while TabAgent ranked \texttt{simple\_note\_add\_content\_to\_note} higher, which appends or prepends content. The ambiguity arises because ``update'' can mean either replace existing content or augment it, and both operations successfully modify the note to include the required items. Similarly, in Spotify tasks, \texttt{spotify\_show\_playlist} (retrieve specific playlist details) and \texttt{spotify\_show\_playlist\_library} (list all playlists) both enable locating a target playlist, with the choice depending on whether the playlist identifier is already known or requires search. In Gmail, \texttt{gmail\_mark\_thread\_archived} (direct archival operation) and the combination of \texttt{gmail\_show\_inbox\_threads} followed by archival represent two valid approaches depending on whether the thread identifier is available from prior context.

A key observation is that these errors disappear largely when evaluated at $Recall@9$ rather than $P@R$. This indicates that TabAgent correctly identifies the relevant functional category and includes both the ground-truth tool and its semantic neighbor within a slightly larger shortlist window. The discrepancy is therefore one of internal ranking rather than omission, and can be mitigated either by accepting a marginally larger shortlist budget in production (moving from $k{=}R_t$ to $k{=}R_t{+}2$) or by collecting additional trajectories that clarify the preference ordering through repeated usage patterns. Alternatively, reducing ambiguity at the tool catalog level (for instance, by consolidating near duplicate endpoints or providing clearer documentation about when to prefer full update versus incremental modification operations) would eliminate this category of error entirely.

\paragraph{Category 3: Genuine required-tool omissions (7\% of errors).}
A small fraction of discrepancies represent true shortlisting failures where TabAgent omits a tool that is necessary for task completion and has no functional substitute in the predicted shortlist. For example, in a Gmail task with the explicit requirement ``Send an email to the project team with the attached report,'' the ground truth includes \texttt{gmail\_send\_email}, but TabAgent's shortlist contained only \texttt{gmail\_create\_draft} and \texttt{gmail\_upload\_attachments\_to\_draft}. While drafting and attachment handling are precursors to sending, they do not fulfill the task's explicit sending requirement, and downstream execution would fail unless the agent's planner compensates by invoking additional tools outside the shortlist. Similarly, in a Phone task requiring ``delete all alarms set for weekdays,'' omitting \texttt{phone\_delete\_alarm} in favor of only \texttt{phone\_show\_alarms} and \texttt{phone\_update\_alarm} leaves the deletion requirement unmet. In an Amazon task (``Return the defective laptop I received last week''), TabAgent's shortlist included order viewing operations (\texttt{amazon\_show\_orders}, \texttt{amazon\_show\_order}) and return status checking (\texttt{amazon\_show\_returns}) but omitted the critical \texttt{amazon\_initiate\_return} endpoint necessary to begin the return process.

These genuine omissions are concerning because they directly threaten task success. However, their rarity (7\% of sampled errors) suggests that TabAgent's decision boundary closely approximates the generative shortlister on the dimensions that matter most for execution. We attribute these failures to the observation that some tasks exhibit low-frequency schema patterns (e.g., explicit delete operations, initiation actions, sending as opposed to drafting) that are underrepresented in the task training set. Delete and create actions appear less frequently in AppWorld trajectories than read and update operations, creating an imbalance that TabAgent's classifier inherits from the logged behavior. TabSynth's paraphrase augmentation mitigates this to some degree, but rewording existing rows, without adding new candidate relationships, may not fully cover tail patterns for rare action types. Increasing the training set size or explicitly oversampling trajectories containing low-frequency operations would likely reduce this error mode further.

\subsection{Downstream Impact and Implications}

Aggregating across categories, 93\% of observed errors (categories 1 and 2) have minimal to low downstream impact on task success. Optional-tool omissions are mitigated by alternative pathways, and semantically similar endpoint confusion is resolved by expanding the shortlist budget slightly or by rank-aware execution strategies that try the top-ranked tools in descending order until the task succeeds. Only the 7\% of genuine omissions pose a high risk of task failure, and even within this category, some tasks may recover if the agent's higher-level planner is sufficiently robust to detect missing capabilities and re-invoke the shortlister with refined context.

This error distribution is consistent with TabAgent's design goal: to reproduce the decision boundaries of a validated generative shortlister at reduced cost and latency, not to surpass it. The framework succeeds in this objective, preserving the core shortlisting capability while introducing a small tail of errors that are amenable to targeted fixing (additional training data, feature engineering for dependency chains, or slightly larger shortlist budgets). The strong shortlist recovery reported in Section~\ref{sec:results_new}, combined with this error analysis, suggests that most shortlist discrepancies would not propagate to execution failures. Re-running the full agent with \textsc{TabHead} in place confirms this directly: task success reaches parity with the generative shortlister (App.~\ref{app:inloop}). This supports the practical viability of the discriminative replacement.

\begin{figure}[t]
\centering
\includegraphics[width=0.55\linewidth]{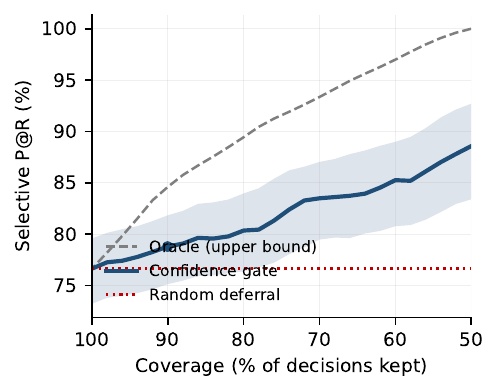}
\caption{Confidence-gated fallback on AppWorld shortlisting ($341$ held-out tasks, five folds). Deferring the least-confident decisions (moving right) lifts selective $\PatR$ on the kept set above random deferral, toward an oracle that defers by true $\PatR$. TabHead's confidence separates its correct from incorrect decisions at AUROC $0.88$, and deferred decisions are routed to the retained LLM. Band is a $95\%$ task bootstrap.}
\label{fig:gate}
\end{figure}

\paragraph{Confidence-gated fallback.} This error tail is separable by the head's own confidence. On the same held-out shortlisting decisions ($341$ tasks, five folds), the margin between TabHead's $R$-th and $(R{+}1)$-th ranked scores predicts whether a decision is fully correct with AUROC $0.88$ (95\% CI $[0.85, 0.92]$, $0.89 \pm 0.01$ across folds, $0.79$ to $0.92$ within each application where both outcomes occur), and a cutoff-free variant (the largest score gap, which needs no oracle set size) reaches $0.78$. Deferring the least-confident decisions to the generative shortlister therefore raises selective $\PatR$ over random deferral at every coverage (Fig.~\ref{fig:gate}). The absolute lift is small at low deferral because base $\PatR$ is already $0.77$. Since $G(t)$ is taken from GPT-5.4's own successful run, deferred cases are recovered by construction, so this establishes that the gate defers the right decisions, not that the LLM independently solves novel compositions. In the live agent the fallback also recovers real tasks: a held-out task that \textsc{TabHead} fails on its own succeeds once its shortlist is deferred to the LLM (App.~\ref{app:inloop}). The current error rate and impact distribution also support deployment without the gate, where a ${>}10^{3}\times$ latency and ${>}10^{5}\times$ cost reduction per call justifies a small tail of correctable errors.

\subsection{Cold start and catalog churn: added and renamed tools}
\label{app:error_analysis:churn}
The gate above defers decisions the head is unsure of. Here we measure the cold-start case that unseen \emph{combinations} of known tools do not cover: the tools below are absent from training entirely or arrive renamed.

\textbf{Added tools, and the cold-start floor.} We withhold one AppWorld application from training and re-admit $N$ of its trajectories (Fig.~\ref{fig:traj_efficiency}b). All figures in this paragraph are macro-averages over the five applications, since the application is the unit being onboarded. The corresponding full-pool, in-catalog reference on that basis is $0.836$. With none of its own traces a newly added application reaches $\PatR$ $0.377$, above zero-shot E5 dense retrieval ($0.328$) and BM25 ($0.281$) on the same basis, and far above chance ($0.086$). Cold start is therefore a degradation to retrieval quality rather than a collapse, because the trace-derived similarity features transfer even where candidate identities do not. The same ordering holds on the pooled task-level basis ($0.324$ against $0.282$, $0.224$ and $0.062$). The sign of early recovery follows catalog size rather than pool size: a single trajectory of the new application is worth $+0.43$ $\PatR$ on Phone ($30$ APIs) and $+0.26$ on SimpleNote ($17$), but is neutral to harmful on the larger catalogs ($-0.05$ Gmail, $-0.01$ Amazon, and $-0.13$ Spotify at $91$ APIs), where one trace makes the usage prior a near-degenerate one-hot over a handful of endpoints and misleads the ranker. We report the direction of this effect rather than a precise magnitude: the two applications that show it most clearly, SimpleNote and Spotify, have only $17$ held-out tasks each, so the per-application deltas carry wide intervals and the ordering across the five applications is the load-bearing observation, not any individual value. The mechanism we attribute it to, a single trace concentrating the usage prior on one or two endpoints while leaving the rest at zero, is consistent with the fact that the effect is negative exactly on the three largest catalogs, but we have not isolated it experimentally. Onboarding a large-catalog application therefore warrants the confidence gate until tens of traces have accumulated.

\textbf{Renamed tools.} We perturb \texttt{api\_name} in the test fold only, leaving training untouched (Table~\ref{tab:rename_churn}). Because the usage prior is keyed on the tool name, a rename breaks that lookup as well as the name text, so we separate the two: carrying the prior through the rename isolates the text effect, while breaking it reproduces real churn, where a renamed tool genuinely has no history. Carrying the prior costs at most $0.1$\,pp under any perturbation, including replacement of every name by an opaque hash. Breaking it costs $24$--$43$\,pp. On this card the head is therefore not relying on memorised tool names, and the practical remedy on a rename is to remap the usage statistics rather than to retrain. Two observations qualify the table: \texttt{camelCase} and hashing are equivalent because the word-level TF--IDF treats \texttt{send\_email} as a single token, so any edit to the name is a complete vocabulary miss, and synonym renaming is the most damaging perturbation, since substituted names collide with the vocabulary of other tools rather than being inert.

That conclusion is conditional on what the card contains, and we isolate the mechanism by ablating the AppWorld card itself rather than by comparing benchmarks. Removing description-bearing fields one at a time and repeating the hash rename (Table~\ref{tab:rename_ladder}) makes the head monotonically more fragile: $+0.1$\,pp with the full card, $-7.1$\,pp without the API description, and $-13.8$\,pp once the description-derived similarity feature is dropped as well and a candidate is identified by its name alone. Repeating the experiment on all four $\tau^2$ agents, whose cards carry no description by construction, gives $-22.3$\,pp, a larger drop than the AppWorld name-only rung, consistent with a card whose candidate features all derive from the tool name. Two honest qualifications: removing these fields also costs baseline accuracy ($0.793 \to 0.764 \to 0.730$), so the lower rungs are weaker heads as well as more fragile ones, and we do not read the $\tau^2$ synonym row, since the substitution list was written for AppWorld vocabulary and leaves most $\tau^2$ tool names untouched. The practical reading is a schema requirement rather than a caveat: a catalog expected to churn should be described in TabSchema by fields that do not depend on the candidate's name, and where it is, renaming reduces to bookkeeping.

\begin{table}[htbp]
\caption{\textbf{Rename fragility is created by removing description-bearing fields.} Cost of replacing every candidate name with an opaque hash at inference, with the usage prior carried across so only the text effect is measured (AppWorld, shipped folds, three seeds). Each row removes one more name-independent field from the card. The $\tau^2$ row is an independent benchmark whose card carries no description by construction, and, with every candidate feature derived from the tool name, it drops further than the AppWorld name-only rung.}
\centering
\footnotesize
\setlength{\tabcolsep}{6pt}
\begin{tabular}{lrrr}
\toprule
\textbf{Card identifies a candidate by} & \textbf{Control} & \textbf{Hashed} & \textbf{$\Delta$} \\
\midrule
name $+$ description $+$ similarity & 0.793 & 0.794 & $+0.1$ \\
name $+$ similarity                 & 0.764 & 0.694 & $-7.1$ \\
name alone                          & 0.730 & 0.592 & $-13.8$ \\
\midrule
$\tau^2$-bench, name alone by construction & 0.766 & 0.543 & $-22.3$ \\
\bottomrule
\end{tabular}
\label{tab:rename_ladder}
\end{table}

\textbf{Removed tools.} The third churn mode is deprecation: tools present at training time that later disappear, with the head left stale. We delete candidates at inference and re-score, recomputing $\PatR$ over the survivors (three seeds). Removing tools the task never needed is not merely harmless but helpful, since the ranking problem shrinks: dropping $10$, $25$ and $50\%$ of non-relevant candidates moves $\PatR$ from $0.793$ to $0.801$, $0.817$ and $0.846$. The head's scores therefore do not depend on the presence of irrelevant options, which is the failure this experiment was designed to detect. Deleting a candidate the task genuinely requires necessarily lowers the attainable score, but the ranking of what remains is not derailed: $\PatR$ over the surviving set falls only to $0.773$, $0.770$ and $0.748$ at the same three rates. Catalog shrinkage is thus the benign direction of churn, and needs no retraining.

\begin{table}[ht!]
\caption{\textbf{Rename damage is the lost usage prior, not the tool name.} Test-time-only perturbation of \texttt{api\_name} on AppWorld shortlisting, three seeds, $\PatR$. Carrying the usage prior through the rename isolates the effect of the name text. Breaking it reproduces realistic churn. Deltas are percentage points against the unperturbed control.}
\centering
\footnotesize
\setlength{\tabcolsep}{6pt}
\begin{tabular}{lrrrr}
\toprule
\textbf{Perturbation} & \textbf{Prior carried} & \textbf{$\Delta$} & \textbf{Prior broken} & \textbf{$\Delta$} \\
\midrule
None (control)       & ---   & ---    & 0.793 & ---     \\
\texttt{camelCase}   & 0.794 & $+0.1$ & 0.552 & $-24.1$ \\
Synonym substitution & 0.792 & $-0.1$ & 0.364 & $-43.0$ \\
Opaque hash          & 0.794 & $+0.1$ & 0.552 & $-24.1$ \\
\bottomrule
\end{tabular}
\label{tab:rename_churn}
\end{table}
\FloatBarrier

\section{In-the-Loop Validation: Live Hot-Swap on AppWorld}
\label{app:inloop}

\textbf{This appendix reports the end-to-end hot-swap experiment in full: the setup, the task-success statistics, a failure attribution that asks whether the shortlister is the bottleneck, the confidence-gated fallback measured in the loop, and an analysis of inherited bias.} The main-text summary is \S\ref{subsec:inloop}.

\paragraph{Setup.} We instrument the CUGA agent on AppWorld and replace only its shortlisting module, following the hot-swap step of the deployment lifecycle (Section~\ref{sec:setup}). Two arms run the identical agent on the GPT-5.4 backbone and differ solely in the shortlister: the stock generative GPT-5.4 shortlister, and \textsc{TabHead}, the deployed head of this paper scoring the live TabSchema card and returning its top $K{=}9$ candidates. Because everything downstream of the shortlist is held fixed, any difference in task success is attributable to the shortlisting decision. We evaluate on $24$ held-out cross-application AppWorld tasks (phone, spotify, and simple\_note), task-disjoint from training, each run over up to three seeds. This is our single complete, single-backbone hot-swap run, which we report in full.

\paragraph{Task success.} Table~\ref{tab:inloop_main} reports Pass@1 and the paired difference, with its standard error and a paired per-task bootstrap $95\%$ confidence interval ($2000$ resamples, matching App.~\ref{app:ci_protocol}). Over the $24$ held-out tasks the two shortlisters are at statistical parity, a paired difference of $-2.1$ points ($n{=}24$, standard error $9.1$, $95\%$ CI $[-18.8, +16.7]$) that includes zero. Following the paper's convention of averaging stochastic methods over seeds (\S\ref{sec:metrics}), we attribute the run-to-run spread to the sampling of the GPT-5.4 agent, not to \textsc{TabAgent}, whose decision is deterministic given the card: a paired gap of this size reflects the non-deterministic LLM backbone's variance across runs rather than the head being less capable.

\subsection{Failure attribution: is the shortlister the bottleneck?}
\textbf{Most tasks that \textsc{TabAgent} fails are tasks the LLM shortlister also fails, so the shortlisting decision is not where these tasks are won or lost.} Table~\ref{tab:inloop_flip} cross-tabulates the two arms by pass or fail. They agree on $79\%$ of tasks, a Cohen's $\kappa$ of $0.52$, which is moderate agreement. Of the $17$ tasks \textsc{TabAgent} fails, $14$, that is $82\%$, are also failures of the GPT-5.4 shortlister, tasks where the agent gets lost or the task is simply hard, and where a better shortlist would not have changed the outcome. Only $3$ of the $24$ paired tasks are ones the LLM solves and \textsc{TabAgent} does not, so the most a perfect shortlist gate could recover is $+3$ tasks, and $2$ go the other way. This is the strongest evidence we have that the swap does not silently degrade task success: the classifier reaches the same outcomes as the generative shortlister on the large majority of tasks, and it shares its failures.

\begin{table}[t]
\centering
\small
\caption{\textbf{Failure attribution on the paired held-out tasks ($n{=}24$).} Each task is scored pass or fail by seed-majority under each shortlister, and the table counts the four joint outcomes. The off-diagonal cells are the only tasks a shortlist swap could change, and both are small and balanced. The large both-fail cell is shared task difficulty, not a shortlisting failure.}
\label{tab:inloop_flip}
\begin{tabular}{lcc}
\toprule
 & GPT-5.4 pass & GPT-5.4 fail \\
\midrule
\textsc{TabAgent} pass & $5$ & $2$ \\
\textsc{TabAgent} fail & $3$ & $14$ \\
\bottomrule
\end{tabular}
\end{table}

\subsection{Confidence-gated fallback, in the loop}
\textbf{The fallback that defers low-confidence decisions to the LLM works as a mechanism in the loop, yet its ceiling on task success is small here because the shortlister is rarely the deciding factor.} App.~\ref{app:error_analysis} shows the head's confidence flags its own shortlist errors well, at AUROC $0.88$. Deferring a shortlist to the LLM does recover tasks in practice, we observe a held-out task that \textsc{TabAgent} fails on its own turn into a success once its shortlist calls are deferred to GPT-5.4, a real recovery rather than one recovered by construction as in App.~\ref{app:error_analysis}. The task-success ceiling is nonetheless set by the attribution above, since only $3$ of the $24$ paired tasks have a shortlist-attributable failure, even a perfect gate lifts task success by at most those $3$ tasks. Confidence that separates correct from incorrect shortlists therefore does not translate into separating solved from failed tasks, because most failures are shared with the LLM and lie downstream of the shortlist. We recommend \textsc{TabAgent} without a gate as the default in this deployment, consistent with the without-gate recommendation of App.~\ref{app:error_analysis}, and keep the gate available for the small, correctable tail.

\subsection{Bias inheritance from the source agent}
\label{app:inloop:bias}
\textbf{\textsc{TabAgent} reproduces the shortlisting boundary of the agent it distills, and it inherits that agent's biases along with its competence, which is the intended behavior for a component built to replace an already validated one.} The head is trained on the successful trajectories of a shortlister we already trust, so the target it learns is that shortlister's decision boundary, and reproducing the boundary means reproducing both its correct calls and its biases. The in-loop concordance measures this inheritance directly, the two arms agree on $79\%$ of task outcomes at Cohen's $\kappa{=}0.52$, and offline the head recovers $92$--$100\%$ of the tools the source agent selects on AppWorld (Recall@9, \S\ref{subsec:retrieval_new}), so at both the outcome level and the decision level the head tracks the source. The inherited skew shows up where the source's does. The usage-prior imbalance that \textsc{TabAgent} absorbs from logged behavior, the under-representation of delete and create actions noted in App.~\ref{app:error_analysis}, is the source agent's imbalance carried through the labels rather than a distortion introduced by the head. This is consistent with the paper's two separate findings, that the TabSchema representation is model-agnostic (App.~\ref{app:model_dependency}) while the label set $G(t)$, and therefore the learned boundary, is inherited from the source agent's successes. We treat inheritance as the design goal and not a defect, the aim is to reproduce a validated component at far lower cost, so absorbing its boundary, biases included, is what a faithful and cheap replacement should do, and the residual arm-to-arm differences we observe sit within the run-to-run stochasticity of the GPT-5.4 agent rather than indicating that the head is better or worse than the shortlister it copies.

\textbf{The head does not amplify the inheritance, it is more consistent than the agent it copies.} Outcome agreement between the two arms must be read against how often each arm agrees with itself, because the GPT-5.4 agent is stochastic and does not reproduce its own task outcomes across seeds. On the held-out tasks the GPT-5.4 shortlister agrees with itself on $93\%$ of seed pairs, while \textsc{TabAgent} agrees with itself on $96\%$, its shortlist being deterministic given the card. The $79\%$ cross-arm agreement sits below the $93\%$ self-agreement ceiling that the agent's own stochasticity allows, so part of the residual disagreement is the agent resampling its trajectory and part is the head departing from the source. A distillation that inherited and then reinforced the source's errors would also grow less stable than the source. We observe the opposite on stability: the head is the more deterministic of the two.

\textbf{A single decision makes the inheritance concrete.} On the held-out task ``initiate returns via FedEx for everything in my last three Amazon orders'', the successful source trajectory used exactly three of the $66$ candidate tools, \texttt{show orders}, \texttt{initiate return}, and \texttt{show return deliverers}. \textsc{TabAgent} places those same three tools at ranks one, two, and three, reproducing both the set the source selected and its ordering. The head is not selecting tools by generic relevance, it is reproducing the specific choice the source agent made, which is what lets it stand in for that agent and what carries the source's preferences forward.

\textbf{Where the source's bias is unacceptable, it should be corrected in the source, not in the head.} Because the head reproduces the source's boundary, the classifier is the wrong place to intervene, a correction applied only to the head would pull it away from the behavior it is meant to preserve. The place to detect and remove an unwanted bias is the upstream agent, before distillation, by auditing its successful traces for the pattern in question, for example the delete and create under-representation of App.~\ref{app:error_analysis}, and either rebalancing the trajectory pool or adjusting the agent until the traces it emits no longer carry the bias. \textsc{TabAgent} then distills the corrected behavior at the same cost. This keeps a clean separation of concerns, the agent owns what a good decision is, and the head owns making that decision cheaply.

\paragraph{Limitations of this validation.} The study covers one agent and one benchmark, and the sample is modest. We launch three seeds per task, as in the offline evaluation (\S\ref{sec:metrics}), and not more, because each seed here is a full end-to-end agent rollout of many GPT-5.4 calls rather than a single scoring pass, so the cost of scaling this evaluation is the very cost \textsc{TabAgent} exists to reduce. The confidence intervals we report fold this limited sample into their width. We present it as a promising in-loop signal that the offline decision transfers to real task success, not as a definitive cross-agent claim.

\section{TabSchema Dependency Analysis Across Frontier LLMs}
\label{app:model_dependency}

TabSchema operates as an offline feature extraction pipeline that runs once on historical trajectory data, not as a runtime dependency. The operational sequence follows three phases: TabSchema processes logged trajectories to generate a feature card as a one-time setup, TabHead trains on the extracted features in under one minute on CPU, and TabHead makes inference predictions using only the compact classifier during repeated per-task execution. This property provides significant flexibility in model selection for TabSchema, as organizations can justify different models during the offline extraction phase to maximize downstream TabHead performance without affecting inference costs.

\subsection{Cross-Model Validation Experiment}

To assess whether TabSchema's feature extraction depends critically on GPT-5.4 or generalizes across frontier LLMs, we conducted a robustness experiment using three different models on CUGA trajectories across AppWorld applications. We applied TabSchema's complete hierarchical orchestration (FeatureAnalyzer, CodeExecutor, CodeValidator, three Judges, MetaJudge) using GPT-5.4 (OpenAI, original configuration), Gemini 2.5 (Google, via API), and Claude 4.5 Sonnet (Anthropic, via API). Each model independently executed the full TabSchema pipeline without access to the others' outputs, producing separate feature cards that we then compared through manual analysis and automated semantic matching using sentence embeddings with BERTScore-style comparison applied to feature names concatenated with feature descriptions.

\begin{table}[ht!]
\caption{Cross-model feature extraction comparison across trajectories. Three-way agreement represents the 9 features proposed by all three models (69\% of GPT-5.4 baseline). GPT-5.4 extracts a balanced feature set, Gemini 2.5 produces expansive coverage with finer-grained text statistics, and Claude 4.5 Sonnet consolidates features into fewer robust representations.}
\centering
\footnotesize
\setlength{\tabcolsep}{10pt}
\renewcommand{\arraystretch}{1.15}
\begin{tabular}{lcccc}
\toprule
\textbf{Model} & \textbf{Total} & \textbf{Unique to} & \textbf{Overlap} & \textbf{Three-way} \\
 & \textbf{Features} & \textbf{This Model} & \textbf{w/ GPT-5.4} & \textbf{Agreement} \\
\midrule
GPT-5.4 & 13 & 4 & --- & 9 \\
Gemini 2.5 & 21 & 12 & 9 (69\%)  & 9 \\
Claude 4.5 Sonnet & 11 & 2 & 9 (69\%)  & 9 \\
\bottomrule
\end{tabular}
\label{tab:cross-model-features}
\end{table}

\begin{table}[ht!]
\caption{Semantic alignment and importance ranking agreement across frontier models. Semantic similarity computed using sentence embeddings with BERTScore-style comparison on feature names concatenated with descriptions, ranging from 0 (unrelated) to 1 (identical). Kendall tau measures agreement between two annotators (part of the research team) who independently ranked shared features by importance for the tool shortlisting task after reviewing feature definitions and sample trajectories. For each model pair, we computed Kendall tau between each annotator's rankings of the two models' features, then averaged across annotators.}
\centering
\footnotesize
\setlength{\tabcolsep}{10pt}
\renewcommand{\arraystretch}{1.15}
\begin{tabular}{lcc}
\toprule
\textbf{Comparison} & \textbf{Semantic Similarity} & \textbf{Importance Ranking} \\
 & \textbf{(BERTScore)} & \textbf{Agreement (Kendall $\tau$)} \\
\midrule
GPT-5.4 vs Gemini 2.5 & 0.79 & 0.65 \\
GPT-5.4 vs Claude 4.5 Sonnet & 0.88 & 0.78 \\
Gemini 2.5 vs Claude 4.5 Sonnet & 0.76 & 0.61 \\
\midrule
Mean across all pairs & 0.81 & 0.68 \\
\bottomrule
\end{tabular}
\label{tab:semantic-alignment}
\end{table}

The three models produced feature sets of varying size with substantial overlap (Table~\ref{tab:cross-model-features}). GPT-5.4 extracted 13 features after MetaJudge filtering, representing balanced coverage of schema signals (taxonomy depth, argument types), state indicators (thought traces, plan status), and dependency patterns (tool co-usage counts). Gemini 2.5 proposed the most expansive feature set with 21 features, reflecting finer-grained decomposition of state signals and detailed text statistics. The additional features Gemini proposed primarily captured secondary text properties such as sentiment indicators in agent reasoning, lexical diversity metrics, and granular temporal markers. Claude 4.5 Sonnet extracted the most conservative feature set with 11 features, prioritizing high-impact signals and consolidating related features into broader representations. For instance, where Gemini proposed separate features for thought content at different lookback windows (current thought, previous thought, two steps ago), Claude consolidated these into a single aggregated thought history feature.

We found that 9 features were proposed by all three models (approximately 69 percent three-way agreement with GPT-5.4's set), with no feature shared by exactly two models. The 9 features with three-way agreement predominantly covered core schema signals (taxonomy depth, argument types, input-output cardinality), primary state indicators (current thoughts, overall status analysis, summary of progress), and high-frequency dependency patterns (co-usage counts, success signals). Gemini's twelve unique features emphasized text-level statistics and sentiment analysis. Claude's two unique features reflected alternative consolidations of temporal state markers. GPT-5.4's four unique features included tool invocation frequency within execution windows, explicit precedence dependencies between tool pairs, argument schema compatibility checks, and step position within the agent trajectory.

GPT-5.4's four unique features target signals the other models did not name explicitly. Tool invocation frequency within the current execution context serves as a strong prior for shortlisting when tasks exhibit iterative patterns. Explicit precedence dependencies extracted from successful tool chains provide crucial signal for multi-step workflows where order matters. Argument schema compatibility checks prevent shortlisting of tools that cannot be invoked due to missing preconditions. Step position within the agent trajectory captures phase-dependent tool selection patterns. The absence of these features in Gemini's and Claude's proposals suggests that GPT-5.4's feature engineering captured dependency and precondition patterns that the other models either consolidated into broader representations or overlooked entirely.

We computed semantic similarity using sentence embeddings with BERTScore-style comparison applied to feature names concatenated with descriptions (Table~\ref{tab:semantic-alignment}). The mean similarity score across all pairwise comparisons was 0.81, with scores of 0.79 between GPT-5.4 and Gemini, 0.88 between GPT-5.4 and Claude, and 0.76 between Gemini and Claude. The higher similarity between GPT-5.4 and Claude (0.88) reflects Claude's consolidation strategy, which produced feature definitions matching GPT-5.4's level of abstraction. To assess whether models agreed on feature importance for the tool shortlisting task, two annotators from the research team independently ranked the 9 shared features by predicted importance after reviewing feature definitions and examining sample trajectories. For each model pair, we computed Kendall tau correlations between each annotator's rankings of the two models' features, then averaged across the two annotators. The mean Kendall tau across all pairwise model comparisons was 0.68, with the highest agreement between GPT-5.4 and Claude (0.78) and the lowest between Gemini and Claude (0.61). Features like intent, task description, current agent thoughts, and overall status analysis received consistently high importance rankings across all three models, while lower-ranked features showed more variation depending on each model's engineering philosophy.

These results support two conclusions about TabSchema's robustness. First, TabSchema extracts genuine decision-relevant structure from execution traces rather than model-specific artifacts, evidenced by 69 percent three-way agreement and 0.81 mean semantic similarity. State and dependency indicators appear in every model's feature set, suggesting these signals are inherent to the execution traces. Second, the variation in feature set size (11 to 21 features) reflects different feature engineering philosophies rather than fundamental disagreement. Gemini's expansive approach captured additional nuance at the risk of including lower-impact features, while Claude's conservative consolidation prioritized robustness at the potential cost of missing granular dependency patterns. GPT-5.4's balanced extraction, which includes both the 9 features with three-way agreement and four additional dependency signals, suggests it occupies a favorable point in the coverage-parsimony tradeoff.

From a practical deployment perspective, organizations are not locked into GPT-5.4 and can select among available frontier models based on API cost structures, data residency requirements, or existing licensing agreements. However, we emphasize that this validation was conducted only on feature extraction. We did not re-run the full TabHead training and evaluation pipeline using Gemini-derived or Claude-derived features due to computational budget constraints, so we cannot definitively quantify the end-to-end impact on metrics like $P@R$ or $Recall@k$. The strong semantic alignment (0.81 mean similarity, 0.68 annotator rank correlation) suggests that downstream performance should be similar for the 9 features with three-way agreement, but the absence of GPT-5.4's unique features in other models' outputs could cause performance gaps that empirical validation would clarify.

\paragraph{Same-model reruns.} We also ran TabSchema three times independently with one model (Claude Opus, used only for this stability check) on a $29$-task subset of the OAK and BPO trajectories, with no shared context between runs. Each run proposed $22$ features. Exact feature names rarely coincide ($1.6\%$ three-way name overlap), but the proposals match semantically ($72$--$95\%$ pairwise coverage at string similarity of at least $0.5$), agree on $16$ of $17$ feature categories ($94.1\%$), and cover $26$ of $28$ target APIs ($92.9\%$). Across three independent TabSynth generations, downstream $\PatR$ varies with a standard deviation of $3.9$ points on OAK and $5.9$ points on BPO.

\subsection{Open-Source Model Limitations}

We attempted to apply TabSchema using open-source models that represent more accessible options for organizations unable or unwilling to use commercial APIs. We tested three open-source alternatives: Llama 3.1 70B Instruct (8 trajectories), Llama 4 Maverick (6 trajectories), and GPT-OSS 120B (5 trajectories). The results revealed significant implementation challenges that currently make open-source alternatives unreliable for TabSchema without substantial additional engineering.

The FeatureAnalyzer stage frequently produced malformed JSON outputs that failed schema validation. Llama 3.1 70B exhibited parsing failures in approximately 6 of 8 attempts, Llama 4 Maverick in 5 of 6 attempts, and GPT-OSS 120B in 4 of 5 attempts, requiring manual intervention in each case. When feature specifications did parse successfully, the CodeExecutor stage generated Python extraction code with systematic errors. Llama 3.1 70B produced syntax errors (mismatched indentation, undefined variables, incorrect dictionary access patterns), with approximately 4 of the 8 trajectories exhausting the three-retry self-repair budget. Llama 4 Maverick showed marginal improvement but still failed to produce executable code on first attempt for 4 of 6 trajectories, with two cases requiring manual debugging after retry exhaustion. GPT-OSS 120B exhibited the most severe code generation issues, with only 1 of 5 trajectories producing valid extraction code within the retry budget. We observed that GPT-OSS frequently outputted code inside the reasoning section of the pydantic structure rather than in the designated code field.

Additionally, all three open-source models occasionally proposed features that were not extractable from the trajectory data, such as referencing fields that do not exist in the CUGA schema or hallucinating tool relationships not present in the catalog. These hallucinated features survived the CodeValidator stage more frequently than with frontier models, particularly for GPT-OSS 120B where 3 of 5 trajectories included at least one invalid feature in the final feature card. Llama 3.1 70B and Llama 4 Maverick showed better validation performance but still produced hallucinated features in approximately 2 of their respective trajectory samples.

Given these difficulties, we concluded that current open-source alternatives are not yet reliable enough for TabSchema's structured orchestration without substantial prompt engineering, schema constraints, or human-in-the-loop supervision. The gap manifests most acutely in the code generation phase, where syntactic correctness, variable scoping, and error handling require precise attention to detail that current open-source models struggle to maintain consistently.

\section{Licenses and Intended Use of Scientific Artifacts}
\label{app:licenses}

For completeness, we summarize the licenses and intended-use compatibility for the scientific artifacts (benchmarks, models, and code) used in this work. All artifacts are used solely for non-commercial academic research, which is consistent with each artifact's stated terms and original release intent. Verification was performed by inspecting the official \texttt{LICENSE} files in the corresponding source repositories and model cards.

\begin{table}[h]
\centering
\footnotesize
\setlength{\tabcolsep}{4pt}
\renewcommand{\arraystretch}{1.15}
\caption{Licenses and intended-use compatibility for artifacts used in this work. ``Compatible'' indicates that the artifact's license and stated intended use permit the non-commercial academic research conducted here.}
\label{tab:artifact-licenses}
\begin{tabular}{l l c}
\toprule
\textbf{Artifact} & \textbf{License} & \textbf{Compat.} \\
\midrule
AppWorld \citep{trivedi2024appworld}     & Apache 2.0          & Yes \\
$\tau^2$-bench \citep{barres2026taubench}  & MIT                 & Yes \\
CUGA \citep{marreed2025towards}          & Apache 2.0          & Yes \\
TabSTAR \citep{arazi2025tabstar}         & MIT (code)          & Yes \\
e5 encoder \citep{wang2022text}          & MIT                 & Yes \\
Llama 3.1-8B-Instruct \citep{dubey2024llama}      & Llama 3.1 Community & Yes \\
Llama 3.2-1B/3B-Instruct \citep{dubey2024llama}   & Llama 3.2 Community & Yes \\
OAK \citep{cugaproject2026oakbench}        & Open license & Yes \\
BPO \citep{shlomov2026benchmarks}        & Open license & Yes \\
SkillsBench \citep{li2026skillsbench}      & Apache 2.0          & Yes \\
Mind2Web \citep{deng2023mind2web}          & CC BY 4.0           & Yes \\
LLMRouterBench \citep{li2026llmrouterbench} & MIT                & Yes \\
$\tau^2$-bench SFT trajectories (HF)       & Apache 2.0          & Yes \\
GPT-5.4 (API)                            & Provider Terms of Service & Yes \\
Claude 3.7 / 4.5 Sonnet, Claude Opus (API) & Provider Terms of Service & Yes \\
Gemini 2.5 (API)                         & Provider Terms of Service & Yes \\
GPT-4.1 / 4.1-mini / o4-mini (API)       & Provider Terms of Service & Yes \\
Qwen2.5-7B \citep{qwen2_5}               & Apache 2.0          & Yes \\
Llama 3.1-70B / Llama-4-Maverick         & Llama Community     & Yes \\
GPT-OSS-120B                             & Apache 2.0          & Yes \\
BM25 \citep{robertson2009probabilistic}  & Public algorithm (n/a) & Yes \\
\bottomrule
\end{tabular}
\end{table}

The seven evaluation benchmarks are released by their respective authors for the evaluation of LLM agents, which is precisely the use we make of them. CUGA is released as an open-source agent framework under Apache 2.0 and is used here as the upstream agent producing the trajectory pool from which TabAgent is trained, consistent with its research-use intent. TabSTAR is released as a tabular foundation model and is used here only as the neural alternative head in our head ablation (App.~\ref{app:head_alt}), matching its intended use for tabular classification. The TabSTAR weights hosted publicly do not currently declare an explicit license field on their model card, so we rely on the MIT license of the corresponding code release, which permits the research use undertaken here. The e5 encoder is released under MIT for general-purpose text representation and is used as the shared encoder for the DSR baselines and the textual side of TabHead. The Llama 3.1 and 3.2 Instruct variants are released under the Llama Community Licenses, which permit non-commercial and small-scale commercial research use. We use them only as generative shortlister baselines. The OAK and BPO benchmarks \citep{shlomov2026benchmarks} are publicly available under an open license, and both use simulated, non-personally-identifying user data. GPT-5.4 is accessed via a commercial API under the provider's standard terms of service, which permit research and evaluation use. BM25 is a public information-retrieval algorithm rather than a licensed artifact. We use standard open-source implementations under their respective permissive licenses. The additional API models (Claude~3.7 and~4.5 Sonnet, Gemini~2.5, GPT-4.1, GPT-4.1-mini, and o4-mini) are accessed under their providers' standard terms of service and appear only in the $\tau^2$ cross-backbone transfer study (Table~\ref{tab:transfer_new}) and the cross-model TabSchema analysis (App.~\ref{app:model_dependency}). Qwen2.5-7B (Apache~2.0) is a generative shortlister baseline, and the larger open weights (Llama~3.1-70B, Llama-4-Maverick, and the Apache-2.0 GPT-OSS-120B) appear only in the open-source feasibility study of App.~\ref{app:model_dependency}.

\section{Terms, Common Questions, and Released Resources}
\label{app:peo}

This appendix gathers the paper's terms in one place, answers the questions a reader is likely to
bring to it, each with a pointer to the full account, and lists what is released.

\paragraph{Keywords.} LLM agents, closed-set decisions, tool shortlisting, tool selection, skill
selection, application selection, GUI element selection, model routing, sub-agent routing, skill
gating, LLM routing, agent efficiency, execution traces, tabular classification, distillation,
CatBoost, AppWorld,
$\tau^2$-bench, OAK, BPO, SkillsBench, Mind2Web, LLMRouterBench.

\subsection{Terms}

\begin{description}
\item[Closed-set decision.] A decision that selects an answer from a fixed, bounded set of
  candidates: which skill to invoke, which tool to call next, which application to route to, or
  which sub-task to gate (Section~\ref{sec:intro}).
\item[Tool.] Any externally callable capability (API endpoint, function, or GUI operation) with a
  typed schema and observable outputs (Section~\ref{sec:intro}).
\item[Tool shortlisting.] The task of picking a small subset from a catalog of hundreds to
  thousands of tools (Section~\ref{sec:intro}).
\item[Feature card.] The tabular layout \textsc{TabAgent} learns from, with one row per (context,
  candidate) pair and one column per behavior-grounded feature (Section~\ref{sec:method}).
\item[TabSchema.] An orchestrated agentic system that proposes features, implements them as
  executable trace-extraction code, and keeps only those a judge ensemble agrees are
  decision-relevant (Section~\ref{sec:tabschema}).
\item[TabSynth.] A schema-aligned augmentation module. A deterministic, rule-based paraphraser
  rewrites the free-text state of an existing (context, candidate) training row while copying the
  candidate, the label, and every schema, state, and dependency feature verbatim, so it cannot
  fabricate a positive (Section~\ref{sec:tabsynth}).
\item[TabHead.] The classifier that scores each (context, candidate) pair over the feature card in
  a single pass. It is a gradient-boosted decision tree (CatBoost) in every main-paper result
  (Section~\ref{sec:tabhead}).
\item[R-precision ($\PatR$).] The share of a task's relevant tools ranked within its first $R_t$
  positions, where $R_t$ is the size of the task's relevant set. It adapts the cutoff to that size,
  captures completeness, and equals Recall@$R$ (Section~\ref{sec:metrics}).
\item[Recall@$k$.] The share of a task's relevant tools found among the top $k$ candidates. It
  evaluates recovery at a fixed budget (Section~\ref{sec:metrics}).
\item[Oracle and normalized score.] On policy and reward decisions, the oracle is the fraction of
  tasks that any candidate can solve, and
  $\mathrm{Norm}=(\text{method}-\text{chance})/(\text{oracle}-\text{chance})$ is the share of the
  chance-to-oracle headroom a method captures (Section~\ref{sec:results_new}).
\item[Best default.] The single best choice applied to every task: one sub-agent or model for all
  tasks, or always activating the skill (Table~\ref{tab:master_new}).
\item[Hot-swap.] Swapping the generative module of a running agent for the discriminative TabHead
  while preserving the surrounding agent architecture, the last step of the deployment lifecycle
  (Section~\ref{sec:setup}).
\end{description}

\subsection{Common questions}

\paragraph{What is TabAgent?}
\textsc{TabAgent} is a framework that replaces a generative decision head in a running LLM agent
with a compact classifier trained on the agent's own successful execution traces
(Section~\ref{sec:intro}). It targets closed-set decisions such as tool shortlisting, skill
selection, application selection, and routing, and it leaves the surrounding agent intact.
TabSchema extracts schema, state, and dependency features from trajectories, TabSynth adds
schema-aligned synthetic rows where trajectory coverage is thin, and TabHead, a CatBoost
classifier, scores every candidate in a single pass (Section~\ref{sec:method}).

\paragraph{Which decisions and benchmarks does the paper evaluate?}
The paper tests \textsc{TabAgent} on three families of closed-set decisions across seven agent
benchmarks (Section~\ref{sec:setup}, Appendix~\ref{app:benchmarks}). Set retrieval is tool
shortlisting on AppWorld \citep{trivedi2024appworld}, $\tau^2$-bench \citep{barres2026taubench},
OAK \citep{cugaproject2026oakbench}, and BPO \citep{shlomov2026benchmarks}, and skill selection on
SkillsBench \citep{li2026skillsbench}. Classification is application selection on AppWorld and
operation and GUI element selection on Mind2Web \citep{deng2023mind2web}. Policy and reward is
sub-agent routing on AppWorld, model routing on LLMRouterBench \citep{li2026llmrouterbench}, and
skill gating on SkillsBench. Candidate sets range from a binary gate to $195$ skills.

\paragraph{How well does TabAgent shortlist tools?}
On AppWorld tool shortlisting, \textsc{TabAgent} with TabSynth reaches Recall@9 of $92$ to $100$
across the Amazon, Gmail, Phone, SimpleNote, and Spotify applications, recovering essentially all
of the tools the GPT-5.4 shortlister used on its successful runs (Section~\ref{sec:results_new},
Table~\ref{tab:master_new}). Under the stricter R-precision it leads or ties every baseline at
comparable cost. Fine-tuned small LLMs score slightly higher, at ${\sim}100$--$250\times$ the
cost even on a GPU. Per-domain heads reach Recall@9 of $96$ to $100$ on $\tau^2$-bench, and OAK and
BPO reach $85$ and $97$.

\paragraph{How much faster and cheaper is TabAgent than the LLM call it replaces?}
The gain in agent efficiency is per decision. One TabHead pass runs in $2.682$\,ms at \$$2.0\times10^{-7}$, against $7.50$\,s and \$$0.052$ for
the GPT-5.4 shortlister it replaces, which is $99.96\%$ lower latency and $99.9996\%$ lower cost
per call, roughly $10^{3}\times$ and $10^{5}\times$ (Section~\ref{sec:results_new},
Table~\ref{tab:cost-runtime}). The head runs on CPU and needs no GPU. A fine-tuned Llama stays
${\sim}100$--$250\times$ TabHead's latency and cost on a V100 GPU, and $5{,}000$--$27{,}000\times$
on a CPU node (Appendix~\ref{app:cost_efficiency}).

\paragraph{Does the classifier matter, or the features?}
The trace-derived feature card carries the decision, and the choice of classifier head barely
matters (Section~\ref{sec:results_new}). On skill selection, with the TabSchema card held fixed, heads from classical
trees to tabular foundation models reach comparable R-precision, and all collapse to an R-precision
of about $0.10$ to $0.21$ when the card is reduced to raw task and candidate text or embeddings
(Figure~\ref{fig:card_new}). The same head trained on CUGA traces beats one trained on text-only
CodeAct traces by $24.7\%$ R-precision, and replacing every tool name with an opaque hash costs at
most $0.1$ point with the full card.

\paragraph{Can TabAgent make routing and gating decisions, such as LLM routing?}
Yes, at the level of a strong fixed default rather than above it. On sub-agent routing over $14$
AppWorld agent systems, model routing over $12$ models on LLMRouterBench, and skill gating on
SkillsBench, \textsc{TabAgent} matches the best fixed default within a paired task-level bootstrap
$95\%$ confidence interval and captures $22$ to $54\%$ of the chance-to-oracle headroom, at orders
of magnitude lower latency and cost than a generative router (Section~\ref{sec:results_new}).
Few-shot GPT-5.4 falls at or below the default on all three decisions. The oracle itself is only
$56$ to $94\%$.

\paragraph{Does swapping TabAgent into a live agent preserve task success?}
In the one live study the paper reports, \textsc{TabAgent} completes AppWorld tasks about as often
as the GPT-5.4 shortlister it replaces. Replacing only CUGA's shortlister with TabHead on $24$
held-out cross-application tasks gives $27.1\%$ Pass@1 against $29.2\%$, a paired difference
of $-2.1$ points whose $95\%$ bootstrap confidence interval, $[-18.8, +16.7]$, includes zero
(Section~\ref{sec:results_new}, Table~\ref{tab:inloop_main}). The two arms reach the same
pass-or-fail verdict on $79\%$ of tasks. The study covers one agent and one benchmark on a modest
sample (Appendix~\ref{app:inloop}).

\paragraph{How many trajectories does TabAgent need?}
\textsc{TabAgent} needs tens to low hundreds of successful traces, not thousands. On AppWorld,
TabHead reaches half of its full-pool R-precision from $8$ trajectories and $90\%$ of it from
$118$, and the four $\tau^2$-bench agents need $37$ to $51$ (Section~\ref{sec:results_new},
Appendix~\ref{app:cost_efficiency:traj}). The requirement depends on the benchmark: tens for a
single homogeneous catalog and low hundreds for a heterogeneous multi-application one. An
application admitted with none of its own traces still reaches an R-precision of $0.377$, above
zero-shot dense retrieval, and recovers as traces accumulate (Appendix~\ref{app:scope_limitations}).

\paragraph{What are the limitations?}
\textsc{TabAgent} learns a decision head's behavior from that head's own successful trajectories,
so its reach is bounded by what those trajectories contain (Appendix~\ref{app:scope_limitations}).
It does not zero-shot to a domain with no trajectory pool. Its cross-agent transfer, a penalty of
about $3.1$ points of R-precision, holds among four ReAct agents that differ only in their LLM
backbone, so it is not evidence of transfer across architectures. It inherits the source
shortlister's biases by design. TabSynth can hurt, as on Spotify ($70$ to $61$ R-precision), and
low-confidence decisions can be deferred to the LLM (AUROC $0.88$).

\paragraph{How is leakage prevented?}
Every feature is read from what the head would see at the moment of the decision, and nothing
written after it, such as the tools later called or the run's outcome, enters the feature card.
Every data-dependent stage is fitted inside the training portion of each fold and never sees
held-out tasks (Section~\ref{sec:metrics}).

\subsection{Where this work sits}

\textsc{TabAgent} sits between three lines of work (Section~\ref{sec:related}). LLM-centric tool
learning \citep{qintoolllm, schick2023toolformer, shen2023hugginggpt, patil2024gorilla} raises
tool-use accuracy but keeps a generative controller on the critical path, and tool-use benchmarks
\citep{li_etal_2023_api, song2023restgpt, tang2023toolalpaca} evaluate tool selection offline and
rarely measure latency and cost, so the paper evaluates on long-horizon, cross-application
environments such as AppWorld \citep{trivedi2024appworld}. Tool retrieval, from BM25 \citep{robertson2009probabilistic} and dense
retrieval \citep{karpukhin_etal_2020_dense} to structure-aware methods \citep{anantha2023protip,
zheng2024toolrerank, qu2024towards}, avoids the LLM but scores a static catalog, and the closest
work casts shortlisting as pointwise classification over static embeddings
\citep{moon2024efficient}. Cost-aware routing and cascades \citep{chen2024frugalgpt,
ong2025routellm, ding2024hybrid} and distillation and self-training \citep{hinton2015distilling,
zelikman2022star, chen2023fireact, zeng2024agenttuning} make the generative call cheaper or smaller
but keep it generative. \textsc{TabAgent} conditions on live execution state with no generative
call, following the view that small, specialized models should carry most agentic work
\citep{belcak2025small}. It extends LLM-driven feature engineering \citep{hollmann2023large} to
agent execution traces and treats tabular learners, from CatBoost \citep{CatBoost} to TabSTAR
\citep{arazi2025tabstar}, as interchangeable heads.

\end{document}